\RequirePackage{fix-cm}

\documentclass[twocolumn,twoside]{svjour3}

\usepackage[switch]{lineno} 
\usepackage{graphicx}
\usepackage{comment}
\usepackage{amsmath,amssymb} 
\usepackage{color}
\usepackage[font=normalsize]{subfig}
\usepackage{blkarray}
\usepackage{threeparttable}
\usepackage{multirow}
\usepackage{makecell}
\usepackage{colortbl}
\usepackage{tikz}
\usepackage{fbox}
\usepackage{array}
\usepackage{booktabs}
\usepackage{wrapfig}
\usepackage{yhmath}
\usepackage{enumerate}
\usepackage{rotating}
\usepackage{diagbox}
\usepackage[authoryear]{natbib}

\usepackage{marginnote} 

\usepackage{url}
\usepackage[colorlinks,linkcolor=red,anchorcolor=blue,citecolor=blue]{hyperref}
\usepackage[linesnumbered,ruled,vlined]{algorithm2e}
\usepackage[misc]{ifsym}
\usepackage{circledsteps}

\usepackage{bbm}
\graphicspath{{./}}

\usepackage{mathtools}

\usepackage{verbatim}

\renewcommand{\aboverulesep}{0pt}
\renewcommand{\belowrulesep}{0pt}

\newcommand{\etal}{et al.}

\newcommand{\ieno}{\textit{i.e.}}

\newcommand{\egno}{\textit{e.g.}}

\newcommand{\tct}{\textcolor{black}}

 \journalname{International Journal of Computer Vision}
\begin{document}

\title{Diffusion Models for Image Restoration and Enhancement: A Comprehensive Survey}

\author{{Xin Li}\textsuperscript{1,2}        \and
        {Yulin Ren}\textsuperscript{1} \and
        {Xin Jin}\textsuperscript{3} \and
        {Cuiling Lan}\textsuperscript{4}  \and
        {Xingrui Wang}\textsuperscript{1} \and
        {Wenjun Zeng}\textsuperscript{3} \and 
        {Xinchao Wang}\textsuperscript{2} \and 
        {Zhibo Chen}\textsuperscript{1}
}


\institute{%
	\begin{itemize}
		\item[\textsuperscript{\Letter}] Xin Li, Xinchao Wang and  Zhibo Chen\\
		xin.li@ustc.edu.cn, xinchao@nus.edu.sg, chenzhibo@ustc.edu.cn
		\item[\textsuperscript{1}] University of Science and Technology of China, Hefei, China
		\item[\textsuperscript{2}] National University of Singapore, Singapore
		\item[\textsuperscript{3}] Eastern Institute for Advanced Study, Zhejiang, China
		\item[\textsuperscript{4}] Microsoft Research Asia, Beijing, China
	\end{itemize}
}

\date{Received: date / Accepted: date}

\maketitle

\begin{abstract}
Image restoration (IR) has been an indispensable and challenging task in the low-level vision field, which strives to improve the subjective quality of images distorted by various forms of degradation. Recently, the diffusion model has achieved significant advancements in the visual generation of AIGC, thereby raising an intuitive question, ``whether the diffusion model can boost image restoration". To answer this, some pioneering studies attempt to integrate diffusion models into the image restoration task, resulting in superior performances than previous GAN-based methods. Despite that, a comprehensive and enlightening survey on diffusion model-based image restoration remains scarce. In this paper, we are the first to present a comprehensive review of recent diffusion model-based methods on image restoration, encompassing the learning paradigm, conditional strategy, framework design,  modeling strategy, and evaluation. Concretely, we first introduce the background of the diffusion model briefly and then present two prevalent workflows that exploit diffusion models in image restoration. Subsequently, we classify and emphasize the innovative designs using diffusion models for both IR and blind/real-world IR, intending to inspire future development. To evaluate existing methods thoroughly, we summarize the commonly used dataset, implementation details, and evaluation metrics. Additionally, we present the objective comparison for open-sourced methods across three tasks, including image super-resolution, deblurring, and inpainting. Ultimately, informed by the limitations in existing works, we propose \tct{nine} potential and challenging directions for the future research of diffusion model-based IR, including sampling efficiency, model compression, distortion simulation and estimation, distortion invariant learning, and framework design. 
The repository is released at \url{https://github.com/lixinustc/Awesome-diffusion-model-for-image-processing/}

\keywords{Diffusion models \and Image restoration \and Image enhancement \and Image processing}
\end{abstract}

\section{Introduction}
\label{sec:introduction}
Image Restoration (IR) has been a long-term research topic in low-level vision tasks, which plays an irreplaceable role in improving the subjective quality of images. The popular IR tasks contain image super-resolution (SR) \citep{SR1_SRCNN-cnn,SR3_EDSR-cnn,SR4_SAN,wang2018esrgan,sr_transformer1,chen2023activating-HST,sr3_swinfir-transformer,gao2023ctcnet}, deblurring \citep{nah2017deep-deblur-old1-cnn,kupyn2018deblurgan-old2,nimisha2017blur-old3-cnn,kim2023mssnet-deblur-cnn,zamir2022restormer_deblur-transformer,tsai2022stripformer_deblur-transformer}, denoising \citep{zhang2017beyond-denoise2-cnn,valsesia2020deep-denoise4-cnn,fan2022sunet-denoise-new1-transformer,liu2022dnt-denoise-new2-transformer,tian2023multi-denoise-new3,zhang2023mm-new4}, inpainting \citep{xie2012image-inpainting-old1,pathak2016context-inpainting-old2,inpainting-new2,yu2023inpaint-new3,li2022mat-inpainting-new4} and compression artifacts removing \citep{dong2015compression-jpeg-old1-cnn,svoboda2016compression-jpeg-old2-cnn,ehrlich2020quantization-jpeg-old3-cnn,wang2022jpeg-new2,jiang2023multi-jpeg-new3}, etc.  To restore the distorted images, traditional IR methods treated the restoration as the signal processing and reduce the artifacts with hand-crafted algorithms from the spatial or frequency perspective \citep{dabov2007image-traditional1,traditional3,mairal2008learning-traditional-4,xu2010image-traditional-5,schmidt2014shrinkage-traditional-8}. With the development of deep learning, numerous IR works collected a series of datasets tailored for various IR tasks, \egno, DIV2K \citep{DIV2K}, LSDIR \citep{li2023lsdir}, Set5 \citep{set5}, and Set14 \citep{Set14} for SR, Rain800 \citep{Rain800}, Rain200 \citep{yang2017deep-rain100}, Raindrop \citep{rain-drop} and DID-MDN \citep{zhang2018density-DID-MDN} for draining, REDS \citep{Nah_2019_CVPR_Workshops_REDS}, and Gopro \citep{Gopro} for motion deblurring, etc. Leveraging these datasets,  the majority of recent works \citep{SR3_EDSR-cnn,nah2017deep-deblur-old1-cnn,valsesia2020deep-denoise4-cnn,dong2015compression-jpeg-old1-cnn,svoboda2016compression-jpeg-old2-cnn,zhang2017beyond-denoise2-cnn,sr_transformer1,swinIR-transformer,sr3_swinfir-transformer,wang2022uformer-transformer,zamir2022restormer_deblur-transformer} focused on improving the representation capability of IR networks for complicated degradation through well-designed backbones based on Convolutional neural networks (CNNs) \citep{krizhevsky2017imagenet-cnn1} or Transformer \citep{vaswani2017attention-transformer}. Although these works achieve great progress in the objective quality (\egno, PSNR, and SSIM), the restored images still suffer from unsatisfactory texture generation, hindering the application of these IR methods in real-world scenarios. 

Thanks to the development of generative models \citep{kingma2013auto-VAE,van2016pixel-autoregressive1,salimans2017pixelcnn++-autoregressive3,rezende2015variational-flow,papamakarios2017masked-flow2,GAN,karras2017progressive-GAN1}, especially the generative adversarial network (GAN) \citep{GAN}, some pioneering IR studies \citep{dosovitskiy2016generating-loss1,yu2016ultra-loss2,johnson2016perceptual-loss3,SRGAN,wang2018esrgan} pinpoint that previous pixel-wise loss, \egno, MSE loss, and L1 loss, are susceptible to the blurring textures, and introduce the adversarial loss from GAN to the optimization of IR network, for enhancing its texture generation capability. For instance, SRGAN \citep{SRGAN} and DeblurGAN \citep{kupyn2018deblurgan-old2} utilize a combination of pixel-wise loss and adversarial loss to achieve perception-oriented SR network and deblurring network, respectively. Following them, two primary directions to improve GAN-based IR arise by enhancing the generator (\ieno, the restoration network) \citep{SRGAN,zhang2019ranksrgan-gene1,wang2018esrgan,wang2021realesrgan,zhang2021designing-gene2} and discriminator \citep{schonfeld2020u-pixel-discri,isola2017image-patch-wise-discri,karras2020analyzing-image-wise-discri2}. In particular, ESRGAN \citep{wang2018esrgan} introduces the powerful RRDB \citep{wang2018esrgan} as a generator for GAN-based SR tasks. Three popular discriminators, including pixel-wise discriminator (U-Net shape) \citep{schonfeld2020u-pixel-discri}, patch-wise discriminator \citep{isola2017image-patch-wise-discri,wang2018fully-patch-wise-4}, and image-wise discriminator \citep{simonyan2014very-image-wise-discri,karras2020analyzing-image-wise-discri2} (\ieno, VGG-like architecture) are designed to focus on the subjective quality at different levels of granularity, (\ieno, from local to global). Although the above progress, most researches on GAN-based IR still face two inevitable but crucial problems: 1) the training of GAN-based IR is susceptible to mode corruption and unstable optimization
and 2) the textures of most generated images seem to be fake and counterfactual.

Recently, diffusion models have emerged as a new branch of generative models, leading to a series of breakthroughs in visual generation tasks. The prototype of the diffusion model can be traced back to the work \citep{sohl2015deep-dpm}, and has been developed by DDPM \citep{DDPM}, NCSN \citep{NCSN} and SDE \citep{SDE}. In general, the diffusion model is composed of the forward/diffusion process and reverse process, where the forward process progressively increases the pixel-wise noise to the image until it satisfies the distribution of Gaussian noise, and the reverse process aims to reconstruct the image by denoising with score estimating \citep{NCSN} or noise prediction \citep{DDPM}. Compared to GANs, the diffusion models yield high-fidelity and diverse generation results, thereby successfully replacing the GANs in a series of fields, such as visual generation \citep{NCSN,DDIM,SDE,DDPM} and conditional visual generation \citep{singh2022conditioning-noisecondition,popov2021grad-app1,babnik2023diffiqa-app11,blattmann2023align-app12}. With the advancement of the vision-language model, diffusion model has been extended to the cross-modality generation, such as StableDiffusion \citep{LDM}, and DALLE-2 \citep{ramesh2022hierarchical-dalle2}. This greatly promotes the development of artificial intelligence generated content (AIGC). We have listed the representative works on diffusion models based on the timeline in Fig. 1 of the \textbf{Supplementary}.

Inspired by the superior generative capability of diffusion models, numerous studies have investigated their application in image restoration tasks, targeting to facilitate the texture recovery. According to the training strategy, these works can be roughly divided into two categories: 1) the first category \citep{SR3,li2022srdiff,sde-sr,weather-diff,deblur-DPM,luo2023refusion,zhou2023pyramid-pydiff,chan2023sud,miao2023dds2m,varanka2024pfstorer,qu2024xpsr,zhang2023diffusion-difftsr} is dedicated to optimizing the diffusion model for IR from scratch via supervised learning, and 2) the second one (\ieno, zero-shot one) \citep{kawar2021snips,jpegddrm,DDRMDRM,ddnm,dps,SIM-SGM,liu2023improved,abu2022adir} strives to exploit the generative priors in the pre-trained diffusion models for IR. Typically, supervised learning-based methods necessitate the collection of large-scale distorted/clean image pairs, while zero-shot-based methods predominantly rely on known degradation modes. These limitations impede the application of these diffusion model-based methods in real-world scenarios, where the distortions are typically diverse and unknown. To further tackle the above issue, some researches \citep{SR3+,murata2023gibbsddrm,blinddps,wang2023exploiting-stable-sr,wei2023raindiffusion,wang2023dr2} have extended the diffusion model to handle blind/real-world image restoration by incorporating real-world distortion simulation, kernel estimation, domain translation, and distortion invariant learning.

Although the diffusion models have shown significant efficacy in image restoration, the associated techniques and benchmarks exhibit considerable diversity and complexity, making them hard to be followed and improved. Moreover, the absence of a comprehensive review for diffusion model-based IR further limits its development. In this paper, we are the first to review and summarize the works on diffusion model-based image restoration methods, aiming to provide a well-structured and in-depth knowledge base and facilitate its evolution within the image restoration community. \tct{More analysis on the advantages of our survey can be found in the Section E of our \textbf{Supplementary}.} 

In this survey, we start by presenting the background of diffusion models in Sec.~\ref{sec:background}, highlighting two foundational modeling methods, \ieno, DDPM \citep{DDPM}, and SDE \citep{SDE}. 
Based on these preliminaries, we shed light on the advances of diffusion models in image restoration from two distinct directions in Sec.~\ref{sec:Diffuion-based IR}: 1) supervised diffusion model-based IR, and 2) zero-shot diffusion model-based IR. In Sec.~\ref{sec:blindir}, we summarize the diffusion model-based IR under more practical and challenging scenarios, \ieno, blind/real-world degradation. This intends to further enhance the capability of the diffusion model-based IR methods in fulfilling the demands of practical application. To facilitate a reasonable and exhaustive comparison, in Sec.~\ref{sec:experiments}, 
a series of comparisons between different DM-based benchmarks across four tasks are provided, including image SR, inpainting, and deblurring, and real-world restoration. In Sec.~\ref{sec:future}, we delve into analyzing the primary challenges and potential directions in diffusion model-based IR to provide some insights and hope to inspire more new works. The final conclusion for this review is summarized in Sec.~\ref{sec:conclusion}.

\section{Background on Diffusion Model (DM)}
\label{sec:background}
Diffusion probabilistic model (\ieno, diffusion model) has brought an evolution in the field of generative models, which transforms the complicated and unstable generation process into several independent and stable reverse processes via Markov Chain modeling. There are three foundational diffusion models that are widely utilized, including DDPM \citep{DDPM}. NCSNs \citep{NCSN} and SDE \citep{SDE}. Among them, NCSNs \citep{NCSN} seeks to model the data distribution by sampling using annealed Lange-vin dynamics with a sequence of decreasing noise scales. In contrast, DDPM \citep{DDPM} models the forward process with a fixed process of adding Gaussian noise, which simplifies the reverse process of the diffusion model into a solution process for the variational bound objective. 
These two basic diffusion models are actually special cases of score-based generative models \citep{SDE}. SDE \citep{SDE}, as the unified form, models the continuous diffusion and reverse processes with stochastic differential equation (SDE). It proves that the NCSNs and DDPM are only two separate discretizations of SDE. We will clarify two commonly-used modeling strategies of diffusion models in image restoration, \ieno, DDPM~\citep{DDPM} and SDE~\citep{SDE} in the following subsections. We also illustrate the NCSN~\citep{NCSN} in the section B.1 of the Supplementary.

\subsection{Denoising Diffusion Probabilistic Model}
DDPM (Denosing diffusion probabilistic model) \citep{DDPM} originates from the diffusion models \citep{sohl2015deepDiffusionModels}, which introduces the simple variational bound objective for diffusion models by setting the variance $\beta_t$ as fixed values. There are two crucial processes in diffusion models, \ieno, the forward process and reverse process.  
In particular, the forward process (\ieno, the diffusion process in DDPM) aims to progressively corrupt the training data to the Gaussian Noise, which is a parameterized Markov chain as:
\begin{equation}
    q(x_{t}|x_{t-1})=\mathcal{N}(x_{t};\sqrt{1-\beta_{t}} \cdot x_{t-1},\beta_{t}\mathbf{I}), 
    \label{eq:diffusion_process}
\end{equation}
where $x_{0},x_{1},...,x_{T}$ are the noise latent variables by adding noises to the training data point $x_0 \sim p_{data}(x)$ progressively with noise schedule as $\beta_{1},...,\beta_{T}\in(0,1)$ for $T$ steps. 
And we can compute the probabilistic distribution of $x_t$ given $x_0$ as:
\begin{equation}
    q(x_{t}|x_{0})=\mathcal{N}(x_{t};\sqrt{\hat{\alpha}_{t}} x_{0},\sqrt{1-\hat{\alpha}_{t}}\mathbf{I}), 
\end{equation}
where $\alpha_{t}=1-\beta_{t}$ and $\hat{\alpha}_{t}=\prod \limits_{i=1}^t \alpha_{i}$. 
When time step $t \to T$ is large enough, the distribution of $x_{T}$ will be a standard Gaussian distribution $\pi(x_{T})\sim\mathcal{N}(0, \mathbf{I})$ since $\hat{\alpha}_{t} \to 0$. 

\begin{figure}[htp]
	\centering
	\includegraphics[width=1\linewidth]{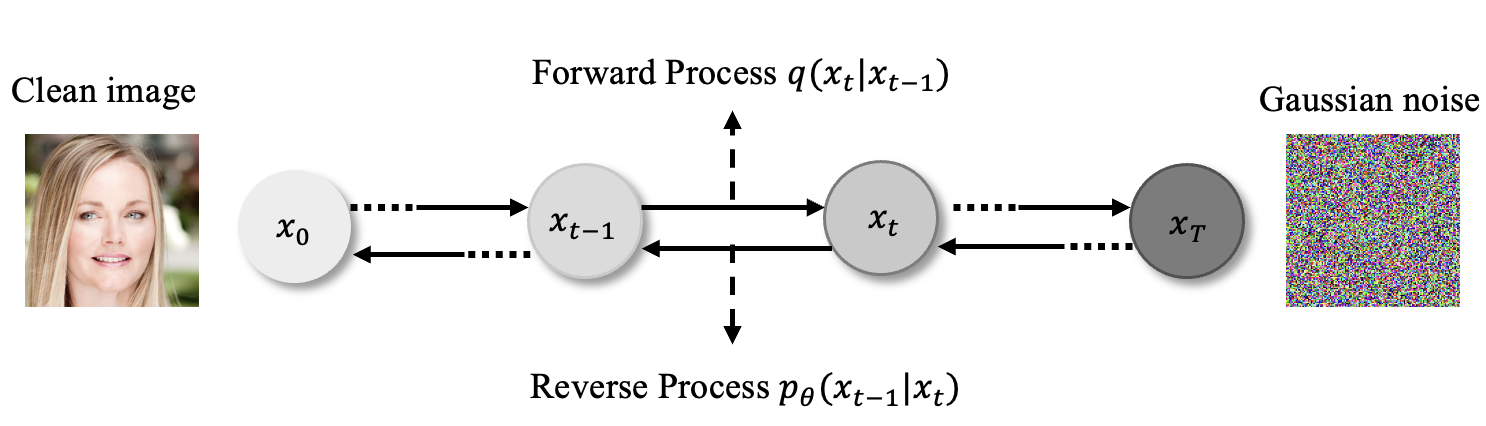}
	\caption{Denoising Diffusion Probabilistic Models.}
	\label{fig:DDPM}
\end{figure}

The reverse process of diffusion models aims to recover the data distribution from the Gaussian noises by approximating the posterior distribution $q(x_{t-1}|x_{t}, x_0)$ as:
\begin{equation}
    q(x_{t-1}|x_{t}, x_0) = \mathcal{N}(x_{t-1}; \Tilde{\mu}_{t}(x_{t},x_0),\Tilde{\beta}_t\mathbf{I}), 
    \label{eq:reverse}
\end{equation}
where $\Tilde{\mu}_{t}(x_{t},x_0)=\frac{\sqrt{\hat{\alpha}_{t-1}}\beta_t}{1-\hat{\alpha}_{t}}x_0 + \frac{\sqrt{\hat{\alpha}_{t}}(1-\hat{\alpha}_{t-1})}{1-\hat{\alpha}_{t}}x_t=\frac{1}{\sqrt{\alpha_t}}(x_t\\-\frac{\beta_t}{\sqrt{1-\hat{\alpha}_t}})\epsilon$ for $\epsilon \sim \mathcal{N}(\mathbf{0}, \textbf{I})$  and $\tilde{\beta_t}=\frac{1-\hat{\alpha}_{t-1}}{1-\hat{\alpha}_t}$. As stated in Eq.~\ref{eq:reverse}, the variance schedule $\beta_t$ is predefined, and thus, it only requires approximating the mean $\mu_\theta (x_t, t) = \Tilde{\mu}_t(x_t, x_o)$ by a denoising network $\epsilon_\theta (x_t, t)$. The optimization objective \citep{DDPM} for denoising network can be written as:
\begin{equation}
    \mathcal{L}_{simple} = \mathbb{E}_{t, x_0, \epsilon}[||\epsilon-\epsilon_\theta(\sqrt{\hat{\alpha}_t}x_0+\epsilon\sqrt{1-\hat{\alpha}_t}, t)||_2^2]
\end{equation}
The direct illustration of denoising diffusion probailisitic models(DDPM) is shown in Fig.~\ref{fig:DDPM}

\subsection{Stochastic Differential Equations (SDEs)}
To unify the approaches with the score-based generative modeling and diffusion probabilistic modeling, SDEs \citep{SDE} exploit the continuous diffusion process through the stochastic differentiable equation (SDE) as:
\begin{equation}
    dx=\mathbf{f}(x,t)dt+g(t)d\mathbf{w}
\end{equation}
where $\mathbf{w}$ is the standard Wiener process, $\mathbf{f}(\cdot,t)$ is called drift coefficient of $x(t)$ and $g(\cdot)$ is called the diffusion coefficient of $x(t)$. Here, the diffusion coefficient can be understood as the degree perturbed by random noise, and the drift coefficients can be designed to ensure the Gaussian distribution, such as DDPM \citep{DDPM} and NCSN \citep{NCSN}. The reverse process of the above continuous diffusion process (\ieno, sampling data from noise) is also a diffusion process and can be modeled by a reverse-time SDE:
\begin{equation}
    dx=[\mathbf{f}(x,t)-g(t)^2\nabla_{x}\log p_{t}(x)]dt+g(t)d\hat{\mathbf{w}}
    \label{eq:reverse_SDE}
\end{equation}
Here $dt$ is an infinitesimal negative time step and $\hat{\mathbf{w}}$ is the standard Wiener process when time flows backward from $T$ to $0$. The core of reverse-time SDE is to estimate the score function using a neural network and then Eq.~\ref{eq:reverse_SDE} can be solved with the score matching \citep{hyvarinen2007some-score-matching,hyvarinen2005estimation-score-matching}. 

\begin{figure*}[h]
	\centering	\includegraphics[width=0.95\linewidth]{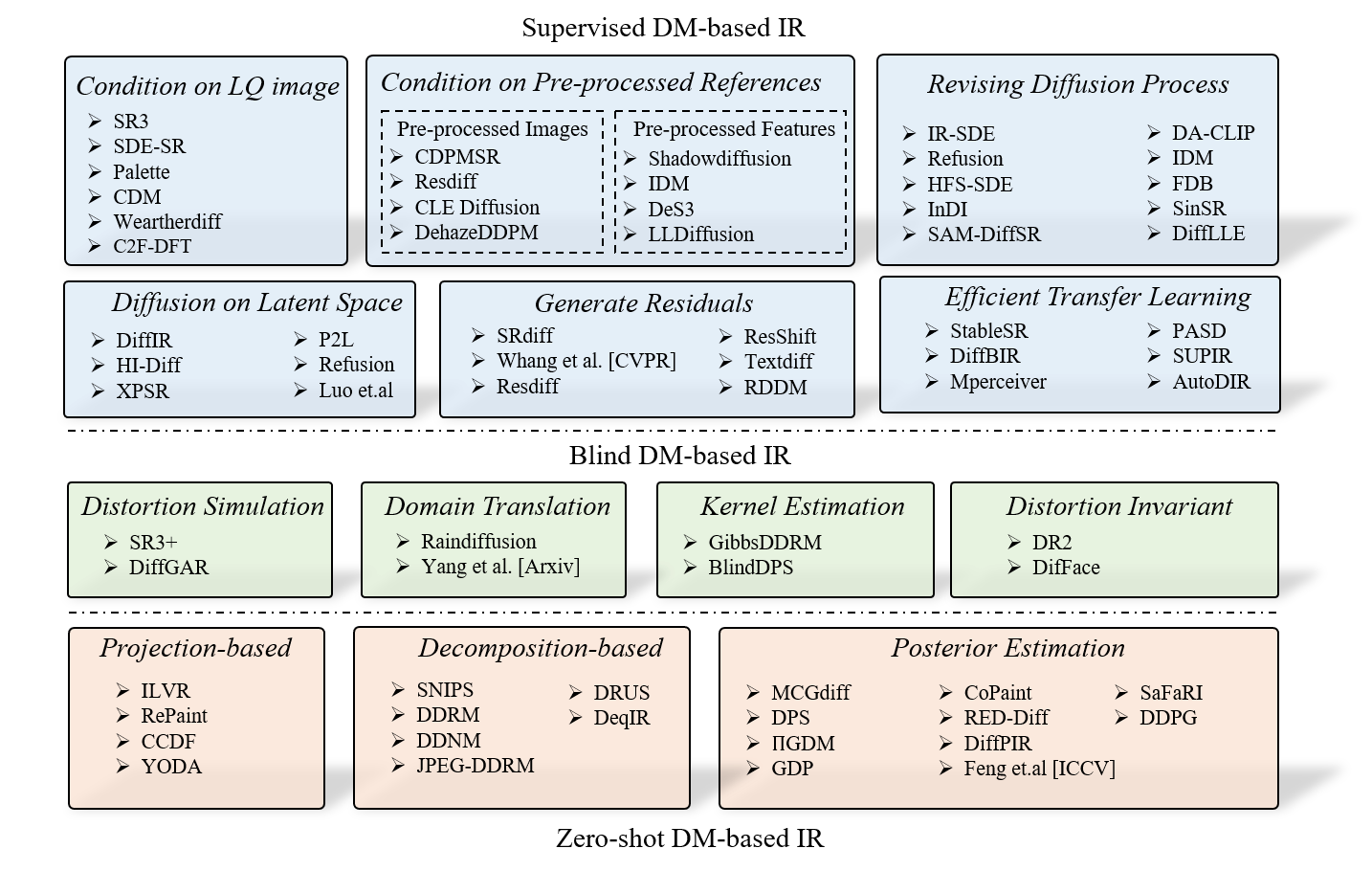}
	\caption{The overview of diffusion model-based image restoration models. This figure categorizes diffusion models into two types based on their training methods, namely the supervised-based models (indicated with a blue background) and zero-shot-based models (indicated with an orange background). Additionally, the figure provides a more detailed classification of models within these two categories according to how conditions are incorporated. The models with a green background are specifically designed for real-world image restoration. All works in this figure can be found in Section C,
 Table 1 and Table 2 of the \textbf{Supplementary}).}
	\label{fig:Overview}
\end{figure*}

DDPM and NCSN can be regarded as discretizations of two different SDEs. When the time variable turns to infinity, the forward process of DDPM will converge to the following: 
\begin{equation}
    dx=-\frac{1}{2}(1-\hat{\alpha}_{t})\mathbf{x}dt+\sqrt{1-\hat{\alpha}_{t}}d\mathbf{w}.
    \label{eq:ddpn_sde}
\end{equation}
And the SDE form of NCSN is as follows:
\begin{equation}
    dx=\sqrt{\frac{d[\sigma^2(t)]}{dt}}d\mathbf{w}. 
    \label{eq:ncsn_sde}
\end{equation}
Here, Eq.~\ref{eq:ddpn_sde} and Eq.~\ref{eq:ncsn_sde} are called the Variance Preserving (VP) SDE and the Variance Exploding (VE) SDE, respectively. More background information about the diffusion model is shown in section B of the \textbf{Supplementary}.

\section{Diffusion model-based Image Restoration Methods}
\label{sec:Diffuion-based IR}
According to whether the diffusion models (DMs) are training-free for IR, we can preliminarily classify the DM-based IR methods into two categories, \ieno, supervised DM-based methods \citep{SR3,deblur-DPM,jin2022shadowdiffusion-driven,qiu2023diffbfr,wang2023exploiting-stable-sr,zhou2023pyramid-pydiff,jiang2023low-WCDM,chen2023hierarchical-HIdiff,chan2023sud,yi2023diff-retinex,gou2023exploiting-zeroair,xu2023stage,wang2023exposurediffusion,wang2024enhancing,zhang2023ultrasound-drus}, and zero-shot DM-based methods \citep{choi2021ilvr,DDRMDRM,dps,GDP,iigdm,zhu2023denoising-diffpir,mardani2023variational-red-diff,fabian2023diracdiffusion,SIM-SGM}. Particularly, the supervised DM-based IR methods entail training the diffusion model from scratch with paired distorted/clean images of IR datasets. Unlike previous GAN-based methods \citep{yuan2018unsupervised-ganr1,dinh2022hyperinverter-ganr6,wang2020transformation-ganr2} that directly take distorted images as input, DM-based IR employ the well-designed conditional mechanism to incorporate the distorted images as guidance during the reverse process. Despite its promising texture generation results, this approach encounters two notable limitations: 1) training the diffusion model from scratch relies on a large quantity of paired training data; 2) collecting paired distorted/clean images in the real world is challenging. In contrast, zero-shot DM-based methods offer an appealing alternative, requiring only distorted images and dispensing with the need for retraining diffusion models. Instead of acquiring the restoration capability from the training datasets of IR, it excavates and exploits the structure and texture priors from the pre-trained diffusion models for image restoration. The core idea stems from the intuition that the pre-trained generative models can be viewed as the structure and texture repository, constructed using real-world datasets with large amounts of data, such as ImageNet \citep{russakovsky2015imagenet}  and FFHQ \citep{FFHQ}. Consequently, an essential challenge faced by zero-shot DM-based IR methods is: \textit{how to extract the corresponding perceptual priors while preserving the data structure from distorted images.} In the subsequent subsections, we first briefly review the representative supervised DM-based IR method: SR3 \citep{SR3}, and zero-shot DM-based IR method: ILVR \citep{choi2021ilvr}. Then we further classify these two types of methods from the perspectives of conditional strategy,  diffusion modeling, and framework, which are summarized in Table 1 and Table 2 of the \textbf{Supplementary}, respectively. In addition, the overall taxonomy of diffusion models is illustrated in Fig.~\ref{fig:Overview}

\subsection{SR3 -- Representative Supervised DM for IR}
\label{sec:sr3}
Unlike the pure image generation task that synthesizes images from noise, image restoration seeks to generate high-quality images from the corresponding low-quality images. Hence, the pivotal challenge of supervised DM-based IR lies in \textit{how to effectively incorporate the degraded/low-quality image into the diffusion model as the condition}. Let us denote the degraded image as $y$. The foundational objective of the diffusion models (DMs) for IR is to learn the posterior distribution $p_{\theta}(x_{t-1}|y,x_{t})$ at time step $t$, such that $x_{0}\sim q(x|y)$ and $x$ denotes the corresponding high-quality image. To achieve this, a pioneering supervised DM-based approach, SR3, is introduced with a straightforward condition strategy. Specifically, it   
 directly concatenates the degraded image with the generated image $x_t$ at $t$ time step, effectively enabling the conditional image generation for SR.

\begin{figure}[h]
	\centering
\includegraphics[width=0.8\linewidth]{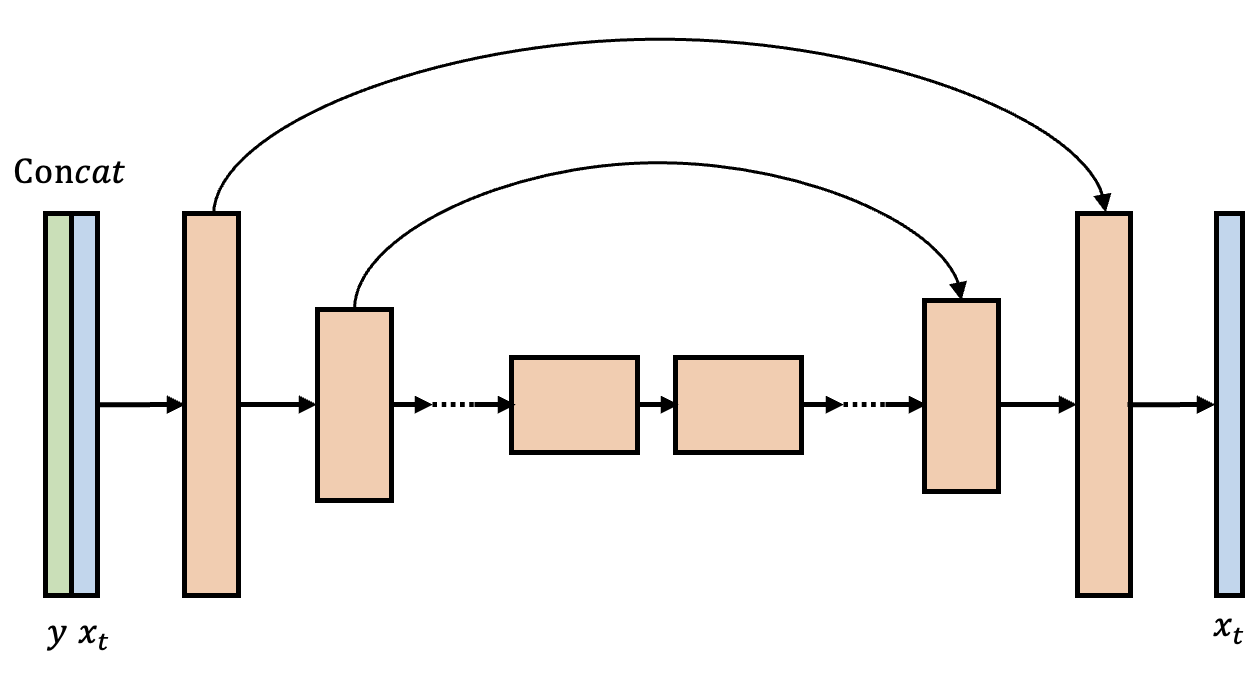}
	\caption{The Backbone of SR3 network}
	\label{fig:SR3}
\end{figure}
As depicted in Fig.~\ref{fig:SR3}, SR3 follows the typical DDPM \citep{DDPM} framework and utilizes the U-Net model as the noise predictor. Given the low-resolution (LR) image $y$, SR3 initially up-samples it to the desired resolution using bicubic interpolation. Subsequently, it concatenates the super-resolved LR image $y$ with the denoised output $x_t$ at $t$ time step, serving as the input of the diffusion model to predict the noise for the $t-1$ step. When reaching $t=0$, the diffusion model can deliver an upsampled high-quality image $x_0$ of $y$ as $x_0 \approx x$.

\begin{figure}[h]
	\centering
 \includegraphics[width=1\linewidth]{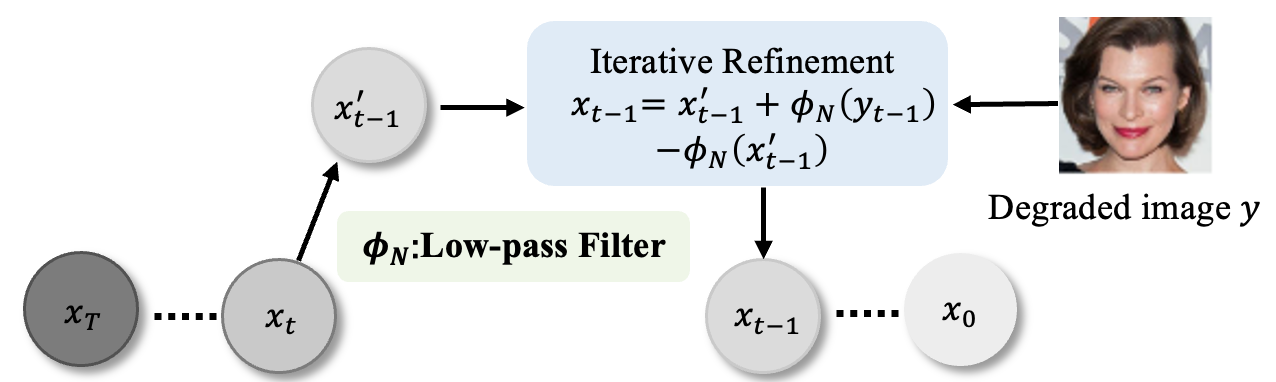}
	\caption{The Backbone of ILVR network}
	\label{fig:ILVR}
\end{figure}

\subsection{ILVR -- Representative Zero-shot DM for IR}

Although the supervised DM-based IR methods have exhibited remarkable performance, the training process entails substantial computational cost and large-scale paired datasets, which might be prohibitive for some researchers. To circumvent this, zero-shot DM-based IR  \citep{choi2021ilvr,repaint,kawar2021snips,CCDF,zhang2022unsupervised-pretrained} is proposed to exploit the intrinsic knowledge within the pre-trained diffusion models. Concretely, it is observed that pre-trained diffusion models for image generation, trained with a large number of natural images, encapsulate considerable prior knowledge about rich textures. And thus, these pre-trained diffusion models can be considered as repositories of texture information. The exploration of reusing such prior knowledge for training-free image restoration is an emerging and promising direction in the low-level vision field.

As the initial work, Choi \etal \citep{choi2021ilvr} introduces the iterative latent variable refinement (\ieno, ILVR) method, which leverages the unconditional diffusion model to enable the training-free conditional generation for image SR and image translation. The pivotal innovation of ILVR involves substituting the low-frequency components in denoised output with their counterpart from the reference image. This substitution process, illustrated in Fig.~\ref{fig:ILVR}, ensures structure and semantic consistency between the generated and reference images, thereby facilitating the conditional generation.  
Particularly, given a reference image $y$ (\egno, the distorted image in IR), at time step $t$, the ILVR  predicts the denoised result for time $t-1$ using the following formula:
 \begin{equation}
    \mathbf{x'}_{t-1}=\sigma_{t}\mathbf{z}+\frac{1}{\sqrt{\alpha_{t}}}(\mathbf{x}_{t}-\frac{1-\alpha_{t}}{\sqrt{1-\hat{\alpha}_{t}}}\mathbf{\epsilon_{\theta}(\mathbf{x}_{t},t)})
    \label{eq:ilvr}
\end{equation}
where $\epsilon_\theta(x_t, t)$ represents the noise predicted by the denoising network, and $z$ denotes the standard random Gaussian noise that governs the randomness of the generated image. 
However, this sampling process inevitably produces inconsistent structures/textures in $x_t$, which necessitates a refinement to align with the structures/textures in reference image $y$ by low-frequency substitution:
\begin{equation}
    x_{t-1}=x'_{t-1}+\Phi_{N}(y_{t-1})-\Phi_{N}(x'_{t-1}).
\end{equation}
Here, $\Phi_{N}$ denotes the low-pass filter designed to bypass the low-frequency component from input, and $y_{t-1}\sim q(y_{t-1}|y)$ is the diffused state of $y$ at $t-1$ steps. Following ILVR, most zero-shot DM-based IR methods \citep{dps,iigdm,DDRMDRM,ddnm,fabian2023diracdiffusion,feng2023score,copaint} predominantly focus on enhancing the refinement strategies within the sampling process, thereby training-free. 
 
\renewcommand\arraystretch{1.5}

\subsection{Supervised DM-based IR.}
\label{sec:Improving diffusion model based Restoration Models}
Motivated by SR3 \citep{SR3}, numerous studies have endeavored to optimize the supervised DM-based IR framework, focusing on enhancing the condition strategy and exploring potential and more efficient generation spaces. With respect to the condition strategy, we categorize these studies into three types based on the conditions: 1) low-quality reference image, 2) pre-processed references, and 3) revising diffusion process. Regarding generation space, the supervised DM-based IR methods can be classified into three distinct groups: image space, residual space, and latent space. If not mentioned, the majority of studies generate the restored images within image space, where the structures and textures are required to be generated directly. In contrast, the residual-space diffusion model focuses on reconstructing the residuals between the low-quality image and its corresponding high-quality image, which simplifies the complexity of generating the whole image. Latent-based methods utilize a well-designed encoder to transform the image into a compact latent space for generation, thereby improving the generation efficiency. This section will elucidate existing studies on supervised DM-based IR in terms of the above three conditional strategies and the last two generation spaces.

\subsubsection{Condition with Low-quality Reference Image}
As highlighted in Sec.~\ref{sec:sr3}, incorporating distorted images as conditions is indispensable and crucial for supervised DM-based IR \citep{chen2023image-PromptSR,wang2023learning-C2F-DFT,li2024blinddiff}. The SR3 method has shown that substantial performance can be achieved through a simple concatenation operation. 
It utilizes a direct concatenation of low-quality reference images with the denoised result at the $t-1$ step as the condition for noise prediction at the $t$ step. With the same condition strategy, Saharial \etal \citep{saharia2022palette} propose a unified diffusion model, termed Palette, for image-to-image translation tasks, which achieves excellent performance on image colorization, inpainting, uncropping, and JPEG artifact removal. Furthermore, they investigate the effects of different optimization objectives for sample diversity and highlight the pivotal role of self-attention within U-Net for the diffusion model. Despite their effectiveness, the aforementioned methods are constrained since they only support a fixed resolution for IR once trained. To adapt the diffusion model for the real-world IR with an arbitrary size, Özdenizci \etal \citep{weather-diff} divide the degraded image and corresponding sampling results $x_t$ 
 into several overlapped patches, and then utilize the patch-wise concatenation as the inputs of the diffusion model for noise prediction. Additionally, to address the inconsistency problem caused by different sampling patches in the overlapped region, this work introduces the mean estimated noise for each pixel within the overlapped region. To enhance the quality of generated images, Ho \etal \citep{CDM} introduce the three-layer cascaded diffusion model based on the SR3 backbone.  The first diffusion model is exploited to achieve the class-conditioned low-resolution image generation, and two additional diffusion models are cascaded to super-resolve the low-resolution generated image, resulting in a higher-resolution and more realistic generated image. In addition to the DDPM-based approaches, there is another work \citep{sde-sr} that explores the various variants of the continuous diffusion model SDE \citep{SDE} for the face super-resolution using predictor-corrector sampling.

\begin{figure}[htp]
	\centering
	\includegraphics[width=0.85\linewidth]{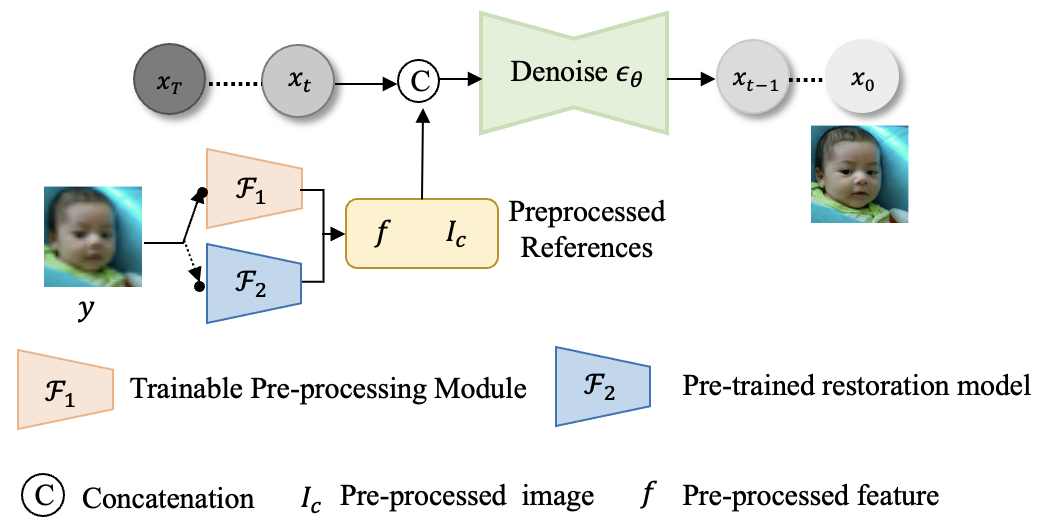}
	\caption{The flowchart of DM-based IR methods using pre-processed references, where the low-quality image is first processed by a pre-trained restoration network or trainable preprocessing module. The output references could be features or clean images}
	\label{fig:preprocessing}
\end{figure}
\subsubsection{Condition with Pre-processed References}
Even though direct concatenation with the low-quality image shows promising performance, the artifacts in low-quality images inevitably cause harmful effects on the generation of the diffusion model, especially for severe and diverse distortions. To mitigate this issue, 
several studies \citep{yin2023cle-CLE-Diffusion,wang2023lldiffusion-LLDiffuion,shang2023resdiff,niu2023cdpmsr,intro-densely-13,moser2023waving,guo2023boundary-BCDiff,wang2024promptrr,wu2023hsr-HSR-diff} strive to enhance the condition by preprocessing low-quality images with either jointly trained modules or pre-trained restoration networks. As depicted in Fig.~\ref{fig:preprocessing}, these works can be grouped into two categories based on the preprocessing strategies, \ieno, condition with pre-processed reference image and feature.

\noindent\textbf{Pre-processed reference image.}
To alleviate the side effects of artifacts in low-quality images, CDPMSR \citep{niu2023cdpmsr} exploits existing super-resolution models, \egno, RCAN \citep{zhang2018image-RCAN}, SwinIR \citep{swinIR-transformer}, EDSR \citep{EDSR-9}, to enhance the low-quality image, thereby providing a high-quality and more reliable condition for diffusion model. Additionally, it eschews the stochastic sampling during the reverse process in favor of a deterministic denoising process, yielding superior image quality and faster inference. Notably, the pre-processing reference image can serve as not only an enhanced condition but also an initially well-restored image. Therefore, ResDiff \citep{shang2023resdiff} utilizes a pre-trained CNN to generate a low-frequency content-abundant image as the initially restored image and exploits the conditional diffusion model to further generate the residuals between the pre-processed distorted image and its corresponding clean image. In contrast, DehazeDDPM \citep{yu2023high-HazeDDPM} tries to add a physics-aware image as an auxiliary condition to achieve complex dehazing task. Thus they generate the transmission map, and haze-free image $I$ through physical modeling. Then, these preprocessed images are then incorporated into diffusion through dynamic fusion.

\noindent\textbf{Pre-processed reference feature.}
Another popular line is to use the feature of the reference image as the condition for the diffusion model \citep{varanka2024pfstorer,guo2023shadowdiffusion,IDM,jin2022des3,xiao2023ediffsr,wang2023lldiffusion-LLDiffuion,liu2024diff-plugin}.  
IDM \citep{IDM}, striving for continuous image super-resolution, first extracts the initial features of the low-resolution image with EDSR \citep{EDSR-9}. Then, the initial features are downsampled to multiple scales, which are utilized as the conditions for different upsampling layers in the diffusion model, with the intent of refining the implicit representation. In contrast, ShadowDiffusion \citep{guo2023shadowdiffusion} leverages a pre-trained transformer backbone to extract the degradation prior (\ieno, the degradation-related features) from the distorted reference image. This extracted degradation prior is exploited as the auxiliary to refine the generated shadow mask and serves as the condition for shadow-free image generation.

\begin{figure}[h]
	\centering
	\includegraphics[width=1\linewidth]{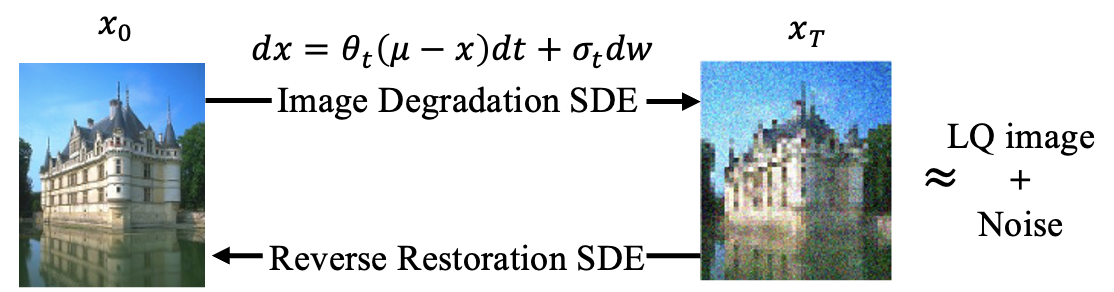}
	\caption{The architecture of IR-SDE \citep{IR-SDE}, where the forward diffusion process approximately models the image degradation processes. The final output of the forward process is equal to a low-quality image with random noise.}
	\label{fig:IR-SDE}
\end{figure}
\subsubsection{Condition by Revising Diffusion Process}
Notably, the above-mentioned supervised DM-based IR methods introduce the conditions by modifying the network while preserving the diffusion process from DDPM \citep{DDPM}. However, this requires the generation process to start from the noise. To obviate this, some studies \citep{IR-SDE,luo2023refusion,luo2023controlling-DACLIP,yang2023difflle-DiffLLE,zhao2023towards-IDM,wang2023sinsr,wang2024sam-diffsr,HFS-SDE,INDI,xiao2023ediffsr,mirza2023learning-fourier-frequency-FDB} condition the diffusion model by modifying the diffusion process, such that the diffused output $x_T$ (\ieno, the start point during the reverse process) approximates the low-quality image corrupted with few Gaussian noises.  
As shown in Fig.~\ref{fig:IR-SDE}, Luo \etal \citep{IR-SDE} modify the forward process with mean-reverting SDE to model the unified degradation process of IR as:
\begin{equation}
    dx = \theta_t (\mu - x)dt + \sigma_t dw,
    \label{eq:ir-sde}
\end{equation}
where $\theta_t$ and $\sigma_t$ are the time-dependent parameters. $\mu$ denotes the distorted image and $x$ represents its corresponding high-quality image. With the mean-reverting SDE, this work successfully models the image degradation and restoration processes with the modified forward and reverse processes, respectively. This avoids the generation from pure noise and achieves better resto-ration performance. Building on IR-SDE \citep{IR-SDE}, the same team introduces the Refusion \citep{luo2023refusion}, which further refines IR-SDE \citep{IR-SDE} by optimizing aspects such as network architecture, noise level, and denoising steps, etc. To reduce the computation cost, Refusion introduces the U-Net compression strategy, thereby enabling efficient sampling in the latent space. 

With the same purpose,  Xie et al \citep{xie2023diffusion} redefine the diffusion process such that the sampling starts from the noisy image. Taking into account the diversity of noise, they derive three separate diffusion processes for the removal of Gaussian, Gamma, and Poisson noises, respectively. In contrast, InDI \citep{INDI} introduces the continuous diffusion process as: 
\begin{equation}
    x_t = (1-t) x + ty + t\epsilon n, t \in [0, 1],
\end{equation}

where $x$ and $y$ are the high-quality images and their corresponding low-quality counterparts. $\epsilon$ is a small constant. It can be interpreted as the step-wise interpolation of high-quality and low-quality images at time step $t$, which decomposes the original single-step prediction of supervised image restoration into several small steps, effectively circumventing the regression-to-the-mean effects often found in conventional supervised image restoration. Different from the diffusion process in the spatial domain, HFS-SDE \citep{HFS-SDE} reformulate the diffusion process for magnetic resonance (MR) reconstruction at frequency space. In this approach, the forward process progressively adds noise into high-frequency space, resulting in the final $x_T$ that is composed of high-frequency noise and low-frequency data. During the reverse process, HFS-SDE employs Predictor-Corrector (PC) method \citep{SDE} for sampling. 
 
\begin{figure}[h]
	\centering
 \includegraphics[width=1.0\linewidth]{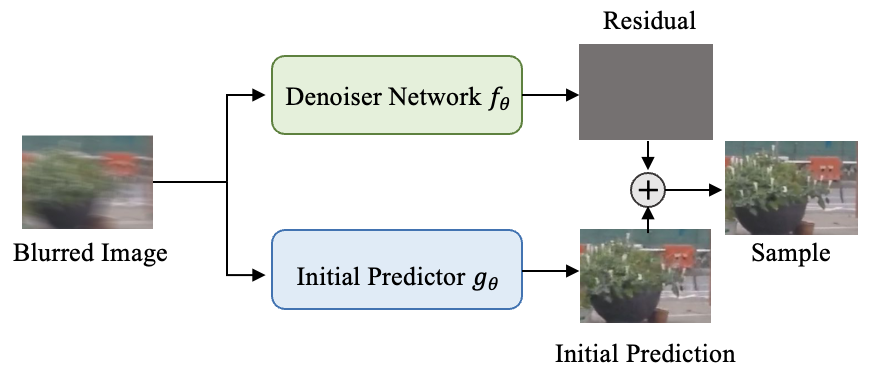}
	\caption{The architecture of Predict and Refine model \citep{deblur-DPM}. The predictor $g_{\theta}$ generates the initial prediction of a clean image, while residual information is modeled by diffusion process.}
	\label{fig:whangdeblur}
\end{figure}

\subsubsection{Generate Residuals}
In the supervised DM-based IR, most studies directly generate high-quality images from noise, which necessitates the concurrent generation of structure and textures. However, regenerating the structures/textures already existing in low-quality images unnecessarily burdens the diffusion model and increases extra resource costs. Motivated by this, some representative studies \citep{li2022srdiff,deblur-DPM,shang2023resdiff,liu2023residual-RDDM,yue2023resshift,gao2022cocodiff-residual2,karchev2022strong-residual1,liu2023textdiff} seek to move the generation process from the image space to the residual space. The objective is to generate the residuals between paired high-quality and low-quality images. As the pioneering work, SRDiff \citep{li2022srdiff} is the first to utilize the diffusion model to predict the residual in SR. Whang \citep{deblur-DPM} introduces a predict-and-refine strategy for image deblurring tasks. As shown in Fig.~\ref{fig:whangdeblur}, this work first predicts an initial deblurring image with the deterministic deblurring network and then generates the residual through a stochastic diffusion model.  ResDiff \citep{shang2023resdiff} mentioned above also adopts this strategy for SR. Resshift \citep{yue2023resshift} introduces a novel Markov chain that shifts residuals between high-quality and low-quality images. This residual shifting technique achieves comparable performance even with a reduced number of sampling steps, as few as 15. Another work RDDM \citep{liu2023residual-RDDM} redefines the forward diffusion process and incorporates residuals to represent directional diffusion from the target domain to the input domain. RDDM predicts both noise and residuals during training to effectively restore images.

\subsubsection{Diffusion on Latent Space}
\begin{figure}[h]
	\centering
	\includegraphics[width=1\linewidth]{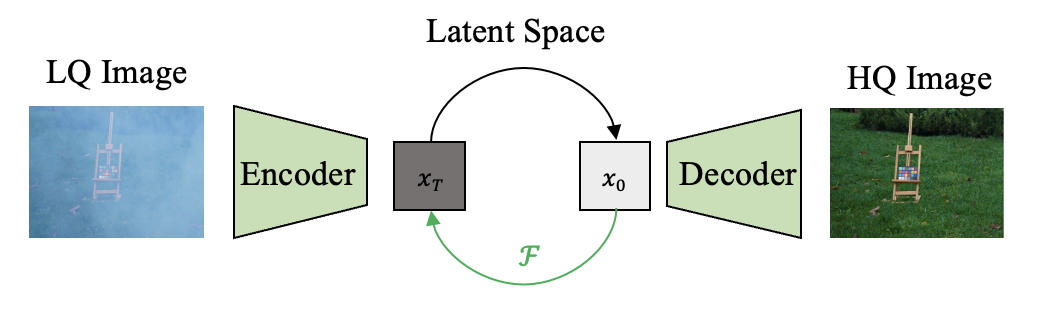}
	\caption{An overview of diffusion model used in Refusion \citep{luo2023refusion}, the low-quality image is first compressed into latent by an encoder. Then the latent information is modeled by a diffusion process. Here, $F$ denotes the forward process, which progressively adds noise to $x_0$, transforming it into $x_T$.}
	\label{fig:refusion}
\end{figure}

To alleviate the training and sampling costs of the diffusion model, some works \citep{LDM,luo2023refusion,xia2023diffir,yang2023pixel-aware,chung2023prompt-P2L,qu2024xpsr,luo2023image-frequency-feng,sun2023improving-CCSR}, try to conduct diffuion in latent space. Among them, stableDiffusion \citep{LDM} is the first work to implement the DM-based generation in latent space. Particularly, it pre-trains an autoencoding model (\ieno, an encoder-decoder architecture) to learn the perceptual latent space, which is able to preserve the perceptual quality of the reconstructed image while reducing the computational complexity. Utilizing the pre-trained autoencoder, StableDiffusion transforms the image-wise diffusion process to the latent space, and then introduces various conditions (\egno, text, segmentation map, and image) into the diffusion model with the cross-attention mechanism. Inspired by this, Refusion \citep{luo2023refusion} introduces the latent-wise diffusion model for image restoration to accelerate the training and sampling, which is shown in Fig.~\ref{fig:refusion}. In contrast to the above works, where latent space is obtained by compressing the original image, DiffIR \citep{xia2023diffir} exploits the latent-wise diffusion model to generate the compact IR priors, which guides dynamic transformer-based restoration network (DIRformer) to achieve better restoration. 

\subsubsection{Efficient Transfer-Learning}
In supervised DM-based IR, efficient and rapid fine-tuning has become a focal point of current research \citep{wang2023exploiting-stable-sr,lin2023diffbir,yang2023pixel-aware,yu2024scaling-SUPIR,jiang2023autodir,ai2023multimodal-mperceriver}. This method principally leverages pre-trained models such as DDPM or Stable Diffusion, extracting certain distortion information through feature extraction and subsequently modulating it into the diffusion model. Pioneering work is StableSR, where a time-aware encoder is trained to extract information from low-quality images and incorporate the learned features into pretrained stable diffusion(trained on LAION-5B \citep{schuhmann2022laion-5B}). On the other hand, DiffBIR \citep{lin2023diffbir} and PASD \citep{yang2023pixel-PASD} introduce conditions by training a UNet encoder using zero convolutions, which bears similarities to ControlNet \citep{zhang2023adding-ControlNet}. Another significant advancement in this area is SUPIR \citep{yu2024scaling-SUPIR}, which builds upon a pre-trained model based on SDXL \citep{podell2023sdxl}. It extracts textual prompts through LLaVA \citep{liu2024visual-LLaVA} and introduces distorted image information using a trimmed Controlnet. 

\renewcommand\arraystretch{1.5}
\subsection{Zero-shot DM-based IR.}
\label{sec:zero-shot}
Different from supervised DM-based IR, zero-shot DM-based IR strives to achieve training-free and data-free image restoration. It generally relies on the pre-trained diffusion models designed for generation tasks and incorporates the condition of low-quality images in the sampling process. The core challenges of this task stem from i) how to maintain the data consistency between low-quality images and generated images, since the pre-trained diffusion models are devoted to preserving the data distribution instead of pixel-wise data consistency; ii) how to excavate the perceptual knowledge aligned with low-quality images, which imposes higher requirements on the design of the condition. In this paper, we roughly summarized the zero-shot DM-based IR methods into three categories, \ieno, projection,  decomposition, and posterior estimation based methods.  

\subsubsection{Projection-based Methods.}
To mitigate the primary challenges in zero-shot DM-based IR, the projection-based method has been introduced in some studies \citep{repaint,choi2021ilvr,CCDF,moser2023yoda}. This approach aims to extract inherent structures/textures from low-quality images as complementary to generated images at each step, which can ensure data consistency.  
For instance, the task of image inpainting involves merely generating the content for the mask region. The unmasked region of the low-quality image can substitute the corresponding part of the denoised image at the $t-1$ step, thereby establishing the condition for data consistency during the sampling process. In line with this, RePaint \citep{repaint} exploits a simple projection for image inpainting task: 
\begin{equation}
    x_{t-1}=m\bigodot x_{t-1}^{known}+(1-m)\bigodot x_{t-1}^{unknown}, 
    \label{eq:unknown}
\end{equation}
where $x_{t-1}^{known}\sim \mathcal{N}(\sqrt{\hat{\alpha}_t}y, 1-\hat{\alpha}_t\mathbf{I})$

is the diffused results at time step $t-1$ by adding noise for masked image $y$. $x_{t-1}^{unknown}$ is sampled from denoised prediction of diffusion model. In contrast, ILVR \citep{choi2021ilvr} employs the low-frequency projection for image super-resolution.
 
Theoretically, at time step $t-1$, the predicted latent variable $x_{t-1}$ and $y_{t-1}$ (\ieno, adding noise to low-resolution image $y$ at the $t-1$ step in the diffusion process) should share the same low-frequency components. Consequently, it substitutes the low-frequency components of $x_{t-1}$ with its counterpart from $y_{t-1}$, ensuring data consistency and establishing an improved condition for the diffusion model. As an advanced solution, CCDF \citep{CCDF} introduces a unified projection method as:
\begin{equation}
    x_{t-1} = Ax'_{t-1} + b,
    \label{eq:linear_projection}
\end{equation}
where $A$ and $b$ are set to achieve data consistency. For instance, in the SR task, the above projection can be instantiated as: 
\begin{equation}
    x_{t-1} = (\mathbf{I}-\mathbf{P})x'_{t-1} + y_{t-1},
    \label{eq:instantiation}
\end{equation}
where $\mathbf{P}$ is the degradation matrix of downsampling. Furthermore, this work proves that generation starting from better initialization can boost the speed in the reverse process.

\subsubsection{Decomposition-based methods.}
It is noteworthy that most image restoration problems can be regarded as linear reverse problems, which can be posed as: 
\begin{equation}
     y = Hx + z,
     \label{eq:linear_degradation}
\end{equation}
where $H$ is the linear degradation operator and $z$ is a contaminating noise. In this setting, the condition probability $p(x|y)$ cannot be directly estimated since the existence of noise $z$. To get rid of the noise $z$, SNIPS \citep{kawar2021snips} and DDRM \citep{DDRMDRM} run the diffusion process in the spectral domain with singular value decomposition (SVD) on the degradation operator $H$. In particular, SNIPS \citep{kawar2021snips} is based on the annealed Langevian dynamics and derives the conditional score function on the spectral space, which achieves great performance on image deblurring, super-resolution, and compressive sensing tasks. Following SNIPS, DDRM \citep{DDRMDRM} further extends the SVD decomposition to the variational objective of linear reverse problems, which reveals a pre-trained DDPM \citep{DDPM}/DDIM \citep{DDIM} can be the optimal solution for it. 
Notably, the above works only focus on the linear reverse problem. In contrast, Kawar et al \citep{jpegddrm} investigate non-linear inverse problems based on the special case of DDRM (\ieno, no noise $z$ in the reverse problem), and extend the pseudo-inverse concept to achieve JPEG artifact correction. For the MRI reconstruction, SVD decomposition is not suitable. To overcome this, Song \etal~train the unconditional generative model for medical images from scratch and then exploit the matrix decomposition in the sampling process to solve the linear reverse problem, which is general for unknown measurement processes.

\begin{figure}[h]
	\centering
	\includegraphics[width=0.9\linewidth]{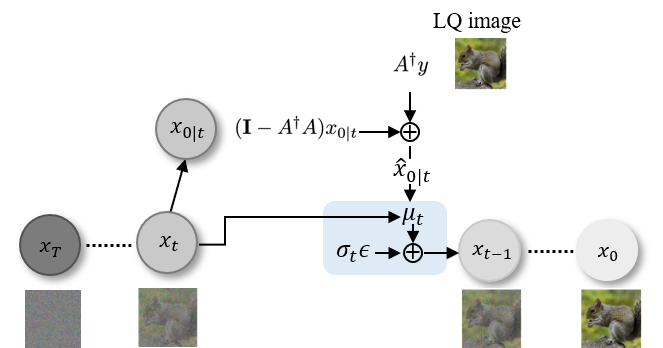}
	\caption{The architecture of DDNM \citep{ddnm},the forward degradation operator $A$ is decomposed into range-null space to meet the realness and data consistency.}
	\label{fig:DDNM}
\end{figure}

With a different purpose, another decomposition strategy, range-null space decomposition is introduced by DDNM \citep{ddnm} to further improve the zero-shot image restoration, where the range space is responsible for the data consistency, and null space is used to improve the reality (\ieno, the perceptual quality). Given the noiseless inverse $y=Hx$, it can be decomposed into:
\begin{equation}
    y=HH^{\dagger}Hx + H(I-H^{\dagger}H)x,
    \label{eq:rn_decom}
\end{equation}
where $H^{\dagger}$ is the pseudo-inverse of degradation operation $H$. We can see the range space $HH^{\dagger}Hx = Hx = y$ can ensure data consistency, and null space $H(I-H^{\dagger}H)x$ has no effects on data consistency since $H(I-H^{\dagger}H)x=\mathbf{0}$. As shown in Fig.~\ref{fig:DDNM}, based on this, DDNM \citep{ddnm} rectifies the prediction of $x_0$ at time step $t$ as:
$\hat{x}_{0|t}= H^{\dagger}y + (I - H^{\dagger}H)x_{0|t}$, where $x_{0|t}$ can be estimated with the noise prediction at time step $t$. With the rectified $\hat{x}_{0|t}$, we can compute the denoised output $x_{t-1}$ at time step $t-1$, which ensures the data consistency and serves as a better condition for the next noise prediction. Moreover, DDNM also exploits the SVD to solve the linear inverse problem with noise, termed DDNM+.

\vspace{0.4em}
\subsubsection{Posterior estimation.}
The projection-based methods have shown superior performance in the inverse problem of image restoration, where a projection-based measurement consistency correction is added after the reverse sampling step of the diffusion model. However, most projection-based works are devoted to the noiseless inverse problems and usually suffer from unsatisfied data consistency, since the projection throws the sample path off the data manifold \citep{mcg}. To solve the general noisy linear reverse problem, some works \citep{mcg,dps,fabian2023diracdiffusion,iigdm,copaint,GDP,cardoso2023monte,garber2023image-DDPG,lee2024spatial-SAFARA,cao2023deep-DeqIR} aim to estimate the posterior distribution $p(x|y)$ with the unconditional diffusion model based on the Bayes theorem. It is equivalent to estimating the conditional posterior $p(x_t|y)$ at each step of the reverse process. Based on the Bayes theorem, it can be derived as:
\begin{equation}
    p(x_t|y) = p(y|x_t)p(x_t)/p(y). 
\end{equation}
And the corresponding score function can be estimated as:
\begin{equation}
    \nabla_{x_{t}}\log p_{t}(x_{t}|y)=\nabla_{x_{t}}\log p_{t}(y|x_{t})+s_{\theta}(x,t),
    \label{eq:score}
\end{equation}
where the $s_{\theta}(x,t)$ could be extracted from pre-trained model while the term $p_{t}(y|x_{t})$ is intractable. From the above equation, we can find that the key factor to achieve better solvers for the inverse problem of image restoration is to accurately estimate the $p(y|x_t)$.

\begin{figure}[h]
	\centering
	\includegraphics[width=1\linewidth]{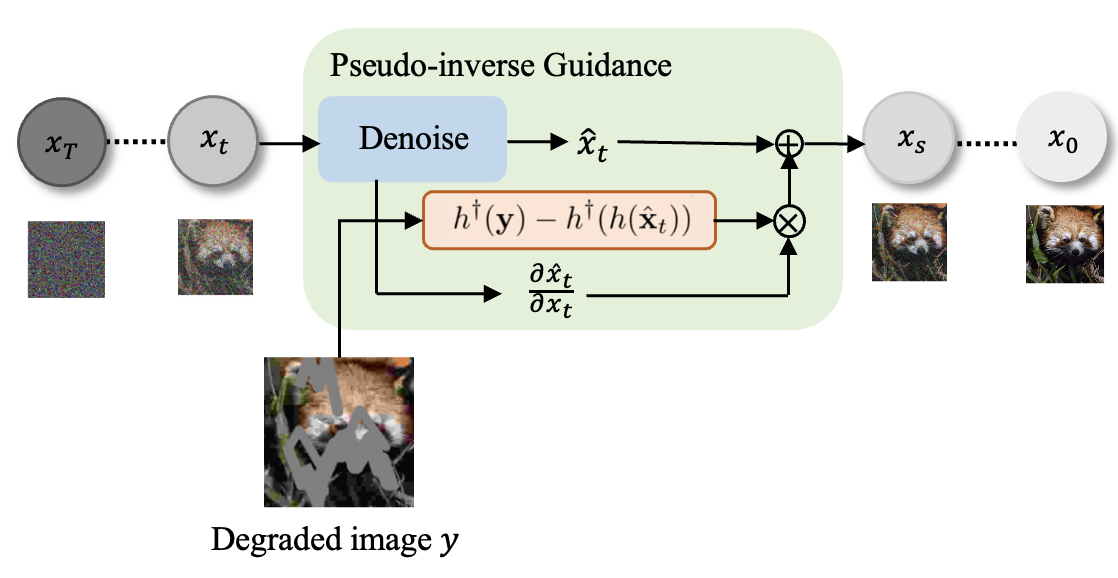}
	\caption{The architecture of $\Pi$GDM \citep{iigdm}. During each denoising step, the pseudo-inverse guidance is employed to encourages data consistency between denoising results and degraded image $y$.  }
	\label{fig:IIGDM}
\end{figure}

As the pioneering works, MCG \citep{mcg} and DPS \citep{dps} approximate the posterior $p(y|x_t)$ with 
$p(y|\hat{x}_o)$, and $\hat{x}_o$ is the expectation given $x_t$ as $\hat{x}_o=E[x_o|x_t]$ \citep{dps} with Tweedie's formula. Concretely, MCG \citep{mcg} considers the data consistency from the perspective of the data manifold, where the manifold constrained gradient is proposed to let the correction lie on the data manifold. 
However, DPS \citep{dps} pinpoints that the projection operation in MCG is harmful to data consistency since it might cause the sampling path off the data manifold. Based on this, DPS \citep{dps} discards the projection step in the reverse process and estimates the posterior as:
\begin{align}
    \nabla_{x_{t}}\log p_{t}(y|x_{t})& \approx \nabla_{x_{t}}\log p(y|\hat{x}_{0})  \notag \\
    & \approx -\frac{1}{\sigma^2}\nabla_{x_{t}} \Vert y-H(\hat{x}_{0}(x_{t})) \Vert_{2}^2 
    \label{eq:DPS}
\end{align}
Following the above works, $\Pi$GDM \citep{iigdm} further expands the Eq.~\ref{eq:DPS} to the unified form for the linear, non-linear, differentiable inverse problem with Moore-Penrose pseudoinverse $h^{\dagger}$ of degradation function $h$ as: 
\begin{equation}
    \nabla_{x_{t}}\log p_{t}(y|x_{t})\approx r_t^{-2} ((h^\dagger(y)-h^\dagger(h(\hat{x}_o)))^{\mathrm{T}}\frac{\partial \hat{x}_o}{\partial x_t})^{\mathrm{T}}
\end{equation}
where $r_t^{-2}$ is set as $\sqrt{\frac{\sigma_t^2}{\sigma_t^2+1}}$ and $h$ is the non-linear degradation function. Based on the equation, $\Pi$GDM developed its pipeline as Fig.~\ref{fig:IIGDM}. 

Different from the above works, some works \citep{GDP,copaint} attempt to model $p(y|x_{t})$ with other strategies. It is noteworthy that the higher conditional probability $p(y|x_{t})$ is equivalent to  a smaller distance between $D(x_{t})$ and $y$ \citep{GDP}. Therefore, 
GDP \citep{GDP} proposes a heuristic approximation of distribution $p(y|x_{t})$ as follows: 
\begin{equation}
    p(y|x_{t})\approx \frac{1}{Z}\exp(-[s\mathcal{L}(\mathcal{D}(x_{t}),y)+\lambda\mathcal{Q}(x_{t})]),
    \label{Eq:GDP}
\end{equation}
where $\mathcal{L}$ and $\mathcal{Q}$ denote the distance metric and the quality loss, respectively. $Z$ is a normalization factor and $s$ is the scaling factor that controls the weight of guidance. However, 
the distance $\mathcal{L}$ is hard to be defined since the noise magnitudes in $x_t$ and $y$ are different. Thus, they substitute $x_t$ with its clean estimation $\hat{x}_0$ in the distance measurement.
With the same purpose, Copaint \citep{copaint} tries to predict the $\hat{x}_{0}$ through the one-step estimation by a neural network.

Instead of modeling intractable distribution $p(y|x_{t})$, Feng~\etal \citep{feng2023score} directly estimates the posterior $p(x_{t}|y)$ from a variational perspective. Following DPI \citep{sun2021deep-dpi1,sun2022alpha-dpi2}, they define a family of distributions $q_{\theta}$ through RealNVP \citep{RealNVP} normalizing flow with parameter $\theta$, which is optimized through a minimal KL-divergence between true posterior and estimated distribution $q_{\theta}$.

\section{Diffusion Models for Blind/Real-world Image Restoration}
\label{sec:blindir}
Although the methods in Sec.~\ref{sec:Diffuion-based IR} have achieved great breakthroughs in image restoration, most of them \citep{dps,SR3,DDRMDRM,li2022srdiff,saharia2022palette,weather-diff,CDM,iigdm,ddnm} focus on solving the synthetic distortions, which usually perform poorly in the out-of-distribution (OOD) real-world/blind degradations.
The reasons stem from the inherent challenges of real-world IR: 1) the unknown degradation modes are hard to be identified. 2) collecting distorted/clean image pairs is non-trivial and even unavailable in the real world. To overcome this, previous works \citep{wang2022blind-simulate3,liang2022efficient-simulating2,wang2022ucl-unsupervised} have attempted to solve it by simulating real-world degradations \citep{wang2021realesrgan,zhang2021designing-sr3+,zhang2022single-simulating1,wang2022blind-simulate3}, and unsupervised learning \citep{wang2021unsupervised,lugmayr2019unsupervised,wang2022ucl-unsupervised}, etc. Inspired by these, some pioneering works \citep{SR3+,blinddps,wang2023dr2,yin2022diffgar,abu2022adir} begin to explore how to exploit diffusion models to solve real-world degradations. In this paper, we divided the DM-based blind/real-world IR \citep{SR3+,yin2022diffgar,dg-dpm,wang2023exploiting-stable-sr,welker2022driftrec,yang2023synthesizing,chan2023sud,wang2023dr2,murata2023gibbsddrm,blinddps,miao2023dds2m} into four categories, \ieno, distortion simulation \citep{SR3+,yang2023synthesizing},  kernel estimation \citep{blinddps,murata2023gibbsddrm}, domain translation \citep{wei2023raindiffusion,yang2023synthesizing}, and Distortion-invariant diffusion model \citep{wang2023dr2,yue2022difface,dg-dpm}.

\subsection{Distortion Simulation}
Notably, the real-world distortions are usually blind/u-nknown, where the distributions are different from the simple synthetic distortions. For supervised-learning-based IR, this requires the restoration network to have strong generalization capability or the synthetic datasets can cover the real-world distortions. From the causality perspective \citep{li2023learningcausalIR}, these two purposes all rely on simulating diverse distortions that are similar to real-world distortions, which we called distortion simulation/augmentation. There are several representative DM-based IR methods \citep{SR3+,yin2022diffgar} utilizing the distortion simulations to improve the robustness of their methods for real-world degradations. The representative method is SR3+ \citep{SR3+}, which is based on the diffusion model from SR3 \citep{SR3} and introduces the second-order degradations simulation of RealESRGAN \citep{wang2021realesrgan} for the training. Similarly, to simulate the real-world degradation, Yang~\etal \citep{yang2023synthesizing} propose to synthesize the real-world distorted/clean training pairs using diffusion models, where the distorted images are initialized with second-order degradation in RealESRGAN \citep{wang2021realesrgan}.

\subsection{Kernel Estimation}
Kernel Estimation is first proposed in blind image restoration \citep{tzikas2009variational-blind0,huang2020joint-blind1,vasu2018non-blind2,oliveira2013parametric-blind3}, where the degradation can be modeled as $y = (x*k)\downarrow_{s} + n$. Here, $k$ is the degradation kernel, and $n$ is the additive noise. Under this setting, the kernel $k$ can be estimated as the guidance to boost the adaptability of the restoration network. Inspired by this, BlindDPS \citep{blinddps} and GibbsDDRM \citep{murata2023gibbsddrm} attempt to solve the blind inverse problem by estimating the unknown degradation kernel in the sampling process. In particular, BlindDPS \citep{blinddps} exploit the DPS \citep{dps} architecture and exploit one parallel diffusion model for the degradation kernel estimation. The diffusion model for kernel estimation is pre-trained on synthetic kernels. Unlike BlindDPS, GibbsDDRM \citep{murata2023gibbsddrm} achieves the sampling process with partially collapsed Gibbs sampler \citep{van2008partially-sampler}, which samples both kernel parameter and image together from the joint posterior $p(x_{t}|k,y)$.

\subsection{Domain Translation}
In the real world, it is hard to collect distorted/clean image pairs. Although some works attempt to simulate the degradation process of real distorted images, the distribution of the synthesized distortion is still far away from the real-world one. 
To further solve the Real-world IR problem, a series of works explore the domain translation techniques for image restoration.
Domain Translation \citep{anoosheh2018combogan-translation0,lin2019learning-translation0.5,chu2018survey-translation1,saunders2022domain-translation2,lugmayr2019unsupervised-translation} aims to translate the image from one domain to another domain. From the domain translation perspective, synthetic distorted images, real-world distorted images, and high-quality images can be regarded as three different domains, that share the same contents.

The domain translation-based works for DM-based IR can be roughly divided into two categories: 1) The first line \citep{yang2023synthesizing} aims to simulate more reliable real-world distorted/clean image pairs by translating the low-quality images from the synthetic domain to the real-world domain. In this way, the simulated datasets can enable the restoration network with better restoration capability for real-world degradation. For instance, Yang~\etal \citep{yang2023synthesizing} 
is the first to exploit the pre-trained diffusion model to synthesize real-world training pairs, where the diffusion model is pre-trained with real-world low-quality images, and the translation is achieved by warping the synthetic low-quality image to noise space (\ieno, the generation inversion).  2) Another popular line \citep{wei2023raindiffusion} exploits unsupervised learning, where domain translation is achieved by the cycle consistency constraint. In particular, two generators construct one cycle path, where one generator aims to translate the distorted image to the distortion-free image, and another generator is utilized to translate the clean image to the distorted image. This enables unsupervised training with unpaired real-world distorted and high-quality images. As the representative work, RainDiffusion \citep{wei2023raindiffusion} proposes to remove the rain with two cooperative branches, where the non-diffusive translation branch aims to utilize the pre-trained cycle-consistent generators to produce initial paired clean/rainy images, and diffusive translation branch leverages the multi-scale diffusion models to refine the results.

\subsection{Distortion-invariant Diffusion Model}
\begin{figure}[h]
	\centering
	\includegraphics[width=1\linewidth]{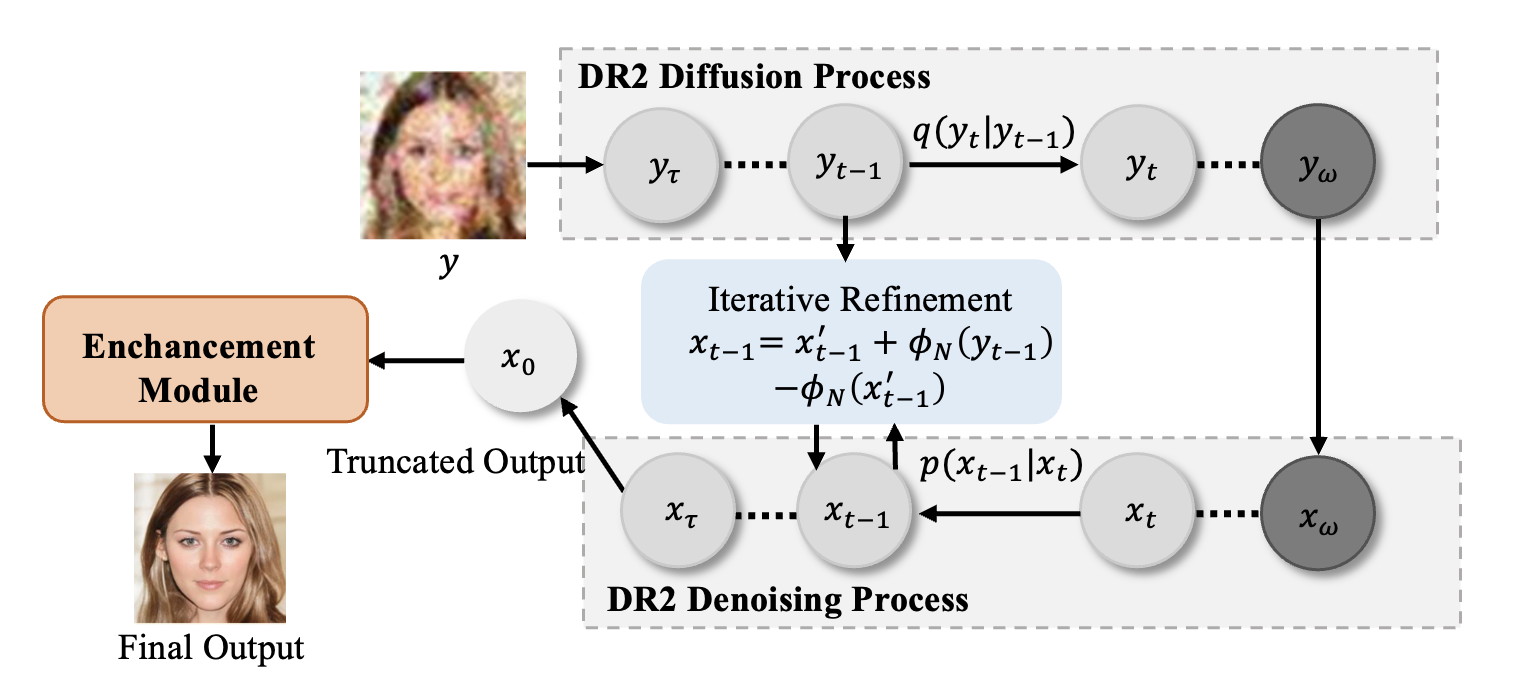}
	\caption{The architecture of DR2 \citep{wang2023dr2}. It consists degradation removal module and an enhancement module. The input image is first sent to a diffusion process to remove degradation information gradually. The output degradation-free image will be enhanced by a pre-trained face restoration network.}
	\label{fig:DR2}
\end{figure}

Since the blind distortions are usually diverse and complicated, the diffusion model is required to own the generation capability for these distortions (\ieno, the distortion-invariant capability) in the real world. To achieve the distortion-invariant diffusion model, DifFace~\etal \citep{yue2022difface} introduces a pre-trained restoration network, \egno, SRCNN \citep{SR1_SRCNN-cnn} or SwinIR \citep{swinIR-transformer}, to obtain an initial clean image as a sampling start point $x_N$, where the restoration network is trained with the second-order degradation from RealE-SRGAN \citep{wang2021realesrgan}, thereby exhibiting good generalization capability and produce the distortion invariant initial clean image for diffusion model. Ren~\etal \citep{dg-dpm} proposes to achieve a distortion-invariant diffusion model with multi-scale degradation-invariant guidance information. They employ distortion augmentation strategies on degraded images to obtain invariant representation with structure information as guidance.  In contrast, Wang \citep{wang2023dr2} exploits the low-pass filter to filter the distortion invariant components in the low-quality image since the different real-world distorted images usually share the same structure information.  As shown in Fig.~\ref{fig:DR2}, they adopt a simple iterative refinement similar to ILVR during the sampling stage. After obtaining the degradation-invariant $\hat{x}_{0}$, they use an enhancement module (Powerful CNN-based or transformer-based restoration methods) to further improve the image quality.

\subsection{Discussion on Some Real-world Application and Limitation.}
\noindent\textbf{Low-light enhancement.}
Real-world low-light image restoration aims to enhance and restore images captured under poor lighting conditions, which is susceptible to low brightness, high noise, color distortion, and reduced contrast, posing significant challenges for restoration methods. To achieve this, LLDiffusion~\citep{wang2023lldiffusion-LLDiffuion} introduces a degradation generation model to capture the degradation representation for low-light enhancement. This degradation-aware condition enhances the performance of diffusion-based low-light image restoration by providing more informative priors. 
To improve the efficiency of the diffusion model while preserving the generation capability, DiffLL~\citep{jiang2023lowDiffLL} decouples the low-light image with wavelet-transform, using the diffusion model only to the low-frequency component. This approach significantly reduces the inference time while preserving the high-frequency details through a dedicated high-frequency restoration module(HFRM).
PyDiffusion~\citep{zhou2023pyramid-pydiff} incorporates a multi-scale generation framework, applying the diffusion model at different scales to mitigate the computational inefficiency of direct high-resolution generation. This hierarchical approach enhances both efficiency and performance.
There are also two diffusion-based works incorporating physically inspired priors to serve as conditioning inputs to the diffusion model, improving restoration quality by explicitly modeling the underlying lighting and reflectance properties. For instance, Diff-Retinex~\citep{yi2023diff-retinex} decomposes low-light images into illumination and reflectance maps based on the Retinex model. LightenDiffusion~\citep{jiang2024lightendiffusion} introduces an unsupervised training mechanism, it decouples the unpaired normal-light image and low-light images to extract the illumination prior from normal-light image and content reflection from low-light image, then combine them with diffusion model to refine it.

As shown in Table~\ref{table:results-low-light}, we have compared recent diffusion-based works on real-world low-light image restoration on commonly-used synthetic dataset LOL~\citep{lol} and three commonly-used real-world datasets, including DICM~\citep{lee2013contrastDICM}, NPE~\citep{wang2013naturalnessNPE}, and VV~\citep{vonikakis2018evaluationVV}. We can observe that DiffLL\citep{jiang2023lowDiffLL} and LightenDiffusion\citep{jiang2024lightendiffusion} achieve optimal generalization in terms of subjective metrics, \ieno, NIQE and PI. However, due to its unsupervised training mechanism, LightenDiffusion exhibits lower objective quality compared to DiffLL~\citep{jiang2023lowDiffLL}.

\begin{table*}[htp]
\centering
\caption{Comparisons of Real-world Low-light Image Restoration. Results are tested on three FR metrics: PSNR$\uparrow$, SSIM$\uparrow$, LPIPS$\downarrow$ and two NR metrics: NIQE$\downarrow$, PI$\downarrow$.}
\resizebox{0.7\textwidth}{!}{
\begin{tabular}{c|ccc|cc|cc|cc}
\hline
\multirow{3}{*}{Models} & \multicolumn{3}{c|}{LOL~\citep{lol}} & \multicolumn{2}{c|}{DICM~\citep{lee2013contrastDICM}} & \multicolumn{2}{c|}{NPE~\citep{wang2013naturalnessNPE}} & \multicolumn{2}{c}{VV~\citep{vonikakis2018evaluationVV}}\\ \hline

                        & PSNR           & SSIM           &LPIPS                    & NIQE  & PI  & NIQE & PI   & NIQE & PI     \\ \hline
\midrule
CLE Diffusion~\citep{yin2023cle-CLE-Diffusion} & 25.51 & 0.890 & 0.160 & 4.505 & 3.361 & 5.249 & 3.512 & 3.240 & 3.470 \\
GDP~\citep{GDP} &  15.90 & 0.542 & 0.337 & 4.358 & 3.552 & 4.032 & 3.097 & 4.683 & 3.431\\
DiffLL~\citep{jiang2023lowDiffLL} & \textbf{26.34}& 0.845 & 0.217  & \textbf{3.636} & \textbf{2.936} & 3.716 & 2.629 & 2.351 & 2.869 \\
LLDiffusion~\citep{wang2023lldiffusion-LLDiffuion} & 24.65 &0.843& 0.075 & - & - & - & - & - & -\\
Diff-Retinex~\citep{yi2023diff-retinex} &  21.98 & \textbf{0.863} & \textbf{0.048} & 4.361 & 3.394 & 4.996 & 3.392 & 3.087 & 3.350\\
PyDiffusion~\citep{zhou2023pyramid-pydiff} & 23.28 & 0.859 & 0.108 &4.499 & 3.792 & 4.082 & 3.268 & 4.360 & 3.678 \\
LightenDiffusion~\citep{jiang2024lightendiffusion}  & 20.45 & 0.803 & 0.192  & 3.724 & 3.144 & \textbf{3.618} & 2.879 & 2.941 & 2.558 \\
\hline
\end{tabular}}
\label{table:results-low-light}
\end{table*}

\noindent\textbf{Adverse weather removal.} As the pioneering work, WeathDiff~\citep{weather-diff} shares the same diffusion architecture with SR3, but it can support arbitrary image restoration by introducing the mean estimated noise of overlapped patches in the reverse process. In contrast, T$^3$-Diffusion~\citep{chen2024teachingt3diffusion} defined a group of degradation-specific sub-prompts and utilizes the dynamical composition of them based on unseen degradation for weather removing, which outperforms DA-CLIP~\citep{luo2023controlling-DACLIP} and Weather-Diff. ReviveDiff~\citep{huang2024revivediff} enhances the perception levels for various weather degradation by introducing the coarse and fine branches into the diffusion model. RainDiff/RainDiffusion~\citep{shen2023rethinkingRainDiff} proposes an unpaired cycle consistency framework for image deraining, eliminating the need for paired training data, making it highly suitable for real-world applications. Depart from that, there are also two diffusion-based works for image dehazing. Dehaze-DDPM~\citep{yu2023high-HazeDDPM} employs a two-stage dehazing pipeline, combining the Atmospheric Scattering Model (ASM) for physical modeling in the first stage and diffusion-based generative recovery of lost structural details in the second. Additionally, Wang et al.~\citep{wang2023frequency-dehazy} enhance dehazing by emphasizing mid-to-high-frequency components in the diffusion process, leveraging a frequency-spectrum filter and skip connections to inject high-frequency details into the U-Net architecture.

\noindent\textbf{Shadow removal.} ShadowDiffusion~\citep{guo2023shadowdiffusion} leverages a pre-trained transformer backbone to extract the degradation prior (i.e., the degradation-related features) from the distorted reference image. This extracted degradation prior is exploited as the auxiliary to refine the generated shadow mask and serves as the condition for
shadow-free image generation. DeS3 \citep{jin2022des3} introduces a novel adaptive attention mechanism to distinguish the underlying objects and shadow regions, which is mask-free compared with existing works on shadow removal. Moreover, it utilizes a pre-trained ViT backbone as loss to enhance the scene understanding capability. As described in DeS3, it can obtain a better performance compared with ShadowDiffusion.

\noindent\textbf{Potential limitations.} The potential limitations of current diffusion-based methods in addressing real-world IR are as:
(i) \textbf{Limited task relevance.} Some diffusion-based image restoration (IR) methods overlook the inherent characteristics of the tasks themselves and instead rely solely on dataset-specific fitting;  (ii) \textbf{Limited comparison reasonability.} 
 As discussed earlier, diffusion-based methods~\citep{weather-diff,wei2023raindiffusion} often use different datasets in some IR task, making it difficult to identify which approach is truly more effective. Furthermore, real-world datasets are challenging to collect and are often insufficient for comprehensive evaluation;
(iii) \textbf{Limited restoration efficiency.}
As stated in Section~\ref{sec:future} of our manuscript, the impressive perceptual quality of diffusion-based IR methods is largely dependent on multiple sampling steps. Although one-step diffusion-based IR methods~\citep{wang2023sinsr, wu2024oneOSEDiff} have been proposed, accelerating the sampling process while maintaining restoration quality remains a significant challenge; (iv) \textbf{Limited pixel-wise consistency.} Diffusion-based real image restoration methods~\citep{yu2024scaling-SUPIR,jiang2024lightendiffusion} share a common weakness with GAN-based approaches, \ieno, they are susceptible to hallucinated textures due to the generative priors stored within the diffusion model. Addressing this issue remains a critical challenge.

\vspace{0.4em}

\section{Experiments}
\label{sec:experiments}
To ensure the efficient and thorough comparison of different diffusion model-based IR methods, we first summarize the popular datasets, experimental configurations, and evaluation metrics for different tasks. Then we have a comparison of existing benchmarks in several typical image restoration tasks, including image super-resolution, inpainting, deblurring, and JPEG artifacts removal.

\subsection{Datasets and Implementation Details}
\label{sec:datasets}
\noindent\textbf{Datasets.}
It is noteworthy that the contents and degradation modes are significantly different across the datasets from different IR tasks. Therefore, we summarize the commonly-used datasets based on IR tasks, including SR, image deblurring, image inpainting, shadow removal, desnowing, draining, and dehazing in Table 3 of section D of the \textbf{Supplementary}.

\vspace{0.3em}
\noindent\textbf{Implementation Details.} 
We summarize the implementation details and datasets of supervised algorithms and zero-shot algorithms in Table 4 and Table 5 of the \textbf{Supplementary}, respectively. For supervised algorithms, we describe the configurations in the training process and testing process, including batch size, training iterations, learning rate, the sampling steps in the training process, and the sampling steps in the inference process. For zero-shot-based methods, we clarify the pre-trained diffusion model, evaluated datasets, and sampling steps in the inference process. The commonly used data augmentation strategies are composed of rotation and flip operations.

\begin{table*}[ht]
\centering
\caption{Quantitative Results of Supervised Models on Super-resolution Task. $\dagger$ denotes that we reproduce the results by retraining.}
\label{tab:supervised_results}
\resizebox{\linewidth}{!}{
\begin{tabular}{l|cccc|cccc|c|c|c|c}
\hline
\multirow{2}{*}{Models} & \multicolumn{4}{c|}{DIV2K} & \multicolumn{4}{c|}{Urban100} & \multirow{2}{*}{Time (s/image)} & \multirow{2}{*}{Parameters} & \multirow{2}{*}{Flops (G)} & \multirow{2}{*}{NFE}\\ \cline{2-9}
 & PSNR & SSIM & LPIPS & FID & PSNR & SSIM & LPIPS & FID & & & \\ \hline
Bicubic & 28.17 & 0.774 & 0.31 & 108.2 & 23.17 & 0.658 & 0.34 & 134.3 & - & - & -& -  \\ \hline \midrule
RealESRGAN+ \citep{wang2021realesrgan} & 28.18 & 0.776 & 0.115 & 13.57 & 24.37 & 0.734 & 0.123 & 20.29 & 0.0327 & 16.7M & 73.43G&1\\
CAL-GAN \citep{CAL-GAN} & 28.96 & 0.790 & 0.108 & 13.22 & 25.33 & 0.762 & 0.117 & 18.47 & 0.0327 & 16.7M & 73.43G&1\\
SeD \citep{li2024sed} & 29.27 & 0.803 & \textbf{0.094} & \textbf{12.51} & 25.93 & 0.780 & \textbf{0.107} & \textbf{16.39} & 0.0327 & 16.7M & 73.43G&1\\ 
RCoT \citep{tang2024residualRCoT} & 29.63 & 0.819 & 0.279 & 20.04 & 24.87 & 0.742 & 0.270 & 33.63 & 0.0164 & 6.74M & 204.74G&1
\\ \hline \midrule
PromptIR$^\dagger$ \citep{potlapalli2023promptir} & 30.73 & 0.844 & 0.262 & 16.75 & 26.71 & 0.804 & 0.209 & 26.41 & 0.0602 & 33.19M & 11.91G&1 \\
Restormer$^\dagger$ \citep{zamir2022restormer_deblur-transformer} & 30.73 & 0.844 & 0.261 & 16.65 & 26.74 & 0.804 & 0.209 & 26.11 & 0.0557 & 26.32M & 10.84G&1\\
RCAN \citep{zhang2018image-RCAN} & 30.78 & 0.844 & 0.255 & 16.41 & 26.74 & 0.807 & 0.196 & 24.23 & 0.0863 & 15.4M & 65.25G & 1\\
IPT \citep{chen2021preIPT} & 30.92 & 0.847 & 0.253 & 15.76 & 27.33 & 0.820 & 0.189 & 22.92 & 0.0379 & 115M & 43.41G & 1\\
SwinIR \citep{swinIR-transformer} & 31.08 & 0.852 & 0.247 & 15.29 & 27.46 & 0.826 & 0.184 & 21.98 & 0.0384 & 11.8M & 37.17G &1\\
HAT \citep{chen2023hat}& 31.22 & 0.855 & 0.245 & 15.33 & 27.97 & 0.836 & 0.173 & 20.49 & 0.0613 & 20.6M & 102.4G &1\\
MambaIRv2 \citep{guo2024mambairv2}& \textbf{31.24} & \textbf{0.858}  & 0.246 & 15.94 & \textbf{28.07} & \textbf{0.838} & 0.171 & 20.83 & 0.0626 & 34.2M & 82.3G&1\\
\hline \midrule
SR3$^\dagger$ \citep{SR3} & 20.37 & 0.629 & 0.370 & 46.04 & 20.60 & 0.635 & 0.288 & 43.55 & 50.4 & 155.3M & 155.2G & 100 \\ 
SRDiff \citep{li2022srdiff}  & 28.62 & 0.789 & 0.130 & 16.93 & 25.14 & 0.758 & 0.138 & 20.65 & 4.5 & 13.2M & 84.22G & 100 \\ 
Resshift \citep{yue2023resshift} & 29.38 & 0.805 & 0.130 & 15.41 & 25.68 & 0.766 & 0.122 & 20.19 & - & - & - & 15\\ 
DiffBIR \citep{lin2023diffbir}& 25.65 & 0.702 & 0.187 & 26.45 & 23.49 & 0.676 & 0.204 & 37.68 & 4.65 &1716.71M & 189.47G&50 \\
SUPIR \citep{yu2024scaling-SUPIR}& 27.19 & 0.728 & 0.192 & 24.94  & 23.26 & 0.669 & 0.182 & 33.38 &7.64 & 4801.1M & 214.43G &50\\\hline 
\end{tabular}
}
\end{table*}

\begin{table*}[ht]
\centering
\caption{Quantitative Results of Zero-shot Models on Super-resolution Task. Best performances are bolded.}
\label{tab:supervised_imagenet_celeba}
\resizebox{0.8\linewidth}{!}{
\begin{tabular}{l|cccc|cccc|c|c|c}
\hline
\multirow{2}{*}{Models} & \multicolumn{4}{c|}{ImageNet 1K} & \multicolumn{4}{c|}{CelebA-HQ test} & \multirow{2}{*}{Time (s/image)} & \multirow{2}{*}{Flops (G)} & \multirow{2}{*}{NFE}\\ \cline{2-9}
 & PSNR & SSIM & LPIPS & FID & PSNR & SSIM & LPIPS & FID & & \\ \hline
Bicubic & 25.98 & 0.699 & 0.461 & 87.56 & 28.91 & 0.786 & 0.355 & 154.69 & - & - & -\\ \hline \midrule
RealESRGAN~\citep{wang2021realesrgan} & 25.94 & 0.705 & 0.123 & 30.29 & 29.48 & 0.801 & 0.099 & 26.60 & 0.0327 & 73.43G& 1\\
CAL-GAN~\citep{CAL-GAN} & 26.92 & 0.736 & 0.122 & 29.14 & 30.00 & 0.815 & 0.101 & 22.40 & 0.0327 & 73.43G& 1\\
SeD \citep{li2024sed} & 27.01 & 0.741 & \textbf{0.109} & \textbf{27.10} & 29.62 & 0.807 & \textbf{0.089} & \textbf{18.96} & 0.0327 & 73.43G& 1\\ 
RCOT \citep{tang2024residualRCoT}& 27.36 & 0.755 & 0.297 & 57.42 & 30.28 & 0.825 & 0.229 & 75.13 & 0.0164 & 204.74G& 1\\ \midrule   
PromptIR$^\dagger$~\citep{potlapalli2023promptir} & 28.35 & 0.787 & 0.274 & 50.13 & 30.99 & 0.842 & 0.224 & 62.79 & 0.0602 & 11.91G& 1\\
Restormer$^\dagger$~\citep{zamir2022restormer_deblur-transformer} & 28.34 & 0.786 & 0.274 & 49.78 & 30.98 & 0.842 & 0.224 & 62.23 & 0.0557 & 10.84G& 1\\
RCAN~\citep{zhang2018image-RCAN} & 28.35 & 0.786 & 0.269 & 50.92 & 30.95 & 0.839 & 0.222 & 60.87 & 0.0863 & 65.25G & 1\\
IPT~\citep{chen2021preIPT} & 28.39 & 0.787 & 0.268 & 50.26 & 30.98 & 0.839 & 0.220 & 59.47 & 0.0379 & 43.41G & 1\\
SwinIR~\citep{swinIR-transformer}& 28.56 & 0.792 & 0.263 & 49.57 & 31.03 & 0.841 & 0.220 & 60.72 & 0.0384 & 37.17G & 1\\
HAT~\citep{chen2023hat} & \textbf{28.67} & 0.795 & 0.262 & 48.20 & 31.08 & 0.841 & 0.220 & 60.69 & 0.0613 & 102.4G & 1\\
MambaIR~\citep{guo2024mambair} & 28.62 & 0.793 & 0.264 & 48.87 & 31.07 & 0.843 & 0.219 & 59.36 & 0.0626 & 82.30G & 1\\
\hline \midrule
ILVR~\citep{choi2021ilvr}  & 27.40 & 0.871 & 0.21 & 43.76 & 31.59 & 0.878 & 0.22 & 32.24 & 41.3 & 1113.75 &250 \\ 
SNIPS~\citep{kawar2021snips}  & 24.31 & 0.684 & 0.21 & 124.54 & 27.34 & 0.675 & 0.27 & 105.19 & 31.4 & - &1000\\ 
DDRM~\citep{DDRMDRM}  & 27.38 & 0.869 & 0.22 & 40.75 & \textbf{31.64} & \textbf{0.946} & 0.19 & 31.04 & 10.1 & 1113.75 &20\\ 
DPS~\citep{dps} & 25.88 & 0.814 & 0.15 & 39.24 & 26.95 & 0.878 & 0.18 & 29.97 & 141.2 & 1113.75 &1000 \\ 
DDNM~\citep{ddnm}  & 27.46 & \textbf{0.871} &0.15 & 40.15 & 31.64 & 0.945 & 0.16 & 28.27 & 15.5 & 1113.75 & 100 \\ 
GDP~\citep{GDP}  & 26.51 & 0.832 & 0.14 & 38.45 & 28.65 & 0.876 & 0.17 & 27.51 & \textbf{3.1} & 1113.76 &20 \\ \hline
\end{tabular}
}
\end{table*}

\begin{figure*}[h]
	\begin{minipage}{0.135\linewidth}
		\vspace{3pt}
       
		\centerline{\includegraphics[width=\textwidth]{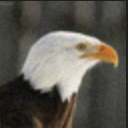}}
	\end{minipage}
	\begin{minipage}{0.135\linewidth}
		\vspace{3pt}
		\centerline{\includegraphics[width=\textwidth]{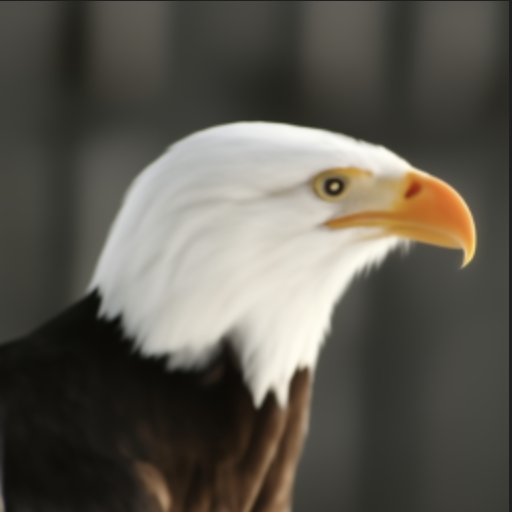}}
	 
	\end{minipage}
	\begin{minipage}{0.135\linewidth}
		\vspace{3pt}
		\centerline{\includegraphics[width=\textwidth]{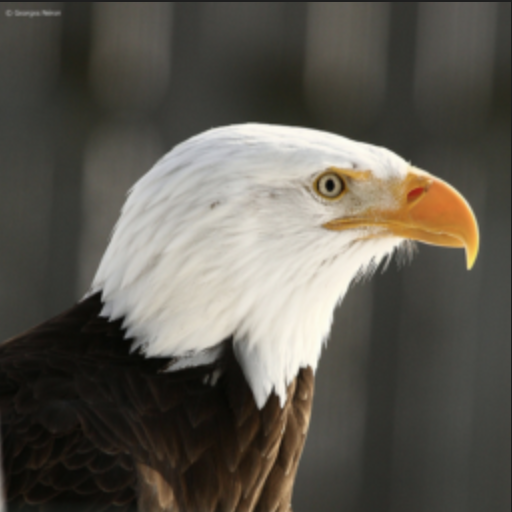}}
	 
	\end{minipage}
 \begin{minipage}{0.135\linewidth}
		\vspace{3pt}
		\centerline{\includegraphics[width=\textwidth]{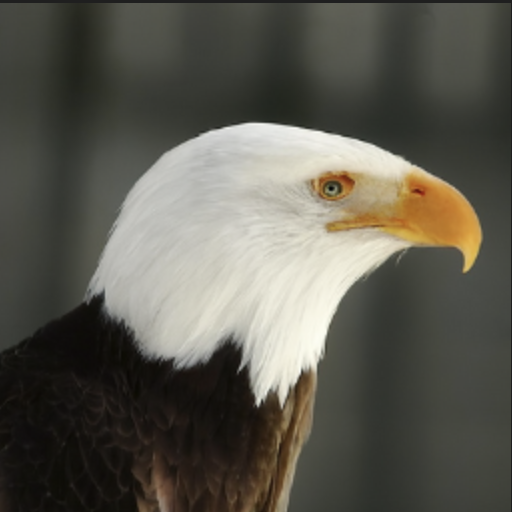}}
	\end{minipage}
 \begin{minipage}{0.135\linewidth}
		\vspace{3pt}
		\centerline{\includegraphics[width=\textwidth]{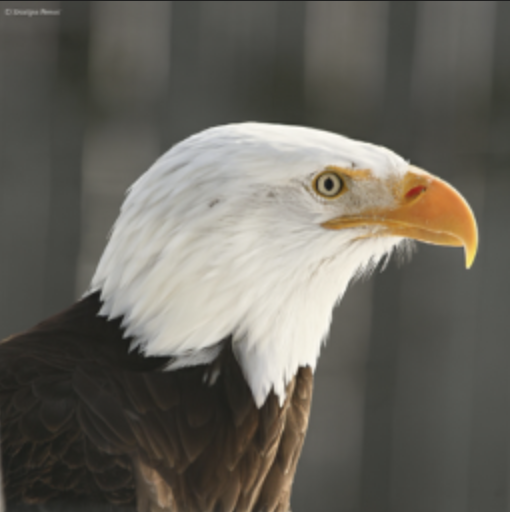}}
	\end{minipage}
 \begin{minipage}{0.135\linewidth}
		\vspace{3pt}
		\centerline{\includegraphics[width=\textwidth]{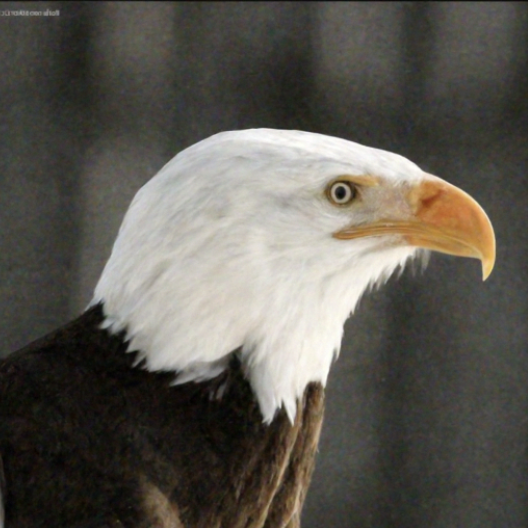}}
	\end{minipage}
\begin{minipage}{0.135\linewidth}
		\vspace{3pt}
        \centerline{\includegraphics[width=\textwidth]{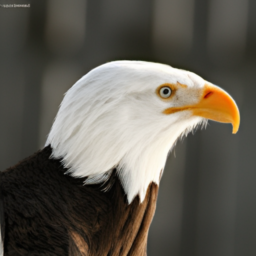}}
	\end{minipage}

\begin{minipage}{0.1354\linewidth}
		\vspace{3pt}
		\centerline{\includegraphics[width=\textwidth]{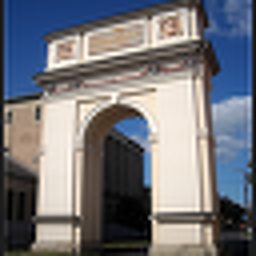}}
		\centerline{Reference}
	\end{minipage}
	\begin{minipage}{0.1354\linewidth}
		\vspace{3pt}
		\centerline{\includegraphics[width=\textwidth]{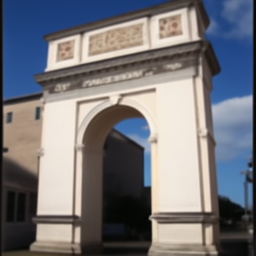}}
	 
		\centerline{DDRM}
	\end{minipage}
	\begin{minipage}{0.1354\linewidth}
		\vspace{3pt}
		\centerline{\includegraphics[width=\textwidth]{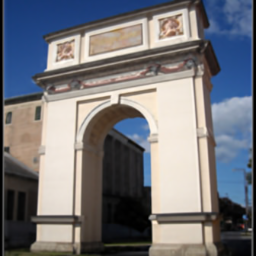}}
	 
		\centerline{DDNM}
	\end{minipage}
 \begin{minipage}{0.1354\linewidth}
		\vspace{3pt}

		\centerline{\includegraphics[width=\textwidth]{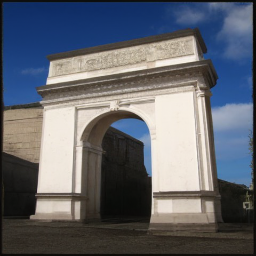}}
		\centerline{DPS}
	\end{minipage}
 \begin{minipage}{0.1354\linewidth}
		\vspace{3pt}

		\centerline{\includegraphics[width=\textwidth]{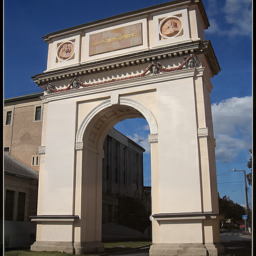}}
		\centerline{GDP}
	\end{minipage}
 \begin{minipage}{0.1354\linewidth}
		\vspace{3pt}
		\centerline{\includegraphics[width=\textwidth]{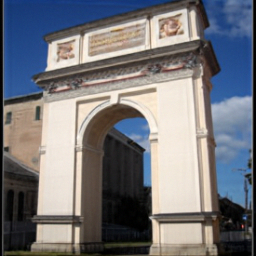}}
          
		\centerline{DiffPIR}
	\end{minipage}
\begin{minipage}{0.1354\linewidth}
		\vspace{3pt}
        \centerline{\includegraphics[width=\textwidth]{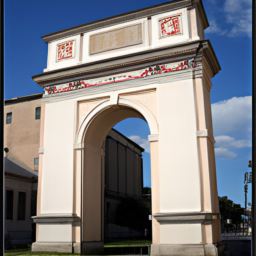}}
		\centerline{StableSR}
	\end{minipage}
	\caption{Qualitative results of diffusion models on Image Super-resolution  }
	\label{Fig:qualitative-sr}
\end{figure*}

\subsection{Experimental Results}
To demonstrate the superiority of different diffusion models, we provide an objective quality comparison for them on multiple tasks. Concretely, we select three commonly investigated IR tasks, including image super-resolution, image deblurring, and image inpainting. The evaluation metrics are composed of PSNR, SSIM \citep{SSIM}, FID \citep{FID}, and LPIPS \citep{LPIPS}. To compare the computation cost and network complexity, we also measure the running time, parameters, and flops for diffusion model-based IR methods. The subjective comparisons of image deblurring, image inpainting, and real-world IR are shown in Fig.~\ref{Fig:qualitative-sr}, Fig. 3, 4, 5, 6 and 7 of section D.6 of the \textbf{Supplementary}.  The comparison of diffusion-based methods on image inpainting and severe composite degradation can be found in the section D.4, D.5 of the \textbf{Supplementary}.

\noindent\textbf{Results on Image SR and Real Image Restoration/SR.}
The experimental settings are as: (i) For the comparison of supervised diffusion model (DM)-based super-resolution (SR) methods, we use full images from DIV2K\citep{DIV2K} and Urban100\citep{huang2015single-urban100} as test datasets instead of their cropped versions. We retrain SR3~\citep{SR3} and SRDiff~\citep{li2022srdiff}  with DF2K~\citep{DIV2K} dataset to keep fair comparison. Additionally, we introduce two widely recognized diffusion-based Real-world SR methods, DiffBIR~\citep{lin2023diffbir} and SUPIR~\citep{yu2024scaling-SUPIR}, which are used to check the generalization capability of diffusion-based real-world image SR methods on synthetic datasets with bicubic degradation. 
(ii) For the comparisons of zero-shot DM-based SR methods, we follow the experimental setups of existing works, \ieno, DPS\citep{dps} and DDNM\citep{ddnm}, to process ImageNet1K~\citep{imagenet1k} and CelebA-HQ test~\citep{Celeba-hq} for consistent comparison. (iii) For the comparison of real-world image restoration, we follow StableSR \citep{wang2023exploiting-stable-sr} to select RealSR~\citep{cai2019toward-realsr}, DRealSR~\citep{wei2020component-drealsr}, and RealPhoto60 as test datasets with central cropping operation. 

The experimental results for supervised diffusion model-based IR models on 4$\times$ image super-resolution are listed in Table~\ref{tab:supervised_results}, which are tested on the DIV2K \citep{DIV2K} and Urban100 \citep{huang2015single-urban100} datasets with the original image. Among supervised DM-based SR methods, we can find that Resdiff \citep{shang2023resdiff} performs exceptionally well on PSNR, SSIM, LPIPS, and FID on DIV2K and Urban100, with a roughly 0.76dB improvement over other diffusion models in terms of PSNR. This is because Resdiff utilizes the diffusion model to generate residual information and uses pre-processed images for conditional generation, thereby ensuring the consistency of the restored image with the high-resolution image at the pixel level. In contrast, DiffBIR~\citep{lin2023diffbir} and SUPIR~\citep{yu2024scaling-SUPIR} are devoted to solving real-world image restoration, thereby achieving better perceptual quality in RealSR, DRealSR datasets as shown in Table~\ref{table:results-real}. For sampling efficiency, ResShit only needs 15 sampling steps, while others need more than 50 sampling steps. The generation of a single image by SRdiff takes 4.5 seconds, costing 84.22 GFlops and 13.2M parameters. This is because SRDiff encodes the low-resolution image into latent space, which reduces the dimension of processing. 

For zero-shot-based diffusion IR models, the quantitative comparisons are shown in Table~\ref{tab:supervised_imagenet_celeba}. We test six open-source diffusion models for 4x super-resolution on two datasets: ImageNet 1K \citep{imagenet1k} and CelebaHQ test \citep{liu2015deep-celeba}. From the table, we can observe that DDRM \citep{DDRMDRM} and DDNM \citep{ddnm} performed well on various metrics, followed by ILVR \citep{choi2021ilvr}. This is because DDRM and DDNM consider data consistency with the low-resolution image from a decomposition perspective, while ILVR only ensures low-frequency consistency. DPS \citep{dps} and GDP \citep{GDP} focus more on the perceptual quality of generated images. 
GDP~\citep{GDP} performs well in perceptual metrics on both datasets because it uses an empirical formula rather than Tweedie's formula for posterior estimation, resulting in some performance improvement relative to DPS. GDP also exhibits the fastest image generation speed, with the reverse sampling steps of 20. The reason for that is it does not need the complex SVD decomposition and computation in DDRM, thereby accelerating image generation. On the other hand, DPS performs corrections after every sampling step (around 1000 steps), making it unable to utilize the DDIM-based sampling method for acceleration. Therefore, generating a single image with DPS takes approximately 141.2 seconds.

By comparing GAN-based IR methods and Diffusion-based methods, we can find that: (i) Diffusion-based IR methods do not always outperform GAN-based IR methods for synthetic degradation. As shown in Table~\ref{tab:supervised_results} and \ref{tab:supervised_imagenet_celeba}, SeD~\citep{li2024sed} consistently achieve optimal reference-based perceptual quality metrics, including LPIPS and FID, which reveals the potential of Gan-based IR methods on synthetic degradation since they also have the advantage on efficiency; (ii) GAN-based IR methods and Diffusion-based IR methods are susceptible to poor pixel-wise consistency in general (\ieno, PSNR, and SSIM) since they are generation-based models. But, we can find some zero-shot DM-based methods in Table~\ref{tab:supervised_imagenet_celeba}, such as DDRM~\citep{DDRMDRM} and ~\citep{ddnm} achieve great pixel-wise consistency. The reason might be that decomposition-based methods can accurately obtain the inverse matrix for bicubic downsampling. Moreover, the utilized diffusion model for zero-shot IR has been pre-trained on face datasets; (iii) Diffusion-based RealSR methods outperform GAN-based SR methods by a large margin across all NR-IQA metrics for real-world degradation, as shown in Table~\ref{table:results-real}, This could be attributed to the fact that these diffusion-based methods are built upon pre-trained diffusion models (\egno, Stable Diffusion \citep{wang2023exploiting-stable-sr}) trained on large-scale natural image datasets. These models inherently possess strong priors that closely align with the real-world image distribution, enabling the super-resolution models fine-tuned on them to generate more realistic restoration results. Nevertheless, due to the nature of diffusion models, where images are sampled progressively from a Gaussian noise distribution, the stochastic generation process is difficult to constrain at the pixel level. This leads to inferior performance on objective metrics such as PSNR and SSIM compared to GAN-based methods, which can better preserve pixel-wise fidelity.

We also select seven state-of-the-art typical deterministic IR methods with various architectures for comparison, including transformer~\citep{dosovitskiy2020image-vit}, mamba~\citep{guo2024mambair} and CNNs~\citep{zhang2018image-RCAN} in Tables~\ref{tab:supervised_results} and~\ref{tab:supervised_imagenet_celeba}. These methods achieve superior performance on objective metrics such as PSNR and SSIM compared to GAN-based and diffusion-based methods. However, this strict pixel-wise fidelity constraint comes at the cost of perceptual quality, leading to inferior performance on subjective evaluation metrics compared to GAN-based and diffusion-based approaches.

\noindent\textbf{Results on Image Deblurring.}
We also evaluate five zero-shot DM-based IR methods on the Gaussian deblurring task using the ImageNet 1K \citep{imagenet1k} and CelebaHQ test \citep{liu2015deep-celeba} datasets. The experimental results are shown in Table 8 of the \textbf{Supplementary}. We can find that
DiffPIR \citep{zhu2023denoising-diffpir} and Dirac-DO \citep{fabian2023diracdiffusion} achieve competitive performance on PSNR and SSIM,  with an average improvement of 1.0 dB to 1.4 dB over DDRM \citep{DDRMDRM} and DDNM \citep{ddnm}. Moreover, Dirac-PO \citep{fabian2023diracdiffusion} and DiffPIR \citep{zhu2023denoising-diffpir} show superior performance on perceptual metrics. DPS \citep{dps} performs well on perceptual metrics, including LIPS and FID, but has a long generation time for each image. DiffPIR utilizes a plug-and-play sampling method and merges the DDIM sampling strategy to ensure both the fidelity and realism of generated images while speeding up the sampling process. Diracdiffion includes perception-optimized (PO) and distortion-optimized (DO) models, and employs an incremental reconstruction and early stopping method to achieve a perception-distortion trade-off. As a result, both models perform exceptionally well on distortion and perception metrics. Among all the models used, DDRM has the shortest sampling time, averaging under 10 seconds per image, as it only uses 20 sampling steps. In terms of model parameters, all zero-shot DM-based IR methods use a pre-trained model with 552.8M parameters for the ImageNet dataset and 126M parameters for the CelebA dataset.

\begin{table*}[htp]
\centering
\caption{Comparisons of some real-world restoration methods. Results are tested on three FR metrics: PSNR$\uparrow$, SSIM$\uparrow$, LPIPS$\downarrow$ and three NR metrics: MUSIQ$\uparrow$, ClipIQA$\uparrow$, ManIQA$\uparrow$.}
\setlength{\tabcolsep}{2pt}
\resizebox{1.0\textwidth}{!}{
\begin{tabular}{c|cccccc|cccccc|ccc|c}
\hline

\multirow{2}{*}{Models} & \multicolumn{6}{c|}{RealSR}                                                                          & \multicolumn{6}{c|}{DRealSR}                                                                          & \multicolumn{3}{c|}{RealPhoto60} & NFE \\ \cline{2-17} 

                        & PSNR           & SSIM           & LPIPS                    & MUSIQ          & ClipIQA  &   ManIQA     & PSNR           & SSIM           & LPIPS                     & MUSIQ          & ClipIQA    &ManIQA     & MUSIQ          & ClipIQA  & ManIQA & -       \\ \hline
\midrule
RealESRGAN~\citep{wang2021realesrgan} & 25.69 & 0.762 & 0.271 & 60.37 & 0.448 & 0.548 & 28.64 & 0.805 & 0.282 & 54.27 & 0.451 & 0.490 & 50.95 & 0.599 & 0.488&1\\
CAL-GAN~\citep{CAL-GAN} & 26.38 & 0.754 & 0.272 & 56.07 & 0.423 & 0.505 & 28.91 & 0.799 & 0.289 & 50.00 & 0.423 & 0.460 & 51.52 & 0.646 & \textbf{0.522}&1\\
SeD~\citep{li2024sed}& 25.56 & 0.754 & 0.271 & 61.03 & 0.540 & 0.546 & 28.35 & 0.794 & 0.285 & 55.75 & 0.530 & 0.500 & 57.14 & 0.690 & 0.519&1\\ 
\hline \midrule
CDC~\citep{wei2020component-drealsr}& \textbf{27.81} & 0.787 & 0.329 & 42.71 & 0.299 & 0.321 & \textbf{32.76} & \textbf{0.877} & 0.291 & 40.27 & 0.356 & 0.283 & 29.52 & 0.369 & 0.268 &1\\ 
SwinIR~\citep{swinIR-transformer} & 26.87 & \textbf{0.800} & 0.291 & 53.73 & 0.306 & 0.349 & 29.06 & 0.834 & 0.329 & 45.42 & 0.330 & 0.307 & 30.82 & 0.332 & 0.244 &1\\ 
MambaIR~\citep{guo2024mambair}  & 26.72 & 0.766 & 0.265 & 53.38 & 0.396 & 0.299 & 29.01 & 0.805 & 0.302 & 45.35 & 0.386 & 0.288 & 31.46 & 0.351 & 0.253 &1\\  

\hline\midrule
StableSR~\citep{wang2023exploiting-stable-sr} & 24.70 &0.707 & 0.300 & 65.88 & 0.623 & 0.428 & 28.15 & 0.752 & 0.328 & 58.51 & 0.636 & 0.387 & 62.24 & 0.651 & 0.361 & 200             \\
DiffBIR~\citep{lin2023diffbir}  & 25.07 & 0.640 & 0.395 & 64.85 & 0.638 & 0.615 & 26.53 & 0.630 & 0.495 & 61.19 & 0.634 & \textbf{0.592} & 64.45 & 0.689 & 0.418 &50\\ 
PASD~\citep{yang2023pixel-PASD}    & 24.29 & 0.663 & 0.343 & \textbf{68.69} & 0.659 & \textbf{0.649} & 27.00 & 0.708 & 0.393 & 61.81 & \textbf{0.677} & 0.585 & 63.28 & 0.712 & 0.477 &20\\ 
DiffIR~\citep{xia2023diffir} & 26.52 & 0.767 &\textbf{ 0.256}& 58.60 & 0.416 & 0.351 & 29.18 & 0.808 & \textbf{0.271} & 50.54 & 0.426 & 0.311 & 49.41 & 0.621 & 0.353 & 4 \\
ResShift~\citep{yue2023resshift} & 25.69 & 0.736 & 0.328 & 56.89 & 0.536 & 0.347 & 27.12 & 0.741 & 0.387 & 51.24 & 0.539 & 0.325 & 36.41 & 0.566 & 0.289 & 15\\
SUPIR~\citep{yu2024scaling-SUPIR}               & 23.47 & 0.537 & 0.518 & 67.06 & \textbf{0.663} & 0.641 & 23.76 & 0.563 & 0.659 & \textbf{65.56} & 0.592 & 0.581 &\textbf{67.04} & \textbf{0.746} & 0.515 &50\\ 
\hline
\end{tabular}
}
\label{table:results-real}
\end{table*}

\section{Challenges and Future Directions}
\label{sec:future}
Although recent researches have achieved remarkable progress on diffusion model-based IR, there are still some challenges to extending them to practical applications since their limited robustness, model complexity, running efficiency, interactability, personalization, and restoration capability. To further improve the development of image restoration, we summarize the primary challenges and propose the potential directions for solving them in this section. 

\subsection{Sampling Efficiency}
It is noteworthy that sampling efficiency is one typical challenge for the diffusion model, where the few sampling steps will cause limited generation fidelity. This inherent problem of the diffusion model damages the training and inference speed of image restoration. As shown in Table~\ref{tab:supervised_results}, SR3 \citep{SR3} needs to take about 50 seconds to restore one image with the size of $224\times 224$, which is largely slower than the existing IR methods.  previous works on diffusion models have attempted to improve the sampling efficiency from four perspectives: 1) modeling the diffusion process with a non-Markov Chain, such as DDIM \citep{DDIM}. 2) designing efficient ODE solvers, \egno, DPM-solver. 3) leveraging the knowledge distillation to reduce sampling steps \citep{zhou2023sparsefusion-distillation,salimans2022progressive-distillation}, and 4) introducing the cross-modality priors with condition mechanism \citep{abu2022adir,wang2023exploiting-stable-sr,liu2023improved}. Under the above progresses, the sampling steps for the diffusion model are largely reduced to $10\sim 20$ steps, which also serves for more fast image restoration. In particular, DDRM \citep{DDRMDRM} reduces the inference speed to 8 seconds for one $224\times 224$ of the image with the sampling strategy of DDIM \citep{DDIM}.

Despite that, the above strategies are not specific to image restoration tasks. Differently, considering the low-quality image in IR contains abundant structural and textual information, some works \citep{IR-SDE,welker2022driftrec,luo2023refusion,zhao2023partdiff} achieve image restoration by sampling from the low-quality image instead of pure noise, which obviates the extra sampling steps in the original DDPM. Among them, PartDiff \citep{zhao2023partdiff} utilizes an intermediate latent stage containing Low-resolution information as a proxy for the start of the inference process. Another work \citep{ma2023solving-ode} tries to accelerate the sampling speed of SR models by exploring the optimal boundary conditions for solving diffusion ODEs.  

Several one-step diffusion models have been proposed to enhance efficiency in various image restoration tasks with knowledge distillation. OSEDiff~\citep{wu2024oneOSEDiff} tackles real-world image super-resolution (Real-ISR) by leveraging the latent information embedded in the low-quality (LQ) image, using it directly as the starting point for diffusion instead of random Gaussian noise. By introducing variational score distillation, OSEDiff~\citep{wu2024oneOSEDiff} aligns the generative process with natural image priors, leading to a notable reduction in inference time compared to multi-step diffusion approaches. Another promising direction is Distribution Matching Distillation (DMD)~\citep{yin2024oneDMD}, which enforces a one-step generator to match the score distribution of the original multi-step diffusion model. Instead of modeling the noise-image correspondence, DMD directly learns the distribution mapping, demonstrating impressive improvements in both inference speed and sample quality. Similarly, SinSR~\citep{wang2023sinsr} focuses on the super-resolution task by distilling a deterministic mapping between the noisy input and the restored image. By refining the inference process into a single step, SinSR achieves up to 10$\times$ speedup while maintaining comparable perceptual quality to multi-step diffusion models.
These one-step diffusion methods present a compelling research direction for further improving the efficiency of diffusion-based image restoration. 
Regardless of the large process, it still poses a challenge to achieve real-time inference while preserving the generative diversity and flexibility of multi-step diffusion models. It will be an important direction to speed up the diffusion model-based IR methods by improving the sampling efficiency.

\subsection{Model Compression}
Model size is also a significant factor impacting the computational cost, limiting the real-time application of diffusion model-based image restoration (IR), such as mobile devices. In particular, DDPM \citep{DDPM} and SR3 \citep{SR3} have 113.7M and 155.3M parameters, respectively, substantially exceeding previous CNN-based \citep{zhang2018image-RCAN,jiang2021towards-jpeg-old4-cnn,li2020learning-lab1} or Transformer-based IR backbones \citep{swinIR-transformer,swinv2-transformer,wang2022uformer-transformer,li2022hst-lab3}. To mitigate this, diffusion model compression is a potential but under-explored research direction toward efficient IR. Model compression \citep{li2023modelcompressionSurvey}, with the goal of reducing computational cost while maintaining task performance, has achieved great breakthroughs from four perspectives: including 1) model pruning, with the aim of removing the unimportant parameters by estimating the importance score of each parameter, 2) model quantization targets for reducing the bit-depth of the floating-point parameters for storage or computing, 3) knowledge distillation is proposed to transfer the knowledge from the complex teacher model to the simple and efficient student model, and 4) low-rank decomposition is devoted to decomposing the parameter tensor into multiple low-rank tensors. On basis of these, some works have taken a step forward to investigate the model compression of the diffusion model. Kim \etal  \citep{kim2023architecturalKDDiffusion_MC} introduce the block-removed knowledge distillation for the diffusion model, which constructs the student model by removing some residual and attention blocks from the U-Net architecture.  Fang \etal \citep{fang2023structuralprunningDiffusion_MC} pinpoint that not all diffusion steps contribute to the generation process, and then, exploit partial diffusion steps to estimate the important and unimportant weights for parameter pruning with Taylor expansion. Additionally, there are pioneering works \citep{shang2023postQuantizationMC,wang2023towardsQuantizationMC} that take the model quantization for the diffusion model to accelerate the sampling process. Although such progress, few works explore how to design the model compression for diffusion model-based IR, which is expected to be developed for real-time application. 

\subsection{Distortion Simulation and Estimation}
Real-world/blind IR is a challenging yet significant task, with the goal of addressing the unknown and intricate degradations encountered in the real world. Unlike synthetic degradation, where the distortion is predefined and paired training samples are available, collecting the paired real-world distorted/clean pairs is non-trivial, thereby preventing the training with supervised learning. To address this limitation, unsupervised learning has been introduced to leverage unpaired real-world distorted/clean images. However, this learning paradigm usually yields an unsatisfied texture consistency between the restored image and the low-quality image. In contrast, distortion simulation serves as another efficient strategy to maintain supervised learning by simulating real-world degradations. Typically, RealESGRAN \citep{wang2021realesrgan} and SR3+ \citep{SR3+} are the representative work to explore the hand-crafted second-order degradations for Real-world IR. Notwithstanding, the hand-crafted distortion simulation is hard to cover all the degradations in the real world. To mitigate this, some works are inspired by domain translation, and introduce the GAN/diffusion model to translate the synthetic distorted image to the real-world one or translate the real-world distorted image to the synthetic one. The former intends to simulate the real-world training pairs for supervised learning \citep{yang2023synthesizing}, while the latter aims to directly utilize the IR network trained with synthetic images \citep{wei2023raindiffusion}. 

Another crucial challenge for real-world/blind IR stems from distortion estimation, which involves explicitly/implicitly identifying the distortion types or levels. In this work, we summarize the utilization of distortion estimation from two perspectives: 1) distortion adaptive learning and 2) solving the inverse problem in IR. From the first perspective, a notable example is kernel prediction for blind IR \citep{blinddps}, where the estimated kernel/representation is used to guide the adaptation of the pre-trained IR model to the unknown degradation. Inspired by this, if we can estimate the distortion type or degree in an explicit/implicit manner, we can achieve the unified IR framework based on distortion adaptive learning. For the second perspective, as stated in Sec.~\ref{sec:zero-shot}, lots of zero-shot diffusion model-based IR methods are based on the modelling of linear reverse problem. This poses the requirements for the identifying of degradation mode, which is necessary for the consistency constrain in diffusion model. Therefore, most of them are devoted into the synthetic distortions since the distortion mode in real world are hard to be identify. One distortion estimation technique is urgently developed to extend the zero-shot diffusion model-based IR to the real-world application.

\subsection{Distortion Invariant Learning}
Recently, we have witnessed the fast evolution of diffusion model-based IR for specific degradation. However, it inevitably suffers from unsatisfied robustness when applied to unseen distortion types and degrees. This raises a foundational question: How to achieve consistent image restoration across diverse distortion types and levels? To achieve this, we propose one direction termed distortion invariant learning (DIL) \citep{li2023learningDIL}, aiming to enable the IR model to be generalized to unknown and diverse degradations. The principle of DIL is to learn the representation that is invariant under various degradation modes and preserve enough structure and textual information for reconstruction.   

Inspired by domain generalization (DG) \citep{zhou2022domain-dg1,gulrajani2020search-dg2,li2018learning-dg3,wang2022generalizing-dg4}, we can present some potential methods to achieve DIL for IR by regarding each distortion mode as one domain. In the DG field,  there are three typical methods to learn the domain invariant feature, including domain alignment \citep{9412797-alignment1,Li_2018_ECCV-alignment3,Shao_2019_CVPR-alignment4}, data augmentation \citep{xu2020robust-augmentation4,mancini2020towards-augmentation1}, and meta-learning \citep{wang2020meta-meta1,jia2020single-meta2,rahman2020correlation-meta3,albuquerque2019generalizing-meta4}. In particular, domain alignment aims to align the representations of source and target domains through minimizing contrastive loss \citep{wang2021understanding-contrastive-loss}, Maximum Mean Discrepancy (MMD), or adversarial learning, etc.   Data augmentation is exploited to extend the domain diversity and consistency, which enables the model to obtain the domain-invariant capability. Meta-learning aims to learn the domain invariant representation by aligning the gradient between different domains, which is from the optimization perspective.  
By regarding the distortion mode as one specific domain, we can obtain several strategies to achieve distortion invariant learning: 1) we can exploit the encoder-decoder architecture for IR and align the representations from different distorted images before the decoder. 2) the second strategy can learn the distortion invariant representation from distortion augmentation, \ieno, simulating various distortions in the real world as much as possible. 3) optimize the empirical risk minimization in IR with meta-learning like \citep{li2023learningDIL}. 

For diffusion model based image restoration, the model is usually composed of two components, noise predictor and condition module. Therefore, we can achieve distortion invariant learning from two perspectives: 1) learning the distortion invariant noise predictor and 2) distortion invariant condition. Obviously, once we implement the distortion invariant condition, we can leave the noise predictor invariant in supervised IR or exploit the pre-trained diffusion model in zero-shot IR. Based on this, some pioneering works attempt to redesign the condition module to achieve the distortion invariant condition, \egno, DifFace \citep{yue2022difface} and DR2 \citep{wang2023dr2}. Notably, the distortion invariant condition also relies on distortion invariant learning for better conditions, which still entails lots of effort for evolution in future work.

\subsection{Framework Design}
As the foundation of image restoration, how to design an effective and powerful IR framework is an ongoing and significant question. We can notice that the most recent diffusion model-based IR methods \citep{SR3,ddnm,DDRMDRM,dps,saharia2022palette,iigdm,GDP,li2022srdiff} are designed based on the U-Net architecture from DDPM \citep{DDPM}, and pursue better frameworks from three perspectives, \ieno, the condition strategy \citep{dps,DDRMDRM,repaint,sr3_swinfir-transformer,IR-SDE}, generation space\citep{LDM,luo2023refusion,xia2023diffir}, noise predictor \citep{xia2023diffir}, respectively. The condition of the diffusion model in IR aims to introduce the structure and textual information from low-quality images. In the early work, SR3 directly select low-quality images as a condition with the concatenation. To improve the condition, some works \citep{niu2023cdpmsr,guo2023shadowdiffusion,IDM,jin2022shadowdiffusion-driven,shang2023resdiff} improve the condition by designing pre-processing networks, such as feature extractors and pre-trained restoration networks. From the generation space, the framework is usually designed from four spaces, including image space, residual space, latent space, and frequency space. Among them, pixel-wise space can preserve more spatial structure and textual information, which can generate high-quality images \citep{niu2023cdpmsr,shang2023resdiff,IDM} or residuals \citep{li2022srdiff,shang2023resdiff,liu2023residual-RDDM,yue2023resshift,liu2023textdiff}, while owning higher computational cost and parameters. In contrast, latent space generation requires fewer computational costs. However, one well-designed encoder and decoder are crucial for latent space generation to make the trade-off between efficiency and fidelity. Frequency space has been widely applied in diffusion model-based image rest-oration \citep{moser2023waving,xu2023stage,mirza2023learning-fourier-frequency-FDB,wang2023frequency-dehazy,guan2023correlated-frequency,luo2023image-frequency-feng}, including wavelet transform, Fourier transform, etc. Compared with image-wise space, frequency space is more good at capturing the global contextual information, where the low-frequency refers to the structural information and the high-frequency represents the texture and style information. These methods can bring more detailed and high-quality image restoration.
Following the DDPM \citep{DDPM}, in most works, the noise predictor is based on the U-Net architecture. For supervised diffusion model-based IR, the modification of the noise predictor is usually achieved by increasing the number of residual blocks in the U-Net or adjusting the channel multipliers at different resolutions, such as SR3. Few works are devoted to investigating how to design brand-new architectures based on transformers for the noise predictor in diffusion model-based IR. Furthermore, how to design the unified and foundational architecture like the painter \citep{wang2023images-painter} for diffusion model-based IR tasks is urgently required to be explored.

\subsection{All-in-one Restoration}
All-in-one image restoration aims to simultaneously address multiple diverse degradations within a single IR model, which has attracted lots of attention in recent years. 
The key challenge lies in enabling diffusion-based IR models to adapt to various degradations while effectively leveraging shared restoration knowledge across different degradation types. The typical work is DA-CLIP \citep{luo2023controlling-DACLIP}, which introduces the degradation-aware CLIP to extract the degradation-aware embedding as the condition for diffusion models. In contrast, MPerceiver \citep{ai2023multimodal-mperceriver} extracts the holistic representation and multi-scale detail representation with the textual and visual branches of CLIP as the condition. Diff-Restore \citep{zhang2024diffrestore} also leverages the CLIP model to extract the degradation and semantic embedding but adopts a distinct conditioning strategy. For unknown hybrid degradation, AutoDIR \citep{jiang2023autodir} introduce the quality assessment model to identify which degradation should be removed and utilizes multiple steps for restoration. UIR-LoRA \citep{hu2022lora} leverages low-rank adaptation (LoRA) to encode task embeddings efficiently. Then it introduces a dynamic weighting strategy based on CLIP similarity between image features and predefined distortion text embeddings. In contrast, GenDeg \citep{rajagopalan2024gendeg} utilizes diffusion models to generate diverse distortions from clean images, expanding training data for all-in-one diffusion-based IR models, which further improves the performance of multiple IR tasks. Although the above works have achieved impressive progress in diffusion-based all-in-one image restoration, a new all-in-one diffusion paradigm needs to be developed based on efficient architecture and new multi-task learning paradigm like Mixtures-of-Experts (MoE).   

\tct{\subsection{Restoration Systems with LLM/MLLM-based Agents.}  
Recent studies have proposed agent-based image restoration frameworks to enhance system intelligence, flexibility, and human interactivity. Leveraging the powerful understanding and reasoning capabilities of large language models (LLMs) and large multimodal models (LMMs), these approaches introduce language-driven agents into the restoration pipeline. For instance, AgenticIR~\citep{zhu2024intelligentAgenticIR}, hybridAgents~\cite{li2025hybrid} and RestoreAgent~\citep{chen2024restoreagent} employ frozen LLMs or vision-language models to analyze input degradations and dynamically select traditional restoration tools (e.g., SwinIR~\citep{liang2021swinir}, FBCNN~\citep{jiang2021towards-fbcnn}, Restormer~\citep{zamir2022restormer_deblur-transformer}).  
Building upon this foundation, recent works such as MAIR~\citep{jiang2025multiMAIR} and 4K-Agent~\citep{Zuo4KAgent} integrate powerful diffusion-based models—such as DiffBIR~\citep{lin2023diffbir}, IR-SDE~\citep{IR-SDE}, and DPS~\citep{dps}—into the agent's toolset. In particular, 4K-Agent employs a frozen LLM/MLLM to perform multi-round reasoning and sequential tool invocation, treating diffusion models as central, not auxiliary, components. This design achieves superior performance, especially under compound, real-world degradations.  
These developments reveal a clear trajectory in image restoration: from static, task-specific models to intelligent agentic systems, and further to fully integrated multimodal frameworks that emphasize interpretability, generalization, and controllability. As such, the development of LLM/MLLM-based restoration age-nts is emerging as a highly promising direction in the field, where the diffusion-based IR models are expected to serve as core generative priors and cross-modality generation tools in this system. }

\tct{\subsection{Vision-Language Prompting for Diffusion-Based Restoration.}
The integration of vision-language models (VLMs) into restoration workflows enables high-level semantic guidance for diffusion models. PASD~\citep{yang2023pixel-PASD} and Upscale-A-Video~\citep{zhou2024upscaleavideo} utilize frozen VLMs like BLIP-2~\citep{li2023blip2} and LLaVA~\citep{liu2024visual-LLaVA} to extract text prompts from degraded images, which are then used to condition diffusion upsamplers. Diff-Restorer~\citep{zhang2024diffrestore} takes this further by injecting semantic and degradation embeddings into the diffusion process. However, since standard VLMs may be misled by visual noise, recent methods like SeeSR~\citep{wu2024seesr} and SUPIR~\citep{yu2024scaling-SUPIR} adapt VLMs to be degradation-aware by retraining (\egno, RAM-based tagging in SeeSR) or introducing lightweight artifact removers. These methods demonstrate the semantic understanding with VLMs, and the VLM-based condition strategies will empower high-fidelity restoration, which is a potential direction for diffusion-based IR.}

\tct{\subsection{Human-in-the-Loop Optimization and Preference Alignment.}
Most existing image restoration (IR) models are optimized using L1, perceptual, or similar handcrafted losses, which may not accurately reflect human preferences or perceptual quality. To bridge this gap, recent works have begun to explore diffusion-based generation with human feedback.  For instance, PrefPaint~\citep{liu2024prefpaint} learns a reward model from human-anno-tated preference data and fine-tunes a pre-trained diffusion backbone via reinforcement learning. InstructRestore~\citep{liu2025instructrestore} enables region-specific, instruction-driven restoration by allowing users to specify both spatial regions and restoration strength using natural language. DSPO~\citep{cai2025dspo} introduces patch-level semantic preference optimization for real-world super-resolution, aligning local instance appearance with human feedback.  
In the future, the human-in-the-loop optimization is expected to open up an exciting direction, such as reinforcement learning with human feedback (RLHF), synthetic preference data generation via diffusion sampling, and fine-grained/region-aware diffusion-based IR restoration under natural-language control.}

\section{Conclusions}
\label{sec:conclusion}
This work presents a comprehensive review of recent popular diffusion models for IR, excavating their substantial generative capability to enhance structure and texture restoration. Initially, we illustrate the definition and evolution of the diffusion model. Subsequently, we provide a systematic categorization of existing works from the perspectives of training strategy and degradation scenario. Concretely, we group existing works into three prominent flows: supervised DM-based IR, zero-shot DM-based IR, and DM-based blind/real-world IR. For each flow, we provide the fine-grained taxonomy based on the techniques and delicately describe their advantages and disadvantages. For evaluation, we summarize the commonly used datasets and evaluation metrics for DM-based IR. And we compare the open-sourced SOTA methods with distortion and perceptual metrics on three typical tasks, including image SR, deblurring, and inpainting. To overcome the potential challenges in DM-based IR, we highlight \tct{nine} potential directions expected to be explored in the future.

\section*{Acknowledgements}
This work was partly supported by the NSFC under Grant No. 
623B2098, 62371434, 62021001, and the China Postdoctoral Science Foundation-Anhui Joint Support Program under Grant Number 2024T017AH.

\noindent We would like to appreciate Prof. Chen Chang
Loy sincerely for his valuable suggestions and help on this paper, which greatly improved the quality
of our work.

\section*{Data Availability Statements}
All datasets and codes used in this work are publicly available and are cited in this work. We have summarized them in~\url{https://drive.google.com/file/d/1z_k1O6yFznDQ1OuyJuazK-YOauA8XHYt/view?usp=drive_link}, which can be utilized to reproduce all experimental results of this work.

\bibliographystyle{unsrt}     
\bibliography{egbib}   
\renewcommand{\thesection}{\Alph{section}}
\renewcommand{\aboverulesep}{0pt}
\renewcommand{\belowrulesep}{0pt}
\begin{figure*}[htp]
	\centering 
	\includegraphics[width=1\linewidth]{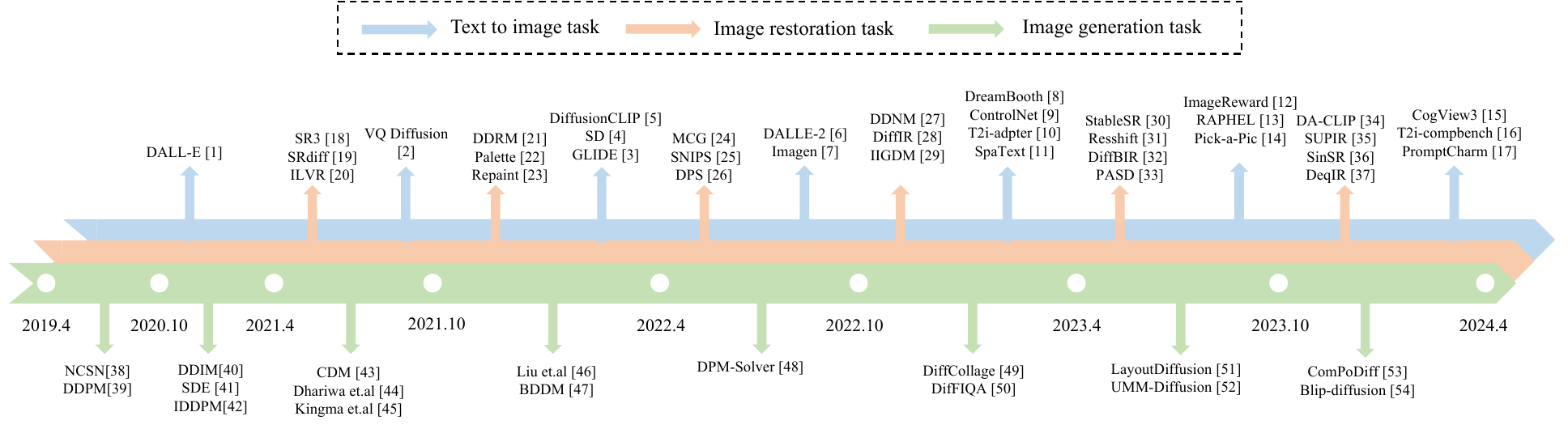}
	\caption{Timeline of Diffusion Model Development in Text-to-Image, Image Restoration, and Image Generation Applications (Text-to-Image in Blue, Image Restoration in Orange, Image generation in Green).}
	\label{fig:timeline}
\end{figure*}
\setcounter{section}{0}
\section{Timeline}

Fig.~\ref{fig:timeline} presents a chronological overview of representative works in text-to-image synthesis \citep{ramesh2021zero-t2i6,gu2022vector-vqdiffusion-backbone,nichol2021glide-t2i5,LDM,kim2022diffusionclip-t2i3,ramesh2022hierarchical-dalle2,saharia2022photorealistic-t2i2,ruiz2023dreambooth-t2i1,zhang2023adding-ControlNet,mou2023t2i-app9,avrahami2023spatext-t2i4,xu2024imagereward,xue2024raphael,kirstain2024pick-a-pic,zheng2024cogview3,huang2023t2i-combench,wang2024promptcharm}, image restoration \citep{SR3,li2022srdiff,choi2021ilvr,DDRMDRM,saharia2022palette,repaint,mcg,kawar2021snips,dps,ddnm,xia2023diffir,iigdm,wang2023exploiting-stable-sr,yue2023resshift,lin2023diffbir,yang2023pixel-PASD,luo2023controlling-DACLIP,yu2024scaling-SUPIR,wang2023sinsr,cao2023deep-DeqIR} and general image generation \citep{NCSN,DDPM,DDIM,SDE,nichol2021improved,CDM,guided-diffusion,kingma2021variational,liu2023more,lam2021bilateral-optimization,DPM-solver,zhang2023diffcollage-app10,babnik2023diffiqa-app11,zheng2023layoutdiffusion,ma2023unified-UMM-diffusion,gu2023compodiff,li2024blip-diffusion} using diffusion models.

\section{Background} 
\subsection{Noise Conditioned Score Networks (NCSNs)}
The purpose of generative models is to learn the probabilistic distribution of the target data. Different from previous likelihood-based \citep{dinh2014nice-likelihood,ho2019flow++-likelihood,chen2018neural-likelihood,grathwohl2018ffjord-likelihood,lee2022autoregressive-likelihood} and GAN-based \citep{karras2019style-gan1,berthelot2017began-gan2,huang2023enhanced-gan9} methods, NCSNs aim to estimate the data distribution from the gradient of the log-density function, \ieno, the score function $\nabla_x \log p(x)$, which guides the sampling progressively forward to the direction of the center of the data distribution. 
Concretely, NCSNs predict the score function of original data with a neural network $s_{\theta}$ parameterized by $\theta$. To avoid the resulting distribution collapses to the low dimensional manifold and the inaccurate score estimation in the low-density region, the annealed Langevian dynamics \citep{Langevin1, Langevin2} is designed for the score-based generative model, where the predefined noises with the monotonically decreasing levels ${\sigma}_{i=1}^L$ are introduced to perturb the data. The original sampling process with Langevin dynamics can be represented as:
\begin{equation}
    \Tilde{x}_{t}=\Tilde{x}_{t-1}+\frac{\epsilon}{2}\nabla_{\Tilde{x}}\log p(\Tilde{x}_{t-1})+\sqrt{\epsilon}z_{t},
    \label{eq:langevin_dynamic}
\end{equation}
where $z_t$ is the random normal Gaussian noise at time step $t$, $\epsilon$ is the fixed step size. When time step $T\xrightarrow[]{}\infty$ and $\epsilon\xrightarrow[]{}0$, the distribution $p(\Tilde{x}_T)$ will equal to original data distribution $p(x)$. After adding the noise with level $\sigma$, the perturbed distribution will be $q_\sigma(\Tilde{x}) \triangleq \int p(x) \mathcal{N}(\Tilde{x}| x, \sigma^2 I)dx$. The noise conditional score network (NCSN) 
  can be optimized toward $s_\theta(\Tilde{x}, \sigma)=-\nabla_{\Tilde{x}}\log q_\sigma(\Tilde{x})$ with the noise score matching objectives as:
\begin{equation}
    \mathcal{L}(\theta,\sigma)=\frac{1}{2}E_{p(x)}E_{\Tilde{x}\sim\mathcal{N}(x,\sigma^2I)}[\Vert s_{\theta}(\Tilde{x},\sigma)+\frac{\Tilde{x}-x}{\sigma^2}\Vert_{2}^2]
\end{equation}
\subsection{Improvements with Diffusion Model}
Based on the foundational diffusion models described above, numerous works are developed to further enhance the diffusion model from the perspectives of optimization strategy, sampling efficiency, model architecture, conditional strategy, etc. 

\vspace{0.6em}

\noindent\textbf{Optimization Strategies.}
To enhance the stability and performance of the diffusion model, several works \citep{nichol2021improved,song2020improved-NCSN,kingma2021variational,lam2021bilateral-optimization,chen2023importance-optimization} have explored the optimization of the variance/noise schedule in both forward and backward processes. Notably, the noise schedule in the forward process controls the perturbation degrees at each step, which is particularly crucial for the inverse process. As the representative work, DDPM \citep{DDPM}, employs the straightforward linear noise schedule for the diffusion process. But this approach often leads to sub-optimal results, especially for the low-resolution image generation \citep{nichol2021improved}. To mitigate this, IDDPM \citep{nichol2021improved} introduces a cosine noise schedule to obviate the negative effects of rapid noise accumulation at the early perturbation stages. Concurrently, Diederik~\etal \citep{kingma2021variational} parameterize the noise schedule with a monotonic neural network and jointly optimize it with the diffusion model. In general, the variance schedule in the reverse process is fixed and computed with the noise schedule in the forward process. However, IDDPM \citep{nichol2021improved} found learning variance can further improve the log-likelihood and thus employed the variance schedule as the learnable linear interpolation of variances in both the forward and reverse processes. In contrast, Analytic-DPM \citep{bao2022analytic-dpm} derives the optimal variances trajectory through analytic estimation, thereby improving the log-likelihood of various DPMs. Depart from the above methods, Jolicoeur-Martineau~\etal \citep{jolicoeur2020adversarial} propose a brand-new sampling procedure, \ieno,  Consistent Annealed Sampling,  which is more stable for the diffusion model than the annealed Langevin method.

\vspace{0.6em}

\noindent\textbf{Sampling Efficiency.}
The generative quality of the diffusion model is critically contingent on a significant number of sampling steps, thereby posing challenges to its efficiency in practical application. To mitigate this, four principal lines of work have been proposed to accelerate the sampling process. 1) The first line involves the handcrafted sampling strategies related to ODE \citep{DDIM,DPM-solver,zhang2022fast,lu2022dpm-ode,karras2022elucidating-ode,ma2023solving-ode}. For instance, DDIM \citep{DDIM} introduces the non-Markovian chain in the forward process, enabling the diffusion model to achieve sampling over an arbitrary number of steps. In contrast, DPM-solver \citep{DPM-solver} strives for a fast ODE solver by analytically computing the linear part of the ODE solution rather than utilizing the black-box ODE solvers. 
Ma~\etal \citep{ma2023solving-ode} propose to sample high qualiy SR images from pretrained SR diffusion models \citep{SR3,li2022srdiff} through solving ODE with optimal boundary conditions. Consequently, this ODE-based method substantially narrows down the sampling steps required to generate high-quality images, limiting them to an acceptable range of 10 to 20. 2) The second paradigm is to revise the diffusion process \citep{lyu2022accelerating-truncation,zheng2022truncated,zhao2023partdiff}. As the representative work, Lou \etal \citep{lyu2022accelerating-truncation} propose to truncate the diffusion process with an early stopping mechanism, and initiate the sampling from the non-Gaussian distribution, which is generated with a pre-trained VAE/GAN model using existing VAE or GAN-based methods \citep{yuan2018unsupervised-ganr1,li2022multi-ganr4,yang2021gan-ganr8,chan2021glean-ganr7,roich2022pivotal-ganr5,pan2021exploiting-ganr9,tampubolon2021snpe-ganr3,dinh2022hyperinverter-ganr6,wang2020transformation-ganr2,karras2019style-gan1,berthelot2017began-gan2,li2017perceptual-gan3,bai2018sod-gan4,zhang2019self-gan5,zhu2017unpaired-gan6,kim2019u-gan7,zhang2022styleswin-gan8,huang2023enhanced-gan9}.  Zhao~\etal \citep{zhao2023partdiff} propose PartDiff to diffuse images to an intermediate latent state which contains information about low-resolution image. During the inference stage, they perform only part of denoising steps from this intermediate state, allowing for faster sampling of SR images. 3) In the third strategy, knowledge distillation is employed to transfer the generation capability from multiple sampling steps to a few sampling steps \citep{luhman2021knowledge-distillation1,meng2023distillation2,salimans2022progressive-distillation,zhou2023sparsefusion-distillation}. 4) The final approach leverages the condition strategy \citep{zhang2023adding-ControlNet,zheng2022fast-condition1,huang2022fastdiff-condition2,LDM,wang2023patch-condition3} to embed the generative prior, thereby optimizing the sampling efficiency.

\vspace{0.6em}

\noindent\textbf{Model Architecture.}
Diffusion models predominantly employ two kinds of architectures, \ieno, CNN-based U-Net, and Transformer-based model. Notably, 
U-Net architecture is preferred in early studies for noise/score prediction \citep{NCSN,DDPM}, benefiting from its capacity for resolution preservation and eliminating the resource cost through the multi-grained downsampled feature space. Following these, a series of efforts have been made in subsequent works to refine the U-Net architecture by incorporating cross attention module \citep{LDM}, group normalization \citep{guided-diffusion,DDPM}, multi-head attention \citep{song2020improved-NCSN,SDE,nichol2021improved}, and position encoding \citep{DDPM}. Recently, the transformer has demonstrated its proficiency in modeling long-range dependencies and unifying the different modalities \citep{luo2020univl-multimodal3,sun2019videobert-multimodal1,lu2019vilbert-multimodal2,yang2022vision-multimodal9,shi2022learning-multimodal8,sun2019learning-multimodal5,zhang2022glipv2-multimodal7,nagrani2021attention-multimodal4}. Consequently, some text-to-image works \citep{bao2023all,sheynin2022knn-vit-backbone1,tang2022improved-vit-backbone2,yang2022your-genvit-backbone,gu2022vector-vqdiffusion-backbone,li2022swinv2-backbone,chai2023layoutdm-backbone,pan20232d-backbone} explore to use transformer backbones \citep{sr_transformer1,sr2_transformer2,sr2_transformer3,swinIR-transformer,swinv2-transformer,sr3_swinfir-transformer,wang2022uformer-transformer,vaswani2017attention-transformer,liu2022dnt-denoise-new2-transformer,zamir2022restormer_deblur-transformer,tsai2022stripformer_deblur-transformer,jiang2022learning-jpeg-new1-transformer,fan2022sunet-denoise-new1-transformer,chen2023learning-sparse-transformer}, such as ViT \citep{dosovitskiy2020image-vit}, Swinv2 \citep{li2022swinv2-backbone}, to replace original U-Net used in most CNN-based methods \citep{dncnn-8,SR2——VDSR-cnn,krizhevsky2017imagenet-cnn1,jiang2021towards-jpeg-old4-cnn,SR3_EDSR-cnn,SR1_SRCNN-cnn,kim2023mssnet-deblur-cnn,nimisha2017blur-old3-cnn} to predict noise in the reverse process, where the time step $t$ and other conditions are fed to the transformer through adaptive layer normalization \citep{yang2022your-genvit-backbone,sheynin2022knn-vit-backbone1,gu2022vector-vqdiffusion-backbone} or cross attention \citep{bao2023all}.
\vspace{0.6em}

\noindent\textbf{Condition Strategy.}
 One effective condition strategy is crucial for the functionality of the diffusion model in the conditional generation. This has motivated numerous works to explore efficient and potent conditional mechanisms for it. Nicol \etal \citep{guided-diffusion} innovatively train an auxiliary classifier to direct the diffusion model, using its gradient to guide image generation toward specific semantics. Another popular line \citep{liu2023more,ho2022classifier-free} introduces the condition to the score estimation/noise prediction model in a classifier-free manner, such as GLIDE \citep{nichol2021glide-t2i5}, and DALLE-2 \citep{ramesh2022hierarchical-dalle2}. To preserve its unconditional generation capability, a null token $\emptyset$ is exploited to replace the condition $\theta(x, c)$ in the diffusion model, enabling $\theta(x)=\theta(x,\emptyset)$, where $c$ denotes the condition, such as textual feature. Thanks to its advantages, a series of works introduce the classifier-free condition to text-to-image tasks \citep{ruiz2023dreambooth-t2i1,saharia2022photorealistic-t2i2,kim2022diffusionclip-t2i3,avrahami2023spatext-t2i4,nichol2021glide-t2i5,ramesh2021zero-t2i6}. Furthermore, apart from the class labels and textual prompts, the diffusion model can integrate other modal conditions, \egno, images, segmentation maps, latent features, which greatly facilitates its application in various requirements, such as Stable Diffusion \citep{LDM}, ControlNet \citep{zhang2023adding-ControlNet}. 
 
\section{Methods}
The supervised and zero-shot diffusion based image restoration models are listed in Table 1 and Table 2, respectively. We also clarify the diffusion-based IR methods in Fig. 2 of our manuscript with the taxonomy of degradation types as follows. 

\textbf{(i) Diffusion-based Image SR:} Diffusion models have significantly advanced image super-resolution (SR) in terms of perceptual quality. As pioneering works, SR3 \citep{SR3} and SRDiff \citep{li2022srdiff} were the first to introduce diffusion models into the SR task. SR3 learns the reverse process directly in the image space, whereas SRDiff operates in the residual space to accelerate the process. Following these approaches, SR-SDE adopts the stochastic differential equation (SDE) diffusion paradigm to replace the denoising diffusion probabilistic model (DDPM) in SR3. However, low-quality images introduce degradation in the diffusion model through direct concatenation, which affects generation quality. To mitigate this, CDPMSR \citep{niu2023cdpmsr} leverages existing super-resolution models, such as RCAN \citep{zhang2018image-RCAN}, SwinIR \citep{swinIR-transformer}, and EDSR \citep{EDSR-9}, to enhance low-quality images. This preprocessing step provides a higher-quality and more reliable condition for the diffusion model. Additionally, CDPMSR \citep{niu2023cdpmsr} replaces stochastic sampling in the reverse process with a deterministic denoising process, leading to superior image quality and faster inference. Notably, the pre-processed reference image serves not only as an enhanced conditioning input but also as an initially well-restored image. ResDiff \citep{shang2023resdiff} takes a different approach by utilizing a pre-trained convolutional neural network (CNN) to generate a low-frequency content-rich image as an initial restoration. It then employs a conditional diffusion model to refine the residuals between the pre-processed distorted image and its clean counterpart. To enable continuous image super-resolution for arbitrary resolutions, IDM \citep{IDM} integrates an implicit neural representation with a denoising diffusion model in an end-to-end framework. The implicit neural representation is utilized in the decoding process to learn a continuous-resolution representation. Additionally, IDM \citep{IDM} introduces a scale-adaptive conditioning mechanism, consisting of a low-resolution (LR) conditioning network and a scaling factor, to enhance reconstruction fidelity across different scales. SAM-DiffSR \citep{wang2024sam-diffsr} introduces the Segment Anything Model (SAM) to segment images and leverage segmentation maps as conditions. This enables region-aware semantic structure adjustments of noise during the diffusion process, refining generation results. However, the aforementioned approaches require retraining the diffusion model.

With the advent of efficient transfer learning, recent works have begun to address this limitation by fixing the pre-trained diffusion model and employing techniques such as adaptor \citep{wang2023exploiting-stable-sr}, and ControlNet \citep{lin2023diffbir}. These methods can preserve the strong texture generation capability of diffusion model, making diffusion-based SR more practical and training-friendly. Pioneering work is StableSR, where a time-aware encoder is trained to extract information from low-quality images and incorporate the learned features into pretrained stable diffusion(trained on LAION-5B \citep{schuhmann2022laion-5B}). On the other hand, DiffBIR \citep{lin2023diffbir} and PASD \citep{yang2023pixel-PASD} introduce
conditions by training a UNet encoder using zero convolutions, which bears similarities to ControlNet \citep{zhang2023adding-ControlNet}. Another significant advancement
in this area is SUPIR \citep{yu2024scaling-SUPIR}, which builds
upon a pre-trained model based on SDXL \citep{podell2023sdxl}. It extracts textual prompts through LLaVA \citep{liu2024visual-LLaVA} and introduces distorted image information using a trimmed Controlnet. 
(Please see the Section 3.3.6 of our manuscript). In contrast, XPSR \citep{qu2024xpsr} explores the integration of high-level and low-level semantics extracted from multimodal large language models (MLLMs). It incorporates these features into the diffusion model through a control encoder architecture, further enhancing the perception of degradation and semantics.

\textbf{(ii) Diffusion-based Unified Model for Multiple Single IR Tasks.}
Several diffusion-based image restoration (IR) methods in Fig. 2 design the unified framework for various independent tasks, such as super-resolution, deblurring, inpainting, draining, dehazing, desnowing.
 These works can be roughly divided into two categories, including supervised DM-based IR works and zero-shot DM-based works.

Supervised DM-based IR methods are typically validated across different tasks and datasets, which is hard to align for comparison. As an early work, Palette \citep{saharia2022palette} formulates image restoration as an image-to-image translation problem. It highlights the importance of the self-attention mechanism within U-Net for diffusion models, achieving remarkable performance on tasks such as image colorization, inpainting, uncropping, and JPEG artifact removal. To improve stability and accelerate the sampling process, IR-SDE \citep{IR-SDE} modifies the forward process using a mean-reverting stochastic differential equation (SDE) to model the unified degradation process in IR:
\begin{equation}
    dx = \theta_t(\mu - x) dt + \sigma_{tdw},
\end{equation}
where $\theta_t$ and $\sigma_t$ are time-dependent parameters, $µ$ denotes the distorted image, and $x$ represents its corresponding high-quality counterpart.  By leveraging the mean-reverting SDE, IR-SDE effectively models both the degradation and restoration processes through modified forward and reverse processes, avoiding generation from pure noise and enhancing restoration quality. Building on IR-SDE, Refusion \citep{luo2023refusion} further optimizes network architecture, noise levels, and denoising steps. To reduce computational costs, it introduces a U-Net compression strategy, enabling efficient sampling in the latent space. In contrast, InDI \citep{INDI} proposes a continuous diffusion process:
\begin{equation}
    x_t = (1-t)x + ty,
\end{equation}
where $x$ and $y$ are the high-quality images and their
corresponding low-quality counterparts. This formulation can be interpreted as step-wise interpolation between high- and low-quality images over time step $t$. By decomposing the single-step prediction of supervised IR into multiple smaller steps, InDI \citep{INDI} effectively mitigates the regression-to-the-mean effect commonly observed in conventional supervised IR approaches. Unlike traditional diffusion processes in the image space.  Resshift \citep{yue2023resshift} introduces a novel Markov chain that shifts residuals between high-quality and low-quality images. This residual-shifting technique achieves comparable performance even with a significantly reduced number of sampling steps, as few as 15. In contrast to methods that directly generate restored images with a diffusion model, DiffIR \citep{xia2023diffir} utilizes a diffusion model to learn degradation priors, incorporating a conditioned transformer to restore images based on these priors. In contrast, C2F-DFT \citep{wang2023learning-C2F-DFT} adopts a transformer-based diffusion model and introduces a coarse-to-fine training strategy to enhance restoration quality.

Another type of DM-based IR works regarding the image restoration task as the linear or non-linear inverse problem. \textbf{The details can be found in Section 3.4.2 and 3.4.3 of our manuscript, which are all unified frameworks for diverse image restoration tasks except for DRUS \citep{zhang2023ultrasound-drus}.}

\textbf{(iii) Diffusion-based All-in-one Image Restoration.}
All-in-one image restoration aims to simultaneously address multiple diverse degradations within a single IR model. The key challenge lies in enabling diffusion-based IR models to adapt to various degradations while effectively leveraging shared restoration knowledge across different degradation types. The typical work is DA-CLIP\citep{luo2023controlling-DACLIP}, which introduces the degradation-aware CLIP to extract the degradation-aware embedding as the condition for diffusion models. In contrast, MPerceiver \citep{ai2023multimodal-mperceriver} extracts the holistic representation and multi-scale detail representation with the textual and visual branches of CLIP as the condition. Diff-Restore \citep{zhang2024diffrestore} also leverages the CLIP model to extract the degradation and semantic embedding but adopts a distinct conditioning strategy. For unknown hybrid degradation, AutoDIR \citep{jiang2023autodir} introduce the quality assessment model to identify which degradation should be removed and utilizes multiple steps for restoration. UIR-LoRA \citep{hu2022lora} leverages low-rank adaptation (LoRA) to encode task embeddings efficiently. Then it introduces a dynamic weighting strategy based on CLIP similarity between image features and predefined distortion text embeddings. In contrast, GenDeg \citep{rajagopalan2024gendeg} utilizes diffusion models to generate diverse distortions from clean images, expanding training data for all-in-one diffusion-based IR models, which further improves the performance of multiple IR tasks.

\textbf{(iv) Diffusion-based Image Weather Removal.}
Recently, there have been some diffusion-based methods for weather removal, which include image deraining, dehazing, and desowing. As the pioneering work, WeathDiff \citep{weather-diff} shares the same diffusion architecture with SR3, but it can support arbitrary image restoration by introducing the mean estimated noise of overlapped patches in the reverse process. In contrast, T$^3$-Diffusion \citep{chen2024teachingt3diffusion} defined a group of degradation-specific sub-prompts and utilizes the dynamical composition of them based on unseen degradation for weather removing, which outperforms DA-CLIP \citep{luo2023controlling-DACLIP} and Weather-Diff. ReviveDiff \citep{huang2024revivediff} enhances the perception levels for various weather degradation by introducing the coarse and fine branches into the diffusion model. 
 RainDiffusion \citep{shen2023rethinkingRainDiff} proposes an unpaired cycle consistency framework for image deraining, eliminating the need for paired training data, making it highly suitable for real-world applications. Depart from that, there are also two diffusion-based works for image dehazing. Dehaze-DDPM \citep{yu2023high-HazeDDPM} employs a two-stage dehazing pipeline, combining the Atmospheric Scattering Model (ASM) for physical modeling in the first stage and diffusion-based generative recovery of lost structural details in the second. Additionally, Wang et al. \citep{wang2023frequency-dehazy} enhance dehazing by emphasizing mid-to-high-frequency components in the diffusion process, leveraging a frequency-spectrum filter and skip connections to inject high-frequency details into the U-Net architecture.

\textbf{(v) Diffusion-based Image Deblurring.}
In Fig.2, there are three works that are devoted to image deblurring, including motion blur, and Gaussian blur. Among them, HI-Diff \citep{chen2023hierarchical-HIdiff} aims to utilize the diffusion model to generate the hierarchical compact priors for real image deblurring, and then utilize them to guide the restoration process of a transformer-based network. Another two works GibbsDDRM \citep{murata2023gibbsddrm} and BlindDPS \citep{blinddps} utilize diffusion model to estimate the blind blur kernel and deblurring. In particular, BlindDPS exploits the DPS \citep{dps} architecture and exploits
one parallel diffusion model for the degradation kernel
estimation. The diffusion model for kernel estimation
is pre-trained on synthetic kernels. Unlike BlindDPS,
GibbsDDRM \citep{murata2023gibbsddrm} achieves the sampling process with a partially collapsed Gibbs sampler \citep{van2008partially-sampler}, which samples both kernel
parameter and image together from the joint posterior
$p(xt|k, y)$(Please see in our manuscript).

\textbf{(vi) Diffusion-based Image Shadow Removal.}
We also clarify some works on diffusion-based shadow removal. ShadowDiffusion \citep{guo2023shadowdiffusion} leverages a pre-trained transformer backbone to extract the degradation prior (\ieno, the degradation-related features) from the distorted reference image. This extracted degradation prior is exploited as the auxiliary to refine the generated shadow mask and serves as the condition for shadow-free image generation. DeS3 \citep{jin2022des3} introduces a novel adaptive attention mechanism to distinguish the underlying objects and shadow regions, which is mask-free compared with existing works on shadow removal. Moreover, it utilizes a pre-trained ViT backbone as a loss to enhance the scene understanding capability. As described in DeS3, it can obtain a better performance compared with ShadowDiffusion.

\begin{table*}[htp]
\caption{Supervised diffusion model-based image restoration models}
 \resizebox{\textwidth}{!}{
 \begin{tabular}{l|c|m{8.5cm}|c}
  \Xhline{1.2pt}
        \textbf{Models(Papers)} & \textbf{Conference\&Year} & \textbf{Methods\&Insights} & \textbf{Target Tasks} \\ 
  \Xhline{1.2pt}
        SR3 \citep{SR3}  & TPAMI 2022 & Channel concatenation in DDPM with LR input  & SR  \\ 
        \rowcolor[gray]{0.9} SRdiff \citep{li2022srdiff}& Neurocomputing 2022 & DDPM with LR encoder information & SR \\ 
         CDPMSR \citep{niu2023cdpmsr} & Arxiv2023  & DDPM with Pretrained SR model & SR\\
        \rowcolor[gray]{0.9} Res-Diff \citep{shang2023resdiff} & AAAI2024   & DDPM with CNN model  &  SR  \\
        SDE-SR \citep{sde-sr} & SIBGRAPI 2022  & SDE with concat LR input  &  SR\\
        \rowcolor[gray]{0.9} LDM \citep{LDM} & CVPR 2022 & DDPM in latent space &  SR, Inpainting\\
        CDM \citep{CDM}  &  J.Mach.Learn.Res.2022 & Cascaded DDPM with condition augmentation  &  SR\\
        \rowcolor[gray]{0.9} ResShift \citep{yue2023resshift} & NeurIPS 2023 & Shift residual, improve the transition efficiency &  SR\\
        PASD \citep{yang2023pixel-aware}  &  Arxiv2023 & Personalized latent diffusion model, pixel aware attention  &  SR\\
        \rowcolor[gray]{0.9} PartDiff \citep{zhao2023partdiff} & Arxiv2023 & Partial Diffusion Model, Reduce cost &  SR\\
        SinSR \citep{wang2023sinsr}  &  CVPR 2024 & Single-step SR generation  &  SR\\
        \rowcolor[gray]{0.9} TextDiff \citep{liu2023textdiff} & Arxiv2023 & Masked guided residual diffusion model &  SR\\
        IDM \citep{IDM}  & CVPR 2023  & Neural representation for continuous SR & SR \\
        \rowcolor[gray]{0.9} DiffIR \citep{xia2023diffir}  & ICCV 2023  & DDPM with transformer-structure &  SR, Deblurring\\
        SAM-DiffSR \citep{wang2024sam-diffsr}  & Arxiv2024 & Structure-Modulated Diffusion Model  &  SR\\
        \rowcolor[gray]{0.9} BlindDiff \citep{li2024blinddiff} & Arxiv2024 & MAP-based optimization &  SR\\
        Diwa \citep{moser2023waving}  &  Arxiv2023 & Discrete Wavelet Transformation with DDPM  &  SR\\
        \rowcolor[gray]{0.9} PromptSR \citep{chen2023image-PromptSR} & Arxiv2023 & Text prompts for degradation priors &  SR\\
        CCSR \citep{sun2023improving-CCSR}  &  Arxiv2023 & Generative adversarial training for enhancing fine details  &  SR\\
        \rowcolor[gray]{0.9} DiffTSR \citep{zhang2023diffusion-difftsr} & CVPR 2024 & Text Diffusion Model guiding Image Diffusion Model  &  SR\\
        EDiffSR \citep{xiao2023ediffsr}  &  IEEE TGRS & DDPM with Efficient activation network &  RSSR\\  
        \rowcolor[gray]{0.9} YODA \citep{moser2023yoda} & Arxiv2023 &  DDPM working on spatial regions of attention maps &  SR\\
        HazeDDPM \citep{yu2023high-HazeDDPM} & Arxiv2023 & DDPM with Atmospheric Scattering Model  &  Dehazing\\      
       \rowcolor[gray]{0.9} WeatherDiff \citep{weather-diff}  & TPAMI 2023  & DDPM with patch-based sampling for size-agnostic tasks &  Deraining, Desnowing, Dehazing\\
        DeblurDPM \citep{deblur-DPM} & CVPR 2022 & Simple DDPM with predict and refine strategy   &  Deblurring\\
      \rowcolor[gray]{0.9}  Palette \citep{saharia2022palette}  & SIGGRAPH 2022  & DDPM with Concatenation of input, achieving multiple I2I tasks& Colorization,Inpainting, Uncropping, JPEG \\
        SR3+ \citep{SR3+}  & Arxiv2023 & Improved SR3 architecture, data augmentation,real-world SR& SR\\
         \rowcolor[gray]{0.9} Refusion \citep{luo2023refusion} & CVPRW 2023  & Latent DDPM with compression strategies, large-scale restoration&  Image Shadow Removal, Dehazing \\
         IR-SDE \citep{IR-SDE}  &ICML 2023 & Mean-reverting SDE  & Image Restoration\\
        \rowcolor[gray]{0.9} DACLIP \citep{luo2023controlling-DACLIP} & ICLR2024 &  CLIP priors with Prompt learning &  All in one IR\\
        Mperceiver \citep{ai2023multimodal-mperceriver} & Arxiv2023 & Stable Diffusion with two tpes of prompts  &  All in one IR\\
         \rowcolor[gray]{0.9} P2L \citep{chung2023prompt-P2L} & Arxiv 2023 & Prompt tuning for Stable Diffusion &  All in one IR\\ DDPG \citep{garber2023image-DDPG} & CVPR 2024 & DDPM with Preconditioned Guidance  &  All in one IR\\
         \rowcolor[gray]{0.9} ZeroAIR \citep{gou2023exploiting-zeroair} & Arxiv 2023 & Sampling with test-time degradation adapter &  Image Restoration\\
        FastDiffusionEM \citep{laroche2023fast-fast-diffusion-em} & Arxiv2023 &  DDPM with Expectation-Minimization estimation & Blind IR\\
        \rowcolor[gray]{0.9} SUPIR \citep{yu2024scaling-SUPIR} & CVPR 2024 & scaling up image restoration  &  Image Restoration\\AutoDIR \citep{jiang2023autodir} & Arxiv2023 & Stable diffusion with blind image quality assessment module   &  All in one IR\\
         \rowcolor[gray]{0.9} LFG-Diffusion \citep{mei2024latent-LFGdiffusion} & WACV 2024 & Conditioned on learned latent feature, fusing noise features &  Image Shadow Removal\\
       Deshadow-Anything \citep{zhang2023deshadow-anything} & Arxiv2023 & Finetuning on large-scale datasets  &  Image Shadow Removal\\
        \rowcolor[gray]{0.9} StableSR \citep{wang2023exploiting-stable-sr}  &  Arxiv2023 & Finetune SD with time-awareness and feature wrapping module& Real World Super-resolution \\
        DiffBIR \citep{qiu2023diffbfr}  & Arxiv2023 &Fine-tune latent diffusion model with ControlNet & Blind Face Restoration \\
        \rowcolor[gray]{0.9} RDDM \citep{liu2023residual-RDDM}  &  Arxiv2023 & Predict residuals, Redefine forward process & Real World SR \\
      PFStorer \citep{varanka2024pfstorer}  & Arxiv2024 &Personalized face restoration & Blind Face Restoration \\
        \rowcolor[gray]{0.9} Diff-Plugin \citep{liu2024diff-plugin}  &  CVPR 2024 & Task-Plugin Module & Image Restoration \\
        Yang \etal \citep{yang2023synthesizing}  &  Arxiv2022  & DDPM for data pairs synthesis for real-world tasks training & Training Data Synthesis  \\
         \rowcolor[gray]{0.9} Adadiff \citep{Adadiff}  & Arxiv2022 & Accelerated DDPM with adaptive diffusion priors& MRI Reconstruction \\
         HFS-SDE \citep{HFS-SDE}  & TMI 2022 & SDE in high-frequency space & MRI Reconstruction \\
        \rowcolor[gray]{0.9} Xie \etal   \citep{xie2023diffusion}  & Arxiv2023 & DDPM with three added noise distributions &Image Denoising \\
        PyDiff \citep{zhou2023pyramid-pydiff}  & Arxiv2023 & Pyramid diffusion with pyramid resolution style& Low-light Enhancement \\       
        \rowcolor[gray]{0.9} DiffLL \citep{jiang2023low-WCDM}  &  Arxiv2023 & Wavelet-based diffusion model with high efficiency& Low-light Enhancement \\
        ExploreDiffusion \citep{wang2023exposurediffusion}  & Arxiv2023 & Integrate DDPMM with a physics-based exposure model& Low-light Enhancement \\
        \rowcolor[gray]{0.9} DiffLL \citep{jiang2023low-WCDM}  &  Arxiv2023 & Wavelet-based diffusion model with high efficiency& Low-light Enhancement \\ Diff-Retinex \citep{yi2023diff-retinex}  & ICCV 2023 & Retinex decomposition with DDPM conditional generation& Low-light Enhancement \\
        \rowcolor[gray]{0.9} DiffLLE \citep{yang2023difflle-DiffLLE}  &  Arxiv2023 &  Diffusion-based domain calibration with unsupervised methods & Low-light Enhancement \\
        LLDiffusion \citep{wang2023lldiffusion-LLDiffuion}  & Arxiv2023 & Degradation-aware learning scheme with DDPM& Low-light Enhancement \\
         \rowcolor[gray]{0.9} DriftRec \citep{welker2022driftrec}& Arxiv2022  &SDE with adaptive strategy, Blind JPEG restoration & JPEG Artifacts Removal\\
        DiffGAR \citep{yin2022diffgar}  & Arxiv2022 & DDPM with encoder in Latent space & Generative artifacts removal\\
        \rowcolor[gray]{0.9} DG-DPM \citep{dg-dpm}  & Arxiv2022  & DDPM with domain generalization for real-world Deblurring & Deblurring \\
        RainDiffusion \citep{wei2023raindiffusion}  & Arxiv2023 &  Circular DDPM for real-world deraining & Real World Deraining \\
        \rowcolor[gray]{0.9} InDI \citep{INDI}  &  Arxiv2023 & Alternative methods proceed by iteratively restoring the input image of Diffusion model & Deblurring,SR, JPEG Restoration  \\
        ShadowDiffusion \citep{guo2023shadowdiffusion} &  CVPR 2023  & DDPM with unrolling-inspired diffusive sampling strategy & Image Shadowing Removal\\
        \rowcolor[gray]{0.9} DDS2M \citep{miao2023dds2m}  &  Arxiv2023 & DDPM with Variational Spatio-Spectral Module  & Hyperspectral Image Restoration \\
        SU$\mathbf{D}^{2}$ \citep{chan2023sud}  & Arxiv2023 & DDPM with few training pairs & Inpainting,Dehazing \\
        \rowcolor[gray]{0.9} DeS3 \citep{jin2022shadowdiffusion-driven}  &  Arxiv2022 & Classifier-driven attention guided DDPM   & Image Shadow Removal \\
        DiffBFR \citep{qiu2023diffbfr}  & Arxiv2023 & DDPM with few training pairs & Blind Face Restoration \\

  \Xhline{1.2pt}
  \end{tabular}}
    \label{table:Supervised-DM}
\end{table*}

\begin{table*}[htp]
\caption{Zero-shot diffusion model based image restoration models}
\resizebox{\textwidth}{!}{
\begin{tabular}{l|c|m{8cm}|c}
\Xhline{1pt}
\textbf{Models(Papers)} & \textbf{Conference\&Year} & \textbf{Methods\&Insights} & \textbf{Target Tasks} \\
\Xhline{1pt}
        ILVR \citep{choi2021ilvr}  & ICCV 2021   & Adds projection into sampling&SR\\
        \rowcolor[gray]{0.9} Repaint \citep{repaint}  & CVPR 2022 &Impose projection into sampling with resampling strategy. & Inpainting\\
        COPAINT \citep{copaint}  &  Arxiv2023& DDPM with Bayes framework, Enhance Coherence  & Inpainting  \\
        \rowcolor[gray]{0.9} GDP \citep{GDP}  & CVPR 2023 & DDPM with Generative diffusion priors  & SR, Deblurring, Inpainting, Colorization  \\
        CCDF \citep{CCDF}  & CVPR 2022 &Adds projection into sampling  & SR, Inpainting, MRI reconstruciton \\
        \rowcolor[gray]{0.9} SNIPS \citep{kawar2021snips}  & NeurIPS 2021  & DDPM in Spectral Space with SVD decomposition & SR, Deblurring, Compressive sensing \\
        DDRM \citep{DDRMDRM}  & ICLR 2022  &DDPM in Spectral Space with SVD decomposition & SR,Deblurring\\
        \rowcolor[gray]{0.9} JPEG-DDRM \citep{jpegddrm}  &  NeurIPS 2022   & JPEG restoration based on DDRM framework & JPEG Restoration \\
        DDNM \citep{ddnm}  & ICLR 2023  & Pretrained DDPM with Null Space, Time travel strategy &  SR, Deblurring, Inpainting, Colorization \\
        \rowcolor[gray]{0.9} MCG \citep{mcg}  & NeurIPS 2022   &  Manifold constraints  & Colorization,CT reconstruction \\
         DPS \citep{dps}  &  CVPR2023 & Posterior sampling &  SR, Inpainting, Deblurring\\
        \rowcolor[gray]{0.9} BlindDPS \citep{blinddps}  & ICLR 2023   & Parallel diffusions for images and kernels & Blind Deblurring \\
        GibbsDDRM \citep{murata2023gibbsddrm}  & Arxiv2023 & Joint posterior sampling & Deblurring, Vocal dereverberation  \\
        \rowcolor[gray]{0.9} Dif-Face \citep{yue2022difface}  & ICLR 2023  & Pretrained DDPM with initialization reverse image& Blind face Restoration\\
        DR2 \citep{wang2023dr2}  & CVPR 2023  & Degradation removal module with enhancement module  & Real-world tasks\\
        \rowcolor[gray]{0.9} DiracDiffusion \citep{fabian2023diracdiffusion}  & Arxiv2023   & Incremental reconstruction with ensured data consistency & Deblurring, Inpainting \\
        MCGdiff \citep{cardoso2023monte}  & Arxiv2023  &Propose efficient Monte Carlo sampler  &  SR,Inpainting,Deblurring\\
        \rowcolor[gray]{0.9} PSLD \citep{rout2023solving-PSLD}  & Arxiv2023   & Solve linear problems with pre-trained latent diffusion models & Inpainting, Denoising, Deblurring, Destriping, SR \\
         $\Pi$GDM \citep{iigdm}  & ICLR 2023   & Pseudoinverse-guided Diffusion Models, non-linear tasks  &  SR, Inpainting, JPEG\\
         \rowcolor[gray]{0.9} SaFaRI \citep{lee2024spatial-SAFARA}  &  Arxiv2024  & Spatial-and-frequency-aware   & SR, Deblurring, Inpainting\\      
         DeqIR \citep{cao2023deep-DeqIR}  & CVPR 2024& Parallel sampling from deep equilibrium point& SR, Deblurring \\
         \rowcolor[gray]{0.9} SIM-SGM \citep{SIM-SGM}  &  ICLR 2022  & Pretrained NCSN with matrix decomposition  & MRI Reconstruction \\      
         ADIR \citep{abu2022adir}  &  Arxiv2022& Adaptive diffusion model with CLIP  & SR, Deblurring \\
        \rowcolor[gray]{0.9} Feng \etal \citep{feng2023score} & ICCV 2023 & Posterior Sampling with Score-Based Priors  & Denoising, Deblurring\\
        Liu \etal \citep{liu2023improved} &  Arxiv2023 & A latent diffusion guider with a lightness-aware VQVAE model. & Colorization\\
        \rowcolor[gray]{0.9} DiffPIR \citep{zhu2023denoising-diffpir} & CVPR 2023 & Integrate plug-and-play method into diffusion& SR, Inpainting, Deblurring\\
        RED-Diff \citep{mardani2023variational-red-diff} &  Arxiv2023 & Regularization, Improve posterior score estimation & Inpainting, SR\\
        \rowcolor[gray]{0.9} DRUS \citep{zhang2023ultrasound-drus} & Arxiv2023 & DDRM-based & Ultrasound Image Reconstruction\\
  \Xhline{1pt}
  \end{tabular}}
    \label{table:Zero-shot-DM}
\end{table*}
\section{Experiments}
\subsection{Datasets}
Table~\ref{tab:datasets} summarizes the datasets used for different IR tasks, including SR, image inpainting, deblurring, denoising, shadow removal, image desnowing, image draining and image dehazing. It is composed of the released year, the number of training samples and testing samples, and short description. 
It is noteworthy that the contents and degradation modes are significantly different across the datasets from different IR tasks. Therefore, we summarize the commonly-used datasets based on IR tasks, including SR, image deblurring, image inpainting, shadow removal, desnowing, draining, and dehazing in Table~\ref{tab:datasets}. 
For traditional image SR (\ieno, bicubic downsampling), the standard training data is typically composed of DIV2K \citep{DIV2K} and Flick2K \citep{wang2019flickr1024}. However, the performance of the diffusion model is inherently bounded by the dataset size. Therefore, SR3 \citep{SR3} train 
diffusion model with ImageNet for natural image SR, and FFHQ \citep{FFHQ} for face SR. In the test process, it utilizes the ImageNet 1K \citep{imagenet1k} for the evaluation of natural image SR and CelebA-HQ for face SR. Depart from that, a series of works also introduce the commonly-used SR testing dataset for evaluation, such as Set5 \citep{set5}, Set14 \citep{Set14}, BSD100 \citep{BSD100}, Manga109 \citep{manga109}, Urban100 \citep{huang2015single-urban100}. For real-world SR, SR3+ \citep{SR3} provide two versions of training data, where the first version is composed of DF2K and OST \citep{OST} (\ieno, DIV2K, Flick2K, and OST300), and the second version contains an extra 61M in-house image and DF2K+OST. For evaluation, the testing data is composed of RealSR \citep{cai2019toward-realsr} and DRealSR \citep{wei2020component-drealsr}, obtained by two DSLR cameras with different lenses. 

For image deblurring, the diffusion model-based methods are usually trained with GoPro training dataset \citep{Gopro}, and validated on Gopro testing dataset, RealBlur-J \citep{Real-blur}, REDS \citep{Nah_2019_CVPR_Workshops_REDS}, and HIDE \citep{HIDE}. In the shadow removal task, ISTD \citep{ISTD} and SRD \citep{SRD} are utilized for the training and evaluation, where ISTD contains 135 scenes with the shadow mask. For the image dehazing task, three typical datasets are used for evaluation, including Haze-4K \citep{liu2021synthetic-haze4k}, Dense-Haze \citep{ancuti2019dense-densehaze}, and RESIDE \citep{li2019benchmarking-reside}. Among them, Haze-4K contains $4,000$ hazy images, Dense-Haze \citep{ancuti2019dense-densehaze} is composed of 33 pairs of out-door hazy and hazy-free images, and RESIDE \citep{li2019benchmarking-reside} collected real-world 443950 training  images and 5342 testing images. Image Desnowing utilizes three datasets, \ieno, CSD \citep{golodetz2018collaborative-csd}, Snow100k \citep{snow100k}, and SRRS \citep{chen2020jstasr-srrs}. The datasets in image deraining contain diverse rain types. For example, Rain100H \citep{yang2017deep-rain100}, Rain100L \citep{yang2017deep-rain100}, Rain800 \citep{Rain800}, DDN-Data \citep{fu2017DDN-data} contains amounts of synthesis rain streaks. RainDrop \citep{rain-drop} collected 1119 pairs of rainy/clean images with various backgrounds and raindrops. Outdoor-Rain \citep{outdoor-rain} considers the rain accumulation, which provides more reasonable modeling for heavy rainy images. SPA-data \citep{wang2019SPA-data} constructs large-scale real-world rain streaks by warping the clean image from multiple continuous rainy images. More details about the above datasets can be found in Table~\ref{tab:datasets}.
For diffusion model based image restoration models, we mainly divide used datasets into two categories according to training ways: datasets for supervised learning and datasets for zero-shot restoration. \\

\noindent\textbf{Training datasets:}\\
DIV2K \citep{DIV2K},Flickr2K \citep{wang2019flickr1024}, LSUN  \citep{yu2015lsun}, ImageNet \citep{russakovsky2015imagenet}, FFHQ \citep{FFHQ}, AFHQ \citep{AFHQ}, fastMRI knee \citep{zbontar2018fastmri}, Places365 \citep{places365}, Celeba-HQ \citep{Celeba-hq}, GoPro \citep{Gopro}, OST \citep{OST}, Places2 \citep{zhou2017places2}, ISTD \citep{ISTD}, SRD \citep{SRD}, Snow100k \citep{snow100k}, Rain200H \citep{yang2017deep-rain100}, Rain800 \citep{Rain800}, Outdoor-rain \citep{outdoor-rain}, Rain-drop \citep{rain-drop}, DIV8K \citep{gu2019div8k}
     \\
\noindent\textbf{Test datasets:} \\
OST300 \citep{OST300}, LSUN \citep{yu2015lsun}, Set5 \citep{set5}, Set14 \citep{Set14}, BSD100 \citep{BSD100}, Celeba-HQ \citep{Celeba-hq}, Manga109 \citep{manga109}, FFHQ \citep{FFHQ}, DIV2K \citep{DIV2K}, ImageNet-1k \citep{imagenet1k}, Gopro \citep{Gopro}, places10k \citep{places365}, Flickr1024 \citep{wang2019flickr1024}, USC-SIPI \citep{USC-SIPI}, SRD \citep{SRD}, ISTD \citep{ISTD}, AFHQ \citep{AFHQ}, Real-Blur \citep{Real-blur}, HIDE \citep{HIDE}, LFWTest \citep{LFW}, WIDER-Test \citep{WIDER-test}, Snow100k(test) \citep{snow100k}, Outdoor-rain(test1) \citep{outdoor-rain}, RainDrop-A \citep{rain-drop}, NTIRE \citep{perez2021ntire}, LOL \citep{lol}, LoLi-Phone \citep{lolo-phone}\\
These mentioned datasets about diffusion models covers different image restoration tasks like super-resolution, deblurring, deraining, denoising, inpainting, JPEG artifacts removal, desnowing, and real-world datasets. The detailed information can be shown in Table~\ref{tab:sm-im} for the supervised model and Table~\ref{tab:zm-im} for the zero-shot model respectively.

 \begin{table*}[htp]
  \centering
  \vspace{3mm}
  \caption{Summary of widely used datasets in different diffusion model-based image restoration tasks}
    \resizebox{\textwidth}{!}{\begin{tabular}{c|l|c|c|c|c}
    \Xhline{1pt}
    \multicolumn{1}{c|}{Task}  & \multicolumn{1}{c|}{Dataset} & Year  & Training & Testing & Short Description \\ \Xhline{1pt}
    \multirow{11}[0]{*}{Image Super-resolution} & DIV2K \citep{DIV2K} & 2017  & 800   & 100  & 2K resolutions \\
        & \cellcolor[gray]{0.9}Flickr2K \citep{wang2019flickr1024} & \cellcolor[gray]{0.9}2017  & \cellcolor[gray]{0.9}2650  & \cellcolor[gray]{0.9}-     & \cellcolor[gray]{0.9}2K resolutions \\
          & Set5 \citep{set5} & 2012  & -     & 5    & Classic 5 images \\
          & \cellcolor[gray]{0.9}Set14 \citep{Set14} & \cellcolor[gray]{0.9}2012  & \cellcolor[gray]{0.9}-     & \cellcolor[gray]{0.9}14    & \cellcolor[gray]{0.9}Classic 14 images \\
          & BSD100 \citep{BSD100} & 2001  & -     & 100   & Objects, Natural images \\
        & \cellcolor[gray]{0.9}Manga109 \citep{manga109} & \cellcolor[gray]{0.9}2015  & \cellcolor[gray]{0.9}-     & \cellcolor[gray]{0.9}109  & \cellcolor[gray]{0.9}109 manga volumes \\
          & Urban100 \citep{huang2015single-urban100} & 2015  & -     & 100   & \textcolor[rgb]{ .129,  .145,  .161}{100 urban scenes} \\
          &\cellcolor[gray]{0.9}OST300 \citep{OST300} & \cellcolor[gray]{0.9}2018  & \cellcolor[gray]{0.9}-     & \cellcolor[gray]{0.9}300  & \cellcolor[gray]{0.9}Outdoor scenes \\
          & DIV8K \citep{gu2019div8k} & 2019  & 1304  & 100    & 8k resolution for large scale factor \\
         &\cellcolor[gray]{0.9}RealSR \citep{cai2019toward-realsr} &\cellcolor[gray]{0.9}2019  &\cellcolor[gray]{0.9}565  &\cellcolor[gray]{0.9}30    &\cellcolor[gray]{0.9}Real world images for SR \\
          & DRealSR \citep{wei2020component-drealsr} & 2020  & 884  & 83    & Large-scale dataset \\
    \hline
    \multirow{3}[0]{*}{Image Deblurring} & GoPro \citep{Gopro} & 2017  & 2103  & 1111   & Blurred images at 1280x720 \\
         &\cellcolor[gray]{0.9}HIDE \citep{HIDE}  & \cellcolor[gray]{0.9}2019  & \cellcolor[gray]{0.9}6397  & \cellcolor[gray]{0.9}2025   & \cellcolor[gray]{0.9}Blurry and sharp image pairs \\
          & RealBlur \citep{Real-blur} & 2020  & 3758  & 980    & 182 different scenes \\
    \hline
    \multirow{3}[0]{*}{Image Denoising}
    &\cellcolor[gray]{0.9}Kodak \citep{franzen1999kodak}&\cellcolor[gray]{0.9}1999  &\cellcolor[gray]{0.9}-  &\cellcolor[gray]{0.9} 24  &\cellcolor[gray]{0.9}Resolution=768x512 \\
          & CBSD68 \citep{martin2001database-cbsd68} & 2001  & - & 68   & Images with different noisy levels\\
           &\cellcolor[gray]{0.9}McMaster \citep{zhang2011color-mcmaster} &\cellcolor[gray]{0.9}2011  &\cellcolor[gray]{0.9}- &\cellcolor[gray]{0.9}18   &\cellcolor[gray]{0.9}Crop size=500x500\\
    \hline
    \multirow{4}[0]{*}{Image Classification} & ImageNet \citep{russakovsky2015imagenet} & 2010  & 1,281,167 & 100,000  & 1000 object classes \\
         & \cellcolor[gray]{0.9}ImageNet1K \citep{imagenet1k} &\cellcolor[gray]{0.9}2020  &\cellcolor[gray]{0.9}-     &\cellcolor[gray]{0.9}1000   &\cellcolor[gray]{0.9}1000 image subset of ImageNet \\
          & LSUN \citep{yu2015lsun}  & 2015  & 120,000 to 3,000,000(each category) & 1000(each category)  & 10 scene categories(Classification) \\
          &\cellcolor[gray]{0.9}Places365 \citep{places365} &\cellcolor[gray]{0.9}2019  & \cellcolor[gray]{0.9}1.8 million &\cellcolor[gray]{0.9}36000  &\cellcolor[gray]{0.9} 434 scene classes \\
    \hline
    \multirow{5}[0]{*}{Face Generation(Recognition)} & LFW \citep{LFW} & 2008  & 13233 & -    & Images from web with 1080 people \\
         &\cellcolor[gray]{0.9}FFHQ \citep{FFHQ}  & \cellcolor[gray]{0.9}2019  &\cellcolor[gray]{0.9}70000 &\cellcolor[gray]{0.9}-     &\cellcolor[gray]{0.9}Various faces at 1024x1024 \\
          & Celeba-HQ \citep{Celeba-hq} & 2018  & 30000 & -  & High quality faces at 1024x1024 \\
          &\cellcolor[gray]{0.9}AFHQ \citep{AFHQ}  &\cellcolor[gray]{0.9}2020  &\cellcolor[gray]{0.9}15000 &\cellcolor[gray]{0.9}-    & \cellcolor[gray]{0.9}Animal faces at 512x512 \\
          & CelebA \citep{liu2015deep-celeba} & 2015  & 202599 & -   & Resolution at 178x218 \\
    \hline
    \multirow{2}[0]{*}{Shadow Removal} 
    &\cellcolor[gray]{0.9}ISTD \citep{ISTD}  &\cellcolor[gray]{0.9}2018  &\cellcolor[gray]{0.9}1330  &\cellcolor[gray]{0.9}540    &\cellcolor[gray]{0.9}135 scenes with shadow mask images \\
          & SRD \citep{SRD}   & 2017  & 2680  & 408    & Large scale dataset for shadow removal \\
    \hline
    \multirow{3}[0]{*}{Image Desnowing} 
     &\cellcolor[gray]{0.9}CSD \citep{golodetz2018collaborative-csd}   &\cellcolor[gray]{0.9}2021  &\cellcolor[gray]{0.9}8000  &\cellcolor[gray]{0.9}2000   &\cellcolor[gray]{0.9}Large scale dataset for desnowing \\
          & Snow100k \citep{snow100k} & 2017  & 50000 & 50000  & Own 1369 realistic snowy images \\
          &\cellcolor[gray]{0.9}SRRS \citep{chen2020jstasr-srrs}  &\cellcolor[gray]{0.9}2020  &\cellcolor[gray]{0.9}15000(paired)+1000(unpaired)&\cellcolor[gray]{0.9}- &\cellcolor[gray]{0.9}Real scenarios from Internet \\
    \hline
    \multirow{6}[0]{*}{Image Deraining} & RainDrop \citep{rain-drop} & 2018  & 1119  & -     & Various scenes and raindrops \\
          &\cellcolor[gray]{0.9}Outdoor-Rain \citep{outdoor-rain} &\cellcolor[gray]{0.9}2019  &\cellcolor[gray]{0.9}9000  &\cellcolor[gray]{0.9}1500   &\cellcolor[gray]{0.9}Outdoor rainy images \\
          & DDN-data \citep{Fu_2017_CVPR-ddndata} & 2017  & 9100  & 4900  & Real-world clean/rainy image pairs \\
          &\cellcolor[gray]{0.9}SPA-data \citep{wang2019SPA-data} &\cellcolor[gray]{0.9}2019  &\cellcolor[gray]{0.9}295000 &\cellcolor[gray]{0.9}1000   &\cellcolor[gray]{0.9}Various natural rain scenes \\
          & Rain-100H \citep{yang2017deep-rain100} & 2017  & 1800    & 100    & Five rain streaks \\
          &\cellcolor[gray]{0.9}Rain-100L \citep{yang2017deep-rain100} &\cellcolor[gray]{0.9}2017  &\cellcolor[gray]{0.9}200    &\cellcolor[gray]{0.9}100    &\cellcolor[gray]{0.9}Five rain streaks \\
    \hline
    \multirow{3}[0]{*}{Image Dehazing} & Haze-4K \citep{liu2021synthetic-haze4k} & 2021  &4000&-  & Attached with associated images \\
          &  \cellcolor[gray]{0.9}Dense-Haze \citep{ancuti2019dense-densehaze} & \cellcolor[gray]{0.9}2019  &\cellcolor[gray]{0.9}33&\cellcolor[gray]{0.9}-  &\cellcolor[gray]{0.9}Out-door hazy scenes \\
          & RESIDE \citep{li2019benchmarking-reside} & 2019  & 443950 & 5342   & Real-world hazy images \\ \Xhline{1pt}
    \end{tabular}}
  \label{tab:datasets}
\end{table*}

\subsection{Evaluation Metrics}
The objective and subjective metrics play a vital role in measuring and comparing the performances between different algorithms of DM-based IR. In this section, we clarify the commonly-used metrics in image restoration in detail, \ieno, PSNR, SSIM \citep{SSIM}, LPIPS \citep{LPIPS}, DISTS \citep{DISTS}, FID \citep{FID}, KID \citep{KID}, NIQE \citep{mittal2012makingNIQE} and PI \citep{blau20182018PIRMchallenge}. 
\begin{itemize}
    \item \textbf{PSNR}, as the most popular metric in image restoration, aims to measure the pixel-wise distance between a distorted image and its corresponding clean image by computing their mean square error(MSE). 
    \item  \textbf{SSIM} \citep{SSIM} is also the traditional image quality assessment (IQA) metric, which intends to satisfy the human visual perception system. Compared with PSNR, it compares the similarity between distorted and clean images from three perspectives: contrast, brightness, and structure. To further improve it, multi-scale information is introduced to SSIM, termed MS-SSIM \citep{wang2003multiscale-MS-SSIM}. In contrast to learning-based IQA metrics, SSIM possesses fast computational speed, while still far from human perception. 
    \item \textbf{LPIPS} \citep{LPIPS} is a full-reference learning-based IQA metric widely applied in perception-oriented image restoration tasks. Instead of utilizing image-wise statistics for quality measurement, it exploits the pre-trained AlexNet as the feature extractor and optimizes the linear layer toward human perception. The lower value LPIPS means the two images are more similar in the perception space. 
    \item \textbf{DISTS} \citep{DISTS} observes that the texture similarity and structure similarity of two images can be measured by the means and correlation of their features from VGG \citep{simonyan2014veryVGG}, respectively. Based on this finding, this work performs an SSIM-like distance measurement for texture and structure similarity in the feature space. 
    \item \textbf{FID} \citep{FID} (Fréchet inception distance) is widely used to measure the fidelity and diversity of generated images, which is the improvement of Inception Score \citep{barratt2018note-IS}. In contrast to IS, which lacks real-world reference images, FID leverages the features from the coding layer of the inception model to model the multivariate Gaussian distribution for sampling images and compute the Fréchet distance between the distribution of generated images and reference images. 
    \item \textbf{KID} \citep{KID} and FID exploit the same features from the inception model for quality assessment while owning the different distance measurement strategy (\ieno, the maximum mean discrepancy (MMD) with a polynomial kernel). In particular, KID is more stable than FID even with few samples. 
    \item \textbf{NIQE} \citep{mittal2012makingNIQE} is an early no-reference/blind image quality assessment metric, where the quality score is computed with the distance between the natural scene static (NSS) of distorted images and natural images with Multivariate Gaussian Model (MGM). 
    \item \textbf{PI} is proposed in PIRM Challenge
     on perceptual SR \citep{blau20182018PIRMchallenge}, aiming to evaluate the perceptual quality of super-resolved images. It is defined as $PI=0.5((10-\mathrm{Ma})+\mathrm{NIQE})$, where $\mathrm{Ma}$ \citep{ma2017learningMa} is a no-reference IQA metric for SR. 
\end{itemize}
Some works also use (Deep Image Structure and Texture Similarity) to evaluate the image quality more correlated to human perception. DISTS is based on a pre-trained VGG network, extracting features for both reference images and test images. The distance measure of DISTS is a modification of SSIM with luminance $l$ and structure $s$ components.
This section will introduce the most widely used evaluation metrics for diffusion model based models in image restoration. The common image quality assessment(IQA) includes objective methods and subjective methods.
Subjective methods require subjects to evaluate and score for different images which are time-consuming and requiring many human resources. Thu, objective methods are widely applied in evaluation for image restoration tasks. \textbf{PSNR} (Peak Signal to Noise Ratio) is the most widely used quality assessment metric in the field of image and video processing. It mainly estimates the quality of the restored image by calculating the mean square error(MSE) of the pixel values of the distorted image and the original high-resolution image. \textbf{SSIM}(Structure Similarity Index Measure) \citep{SSIM}is another traditional quality assessment metric. It meets better in human visual perception than PSNR. It mainly compares the similarity between two images from three perspectives: contrast, brightness, and structure. PSNR and SSIM are two basic metrics to evaluate the data consistency in pixel levels between generated images and original high-quality images. However, they could not show perceptual quality of images. 
Diffusion model have exceptional image generation ability on realness and high-fidelity. Thus, applying perceptual quality metrics to generated images are crucial in image generation tasks. \textbf{LPIPS}(Learned Perceptual Image Patch Similarity) \citep{LPIPS} is widely applied by many researchers to evaluate the perceptual quality of generated images.  The core idea of LPIPS is to compute the similarity between different patches of two images leveraging a pretrained CNN-based network like Resnet-50. This quality is more in line with human vision system. The lower value of LPIPS means two images are more similar to each other. 

\subsection{Implementation details}
We summarize the implementation details of supervised and zero-shot DM-based IR methods in Table~\ref{tab:sm-im}, and Table~\ref{tab:zm-im}, respectively. For supervised DM-based IR, we clarify the training dataset, testing dataset, and some crucial implementation details, including batch size, iterations, learning rate, and the number of sampling steps in the training and inference stages. For zero-shot DM-based IR, we summarize the implementation details from the perspectives of testing datasets, pre-trained models, and sampling steps.

\begin{table*}[h]
\caption{Implementation details of supervised diffusion model-based methods. BS, Iters, LR, Step-T, and Steps-I denote the batch size, training iterations, learning rate, and the number of sampling steps in the training and inference stages, respectively.}
\centering
\renewcommand{\arraystretch}{1.5}
\resizebox{\textwidth}{!}{
\begin{tabular}{c|lllccccc}
\hline
\multirow{2}{*}{Target Tasks}   & \multirow{2}{*}{Methods} & \multirow{2}{*}{Training Datasets}                                                          & \multirow{2}{*}{Testing Datasets}                                                   & \multicolumn{5}{c}{Implementaion Details}                                                                                               \\ \cline{5-9} 
                                  &                          &                                                                                             &                                                                                     & BS & Iters & LR                                                        & Steps-T& Steps-I \\ \hline
\multirow{10}{*}{Super-resolution} & SR3 \citep{SR3}                      & FFHQ \citep{FFHQ}, ImageNet \citep{russakovsky2015imagenet}                                                                              & CelebA-HQ \citep{Celeba-hq},ImageNet 1k \citep{imagenet1k}                                                               & 32        & 1M         & 1e-4                                                                 & 2000             & 2000                \\ 
                                  &    
 \cellcolor[gray]{0.9}SRDiff \citep{li2022srdiff}                   & \cellcolor[gray]{0.9}CelebA \citep{liu2015deep-celeba}, DIV2K \citep{DIV2K}+Flickr2K \citep{wang2019flickr1024}                                                                      & \cellcolor[gray]{0.9}Dev split, DIV2K(100 test) \citep{DIV2K}                                                          & \cellcolor[gray]{0.9}16        & \cellcolor[gray]{0.9}400K       & \cellcolor[gray]{0.9}2e-4                                                                 & \cellcolor[gray]{0.9}100              & \cellcolor[gray]{0.9}100                 \\
                                  & CDPMSR \citep{niu2023cdpmsr}                   & DIV2K \citep{DIV2K}                                                                                       & Set5 \citep{set5}, Set14 \citep{Set14}, Urban100 \citep{huang2015single-urban100}, BSD100 \citep{BSD100}, Manga109 \citep{manga109} & 16        & 300K       & 1e-4                                                                 & 1000             & 100                 \\
                                  & \cellcolor[gray]{0.9}SR3+ \citep{SR3+}                     & \cellcolor[gray]{0.9}DIV2K \citep{DIV2K},Flick2K \citep{wang2019flickr1024}, OST300 \citep{OST300}                                                  & \cellcolor[gray]{0.9}RealSR \citep{cai2019toward-realsr}, DRealSR \citep{wei2020component-drealsr}                                                                     & \cellcolor[gray]{0.9}256       & \cellcolor[gray]{0.9}1.5M       & \cellcolor[gray]{0.9}-                                                               & \cellcolor[gray]{0.9}2000             & \cellcolor[gray]{0.9}256                 \\
                                  & Resdiff \citep{shang2023resdiff}                  & FFHQ, DIV2K                                                                                 & CelebA(5000) \citep{liu2015deep-celeba}, DIV2K(100) \citep{DIV2K},Urban100(20) \citep{huang2015single-urban100} & 4         & 250k       & 1e-4                                                                 & 1000             & -                   \\
                                  & \cellcolor[gray]{0.9}SDE-SR \citep{sde-sr}                   & \cellcolor[gray]{0.9}FFHQ \citep{FFHQ}                                                                                        & \cellcolor[gray]{0.9}CelebA \citep{liu2015deep-celeba}                                                                              & \cellcolor[gray]{0.9}-    & \cellcolor[gray]{0.9}1M         & \cellcolor[gray]{0.9}2e-4                                                                 & \cellcolor[gray]{0.9}2000             & \cellcolor[gray]{0.9}2000                \\
                                  & CDM \citep{CDM}                      & LSUN \citep{yu2015lsun}                      & LSUN \citep{yu2015lsun}                                                                                & 1024      & 40k        & 1e-4                                                                 & 2000             & 100                 \\
                                  
                                   &  \cellcolor[gray]{0.9}StableSR \citep{wang2023exploiting-stable-sr}                   & \cellcolor[gray]{0.9}DIV2K \citep{DIV2K}, Flickr2K \citep{wang2019flickr1024}, DIV8K \citep{gu2019div8k}, FFHQ \citep{FFHQ}                                                                                       & \cellcolor[gray]{0.9}RealSR \citep{cai2019toward-realsr}, DRealSR \citep{wei2020component-drealsr}                                                                              & \cellcolor[gray]{0.9}192    & \cellcolor[gray]{0.9}--         & \cellcolor[gray]{0.9}5e-5                                                                 & \cellcolor[gray]{0.9}2000             & \cellcolor[gray]{0.9}200                \\

                                  & PASD \citep{yang2023pixel-PASD}                   & DIV2K \citep{DIV2K}, Flickr2K \citep{wang2019flickr1024}, OST \citep{OST}, FFHQ \citep{FFHQ}                                                                                        & RealSR \citep{cai2019toward-realsr}, DRealSR \citep{wei2020component-drealsr}                                                                              & 4    & 500k         & 5e-5                                                                 & --             & 20                \\
                                  
                                  & \cellcolor[gray]{0.9}IDM \citep{IDM}                      & \cellcolor[gray]{0.9}FFHQ \citep{FFHQ}, DIV2K \citep{DIV2K}
                                  & \cellcolor[gray]{0.9}DIV2K(100) \citep{DIV2K}, LSUN \citep{yu2015lsun}                                                                    & \cellcolor[gray]{0.9}64    & \cellcolor[gray]{0.9}1.5M     & \cellcolor[gray]{0.9}1e-4  & \cellcolor[gray]{0.9}2000           & \cellcolor[gray]{0.9}2000             \\ \hline
\multirow{4}{*}{IR}               & LDM \citep{LDM}                      & ImageNet \citep{russakovsky2015imagenet}                                                                                    & ImageNet 1K \citep{imagenet1k}                                                                         & 32        & 1M         & 1e-4                                                                 & 2000             & 2000                \\
 & \cellcolor[gray]{0.9}SUPIR \citep{yu2024scaling-SUPIR}                      &                                                                   \cellcolor[gray]{0.9}DIV2K \citep{DIV2K}, Flickr2K \citep{wang2019flickr1024}, OST \citep{OST}, FFHQ \citep{FFHQ}                  & \cellcolor[gray]{0.9}RealSR \citep{cai2019toward-realsr}, DRealSR \citep{wei2020component-drealsr}, RealPhoto60 \citep{yu2024scaling-SUPIR}                                                                         & \cellcolor[gray]{0.9}256        & \cellcolor[gray]{0.9}--         & \cellcolor[gray]{0.9}1e-5                                                                 & \cellcolor[gray]{0.9}100             & \cellcolor[gray]{0.9}50               \\
                                  & Palette \citep{saharia2022palette}                  & ImageNet \citep{russakovsky2015imagenet},Places2 \citep{zhou2017places2}                                                                            & imagenet- ctest10k \citep{russakovsky2015imagenet} places10k \citep{zhou2017places2}                                                            & 512       & 500K       & 1e-4                                                                 & 2000             & 1000                \\ 
                                  & \cellcolor[gray]{0.9}DiffIR \citep{xia2023diffir}                   & \cellcolor[gray]{0.9}Places-Standard \citep{places365}, DF2K \citep{DIV2K,wang2019flickr1024},CelebA-HQ \citep{Celeba-hq} &\cellcolor[gray]{0.9}CelebA \citep{liu2015deep-celeba}, Set5 \citep{set5}, Set14 \citep{Set14}, Urban100 \citep{huang2015single-urban100}                                                       & \cellcolor[gray]{0.9}30        & \cellcolor[gray]{0.9}-          & \cellcolor[gray]{0.9}2e-4                                                                 & \cellcolor[gray]{0.9}4                & \cellcolor[gray]{0.9}4                   \\ \hline
\multirow{3}{*}{Shadow Removal}   & Shadowdiffusion \citep{guo2023shadowdiffusion}          & ISTD \citep{ISTD}, ISTD+ \citep{le2019shadow-istd+}, SRD \citep{SRD}                                                                            & ISTD-test \citep{ISTD}, ISTD+test \citep{le2019shadow-istd+}, SRD-test \citep{SRD}                                                    & -    & -     & 3e-5                                                                 & 1000             & 25                  \\
                                  & 
  \cellcolor[gray]{0.9}DeS3 \citep{jin2022shadowdiffusion-driven}                     & \cellcolor[gray]{0.9}SRD \citep{SRD}                                                                                         & \cellcolor[gray]{0.9}SRD \citep{SRD}                                                                                 & \cellcolor[gray]{0.9}-    & \cellcolor[gray]{0.9}-& \cellcolor[gray]{0.9}-                                                               & \cellcolor[gray]{0.9}1000             & \cellcolor[gray]{0.9}25                  \\
                                  & Refusion \citep{luo2023refusion}                 & Flickr1024 \citep{wang2019flickr1024} (Random 800   images)                                                            & Flickr1024 \citep{wang2019flickr1024} (Select 112 images)                                                      & 8         & 500K       & 3e-4                                                                 & 100              & 100                 \\ \hline
\multirow{2}{*}{Image Deblurring} &  \cellcolor[gray]{0.9}Deblur-DPM \citep{deblur-DPM}               & \cellcolor[gray]{0.9}GoPro \citep{Gopro}                                                                                       & \cellcolor[gray]{0.9}HIDE \citep{HIDE}                                                                                & \cellcolor[gray]{0.9}32        & \cellcolor[gray]{0.9}-     & \cellcolor[gray]{0.9}1e-4                                                                 & \cellcolor[gray]{0.9}2000             & \cellcolor[gray]{0.9}10-500              \\
                                  & DG-DPM \citep{dg-dpm}                   & GoPro \citep{Gopro}                                                                                       & RealBlur-J \citep{Real-blur}, REDS \citep{Nah_2019_CVPR_Workshops_REDS}, HIDE \citep{HIDE}                                                              & 256       & 60k        & 1e-4                                                                 & 1000             & 20-1000             \\ \hline
\multirow{2}{*}{Image Deraining}  \rule{0pt}{12pt}&  \cellcolor[gray]{0.9}WeatherDiff \citep{weather-diff}            &\cellcolor[gray]{0.9}Snow100k \citep{snow100k}, Rain-drop \citep{rain-drop}, Outdoor-rain \citep{outdoor-rain}                                                        &\cellcolor[gray]{0.9}Outdoor-rain(test) \citep{outdoor-rain}, Snow100k(test) \citep{snow100k}, RainDrop-A \citep{rain-drop}                                   &\cellcolor[gray]{0.9}64        &\cellcolor[gray]{0.9}2000k      & \cellcolor[gray]{0.9}2e-5                                                                 & \cellcolor[gray]{0.9}-          & \cellcolor[gray]{0.9}10,50               \\
                                  & RainDiffusion \citep{wei2023raindiffusion}            & Rain200H \citep{yang2017deep-rain100}, Rain200L \citep{yang2017deep-rain100}, Rain800 \citep{Rain800}, DDN-Data \citep{Fu_2017_CVPR-ddndata}                                                       & Rain200H \citep{yang2017deep-rain100}, Rain200L \citep{yang2017deep-rain100}, Rain800 \citep{Rain800}, DDN-Data \citep{fu2017DDN-data}                                              & 4         & -     & 2e-5                                                                 & -           & -              \\ \hline
\end{tabular}}
\label{tab:sm-im}
\end{table*}

\begin{table*}[htp]
\caption{Implementation details used in zero-shot diffusion model-based methods}
\resizebox{\textwidth}{!}{
\begin{tabular}{llll}
\Xhline{1pt}
Methods & Training Datasets                                      & Test Datasets &Inference Steps \\ \Xhline{1pt}
RED-Diff \citep{mardani2023variational-red-diff}                                     & Pretrained on ImageNet \citep{russakovsky2015imagenet}                                       & ImageNet 1K \citep{imagenet1k}                       & 1000                \\
\rowcolor[gray]{0.9} Rapaint \citep{repaint}                                      & Pretrained on CelebA-HQ \citep{Celeba-hq} and ImageNet \citep{russakovsky2015imagenet}                              & CelebA-HQ \citep{Celeba-hq}, ImageNet \citep{russakovsky2015imagenet}                     & 250                 \\
CoPaint \citep{copaint}
& Pretrained on ImageNet \citep{russakovsky2015imagenet}, CelebA-HQ \citep{Celeba-hq}                            & CelebA-HQ \citep{Celeba-hq},ImageNet \citep{russakovsky2015imagenet}                  & 250                 \\
\rowcolor[gray]{0.9} ILVR \citep{choi2021ilvr}                                         & Pretrained on FFHQ \citep{FFHQ},AFHQ \citep{AFHQ}, LSUN\citep{yu2015lsun},  Places365 \citep{places365} & Images from the website           & 250             \\
CCDF \citep{CCDF}                                         & Pretrained on FFHQ \citep{FFHQ}, AFHQ \citep{AFHQ} , fastMRI knee \citep{zbontar2018fastmri}                     & FFHQ \citep{FFHQ}, AFHQ \citep{AFHQ}, fastMRI knee \citep{zbontar2018fastmri}          & 20                  \\
\rowcolor[gray]{0.9} SNIPS \citep{kawar2021snips}                                        & Pretrained NSCN on CelebA \citep{liu2015deep-celeba},                                               & CelebA \citep{liu2015deep-celeba}, LSUN \citep{yu2015lsun}                         & 500                 \\
DDRM \citep{DDRMDRM}                                         & Pretrained on CelebA-HQ \citep{Celeba-hq}, LSUN \citep{yu2015lsun}, ImageNet \citep{russakovsky2015imagenet}                        & FFHQ \citep{FFHQ}, ImageNet 1K \citep{imagenet1k},LSUN \citep{yu2015lsun}                   & 20                  \\
\rowcolor[gray]{0.9} DDNM \citep{ddnm}                                         & Pretrained on ImageNet \citep{russakovsky2015imagenet}, CelebA-HQ \citep{Celeba-hq}                           & ImageNet 1K \citep{imagenet1k} CelebA \citep{liu2015deep-celeba}                   & 100                 \\
Bahjat Kawar et.al \citep{jpegddrm}                           & Pretrained on ImageNet \citep{russakovsky2015imagenet}                                       & ImageNet 1K \citep{imagenet1k}                           & 20                  \\
\rowcolor[gray]{0.9} MCG \citep{mcg}                                          & Pretrained on FFHQ \citep{FFHQ},ImageNet \citep{russakovsky2015imagenet}                                 & FFHQ \citep{FFHQ},LSUN \citep{yu2015lsun}                              & 1000                \\
DPS \citep{dps}                                          & Pretrained on ImageNet \citep{russakovsky2015imagenet}                                                                 & FFHQ \citep{FFHQ}, ImageNet 1K \citep{imagenet1k}                          & 1000                \\
\rowcolor[gray]{0.9} GibbsDDRM \citep{murata2023gibbsddrm}                                    & Pretrained on FFHQ \citep{FFHQ}, AFHQ \citep{AFHQ}                                    & FFHQ \citep{FFHQ}, AFHQ \citep{AFHQ}                      & 50                  \\
DifFace \citep{yue2022difface}                                      & Pretrained on FFHQ \citep{FFHQ}, ImageNet \citep{russakovsky2015imagenet}                                & Celeba-HQ \citep{Celeba-hq}                        & 250                 \\
\rowcolor[gray]{0.9} Dirac-Diffusion \citep{fabian2023diracdiffusion}                              & SDE-VP trained on CelebA \citep{liu2015deep-celeba} , ImageNet \citep{russakovsky2015imagenet}                                       & CelebA \citep{liu2015deep-celeba}, ImageNet \citep{imagenet1k}                  & 100              \\
IIGDM \citep{iigdm}                                        & Diffusion model trained on ImageNet \citep{russakovsky2015imagenet}                                      & ImageNet \citep{imagenet1k}                 & 100                 \\
\rowcolor[gray]{0.9} ADIR \citep{abu2022adir}                                         & Pretrained on ImageNet \citep{russakovsky2015imagenet}                                      & DIV2K \citep{DIV2K}                             & 1,000           \\ \Xhline{1pt}
\end{tabular}}
\label{tab:zm-im}
\end{table*}

\begin{figure*}[h]
	\includegraphics[width=1\linewidth]{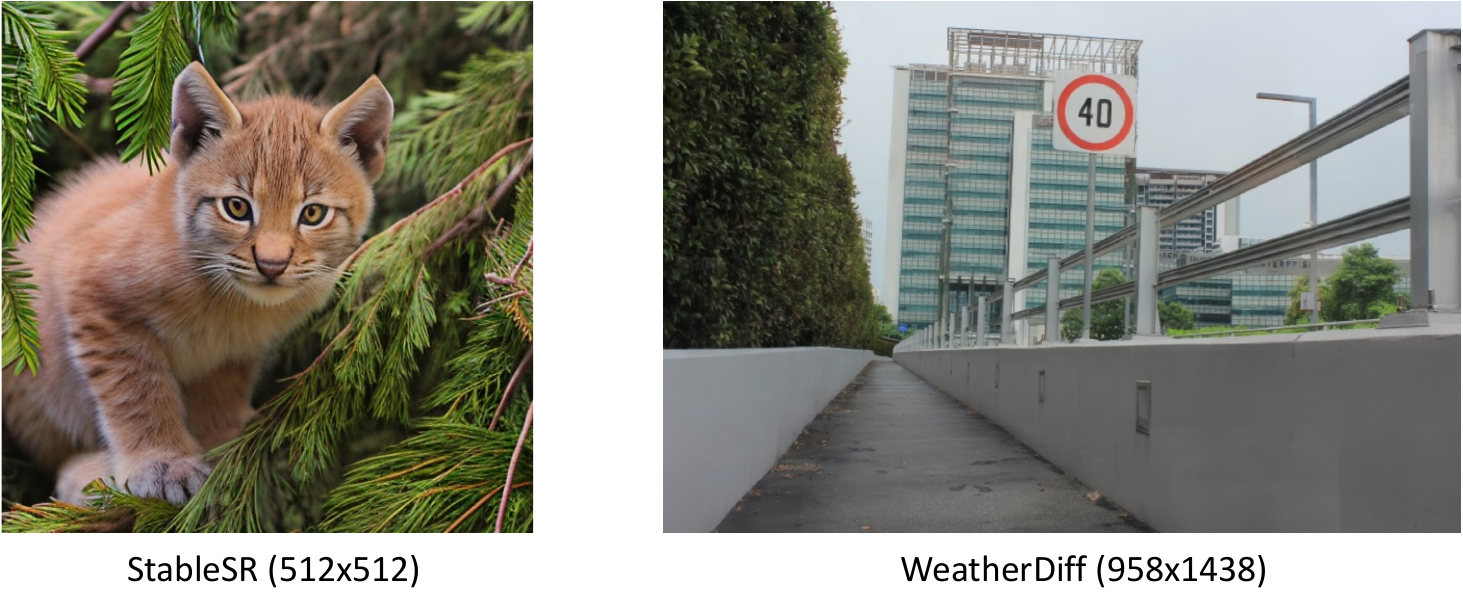}
	\caption{Qualitative results of restored images at arbitrary size.}
	\label{fig:arbitarary}
\end{figure*}

\vspace{0.5em}
\noindent\textbf{Image Restoration with Arbitrary Size.}
In general, the resolution of generated images from the diffusion model is required to be consistent with the optimization process. This limitation hinders the  diffusion model-based image restoration from processing distorted images with arbitrary sizes, especially for high-resolution images, such as 2K, and 4K. An intuitive solution is that we can generate each part of the whole image, and then stitch them into one image. However, it will cause severe mismatching issues and inconsistencies at the edges of each part, stemming from the inherent  randomness within the diffusion model.  
Recently, several papers \citep{dg-dpm,GDP,weather-diff,wang2023exploiting-stable-sr} have proposed severe effective solutions for this problem. In particular, DG-DPM \citep{dg-dpm} uses fully-convolution layers to handle inputs of arbitrary sizes, but the method is computationally expensive due to the large network structure. In contrast, Weather-diff \citep{weather-diff} and GDP \citep{GDP} both adopt patch-based restoration methods. They extract overlapping patches from input images and input each patch into the diffusion process for denoising. The overlapped parts of patches are averaged in the noise dimension to keep the consistency between different patches. Stable-SR \citep{wang2023exploiting-stable-sr} also adopts a patch-based method while it uses a Gaussian filter to smooth the noise in the overlapped parts of patches. We have shown their effectiveness in Fig.~\ref{fig:arbitarary}, where the overlapped regions among different patches are almost indistinguishable.

For traditional image SR (\ieno, bicubic downsampling), the standard training data is typically composed of DIV2K \citep{DIV2K} and Flick2K \citep{wang2019flickr1024}. However, the performance of the diffusion model is inherently bounded by the dataset size. Therefore, SR3 \citep{SR3} train 
diffusion model with ImageNet for natural image SR, and FFHQ \citep{FFHQ} for face SR. In the test process, it utilizes the ImageNet 1K \citep{imagenet1k} for the evaluation of natural image SR and CelebA-HQ for face SR. Depart from that, a series of works also introduce the commonly-used SR testing dataset for evaluation, such as Set5 \citep{set5}, Set14 \citep{Set14}, BSD100 \citep{BSD100}, Manga109 \citep{manga109}, Urban100 \citep{huang2015single-urban100}. For real-world SR, SR3+ \citep{SR3+} provide two versions of training data, where the first version is composed of DF2K and OST \citep{OST} (\ieno, DIV2K, Flick2K, and OST300), and the second version contains an extra 61M in-house image and DF2K+OST. For evaluation, the testing data is composed of RealSR \citep{cai2019toward-realsr} and DRealSR \citep{wei2020component-drealsr}, obtained by two DSLR cameras with different lenses. 

For image deblurring, the diffusion model-based methods are usually trained with GoPro training dataset \citep{Gopro}, and validated on Gopro testing dataset, RealBlur-J \citep{Real-blur}, REDS \citep{Nah_2019_CVPR_Workshops_REDS}, and HIDE \citep{HIDE}. In the shadow removal task, ISTD \citep{ISTD} and SRD \citep{SRD} are utilized for the training and evaluation, where ISTD contains 135 scenes with the shadow mask. For the image dehazing task, three typical datasets are used for evaluation, including Haze-4K \citep{liu2021synthetic-haze4k}, Dense-Haze \citep{ancuti2019dense-densehaze}, and RESIDE \citep{li2019benchmarking-reside}. Among them, Haze-4K contains $4,000$ hazy images, Dense-Haze \citep{ancuti2019dense-densehaze} is composed of 33 pairs of out-door hazy and hazy-free images, and RESIDE \citep{li2019benchmarking-reside} collected real-world 443950 training  images and 5342 testing images. Image Desnowing utilizes three datasets, \ieno, CSD \citep{golodetz2018collaborative-csd}, Snow100k \citep{snow100k}, and SRRS \citep{chen2020jstasr-srrs}. The datasets in image deraining contain diverse rain types. For example, Rain100H \citep{yang2017deep-rain100}, Rain100L \citep{yang2017deep-rain100}, Rain800 \citep{Rain800}, DDN-Data \citep{fu2017DDN-data} contains amounts of synthesis rain streaks. RainDrop \citep{rain-drop} collected 1119 pairs of rainy and clean images with various backgrounds and raindrops. Outdoor-Rain \citep{outdoor-rain} considers the rain accumulation, which provides more reasonable modeling for heavy rainy images. SPA-data \citep{wang2019SPA-data} constructs large-scale real-world rain streaks by warping the clean image from multiple continuous rainy images. More details about the above datasets can be found in Table~\ref{tab:datasets}.

\subsection{Results on Image Inpainting.}
We validate the performance of five zero-shot diffusion models on the image inpainting (narrow mask) task, as shown in Table~\ref{tab:inpainting-quantitative}. In addition to the three multi-task models, DPS \citep{dps}, DDRM \citep{DDRMDRM}, and DDNM \citep{ddnm}, we also add two models, Repaint \citep{repaint} and Copaint \citep{copaint}, specifically designed for image inpainting. In terms of distortion metrics, similar to the case of super-resolution, DDRM and DDNM achieve better performance. In terms of perceptual quality, DPS demonstrates much better perceptual performance on CelebA-HQ than on ImageNet, whereas Repaint and Copaint outperform the other models. Copaint is superior to the Repaint model, reducing the FID metric by 0.08dB and 1.33dB on CelebA-HQ and ImageNet datasets, respectively. This is because Copaint considers the coherence between the unrevealed and revealed regions from the perspective of Bayesian posterior estimation, which is more theoretically supported than the resampling strategy used in Repaint. At the same time, Copaint also employs the time travel method in DDNM \citep{ddnm} for better restoration quality, but it also increases the computation complexity. In terms of running time, DDRM remains the fastest model for generating a single image due to its low number of sampling steps (with running time almost linearly correlated with NFE). Although CoPaint has a shorter step size of 250 steps compared to DPS, it employs time travel, which increases the sampling runtime to approximately 298 seconds for generating a single image. Similarly, Repaint also requires a similar amount of time for image generation, but CoPaint's sampling time is slightly faster than that of Repaint.

\begin{table*}[htp]
\centering
\caption{Quantitative Results of Diffusion Models on Image Inpainting Task}
\resizebox{0.95\textwidth}{!}{\begin{tabular}{c|l|cccc|cccc|c|c|c}
\hline
                             &                          & \multicolumn{4}{c}{ImageNet 1K}                                                                                         & \multicolumn{4}{c|}{CelebA-HQ}                                                                                          &                                 &                                         &                                      \\ \cline{3-10}
\multirow{-2}{*}{Target}     & \multirow{-2}{*}{Models} & PSNR                         & SSIM                         & LPIPS                       & FID                         & PSNR                         & SSIM                         & LPIPS                       & FID                         & \multirow{-2}{*}{Time(s/image)} & \multirow{-2}{*}{Parameters}            & \multirow{-2}{*}{Flops (G)}          \\ \hline
&Masked Imags     & 14.86                        & 0.712                        & 0.37                        & 79.83                        & 15.67                        & 0.794                        & 0.36                       & 181.60                        &           -                 &                   -    & -    \\
                             & Repaint \citep{repaint}                  & 31.87                        & 0.967                        & \textbf{0.07}& 4.31                        & 35.23                        & 0.982                        & 0.04                        & 6.19                        & 353.4                           &  \multirow{5}{*}{552.7M(ImageNet)} & 10736.60 \\
                             & DDRM \citep{DDRMDRM}                     & 31.72                        & 0.966                        & 0.18                        & 8.82                        & \textbf{35.73 }& 0.967                        & 0.16                        & 10.82                       &  \textbf{8.8 }     &  &  1113.75   \\
                             & DPS \citep{dps}                      & 30.87                        & 0.929                        & 0.23                        & 28.32                       & 33.48                        & 0.943                        & 0.08                        & 5.37                        & 136.3                           & & 1113.75  \\
                             & DDNM \citep{ddnm}                     & \textbf{32.06}& 0.969                        & 0.11                        & 7.89                        & 35.64                        & \textbf{0.982} & 0.11                        & 7.54                        & 16.4                            &  & 1113.75  \\
\multirow{-5}{*}{Inpainting} & Copaint \citep{copaint}                  & 31.93                        &\textbf{0.976} & 0.08                        &  \textbf{4.23}& 34.97                        & 0.975                        & \textbf{0.04} &  \textbf{4.86 }& 293.8                           &  &  6083.27  \\ \bottomrule
\end{tabular}}
\label{tab:inpainting-quantitative}
\end{table*}

\subsection{Results on Composite Degradation.}
To investigate how the diffusion-based methods would behave under extra noise and composite degradations, we construct the DIV2K severe dataset following \citep{li2023learningDIL}, incorporating degradations of severe noise, blur, JPEG compression artifacts, and downsampling. The experimental results are shown in Table~\ref{table:results-blind}. We can observe that (i) the extra noise and composite degradations will cause severe performance drops. For instance, SUPIR \citep{yu2024scaling-SUPIR} achieves the PSNR of 27.19dB with simple bicubic degradation while only achieving 19.06dB in Table~\ref{table:results-blind}; (ii) Diffusion-based RealSR methods still persevere greater generalization capability compared with other diffusion-based IR methods since they introduce the distortion simulation or collected more datasets for training to eliminate the biases on unseen degradations.

\begin{table*}[htp]
\centering
\caption{Comparisons of severe mix-degradation on DIV2K. Results are tested on four FR metrics: PSNR$\uparrow$, SSIM$\uparrow$, LPIPS$\downarrow$ and two NR metrics: MUSIQ$\uparrow$, ClipIQA$\uparrow$, ManIQA$\uparrow$.}
\setlength{\tabcolsep}{2pt}
\resizebox{0.6\textwidth}{!}{
\begin{tabular}{c|cccccc}
\hline

\multirow{2}{*}{Models} & \multicolumn{6}{c}{DIV2K Severe} \\ \cline{2-7} 

                        & PSNR           & SSIM           & LPIPS                    & MUSIQ          & ClipIQA  &   ManIQA     \\ \hline
\midrule
RealESRGAN \citep{wang2021realesrgan} & 21.97 & 0.533 & 0.489 & 55.74 & 0.550 & 0.474 \\
CAL-GAN \citep{CAL-GAN}& 21.62 & 0.488 & 0.488 & 53.53 & 0.472 & 0.462\\
SeD \citep{li2024sed} & 21.87 & 0.525 & 0.468 & 59.68 & 0.549 & \textbf{0.492} \\
\hline \midrule
SwinIR \citep{swinIR-transformer} & \textbf{22.98} & \textbf{0.577} & 0.631 & 35.58 & 0.268 & 0.212\\
MambaIR \citep{guo2024mambair} & 22.32 & 0.539 & 0.413 & 57.14 & 0.523 & 0.343 \\
\hline\midrule
StableSR \citep{wang2023exploiting-stable-sr} & 20.85 & 0.389 & 0.536 & 56.50 & 0.586 & 0.352\\
DiffBIR \citep{lin2023diffbir} & 21.16 & 0.391 & 0.558 & 51.53 & 0.548 & 0.395\\
SUPIR \citep{yu2024scaling-SUPIR}   & 19.06 & 0.268 & 0.666 & 60.06 & \textbf{0.638} & 0.478\\
DiffIR \citep{xia2023diffir} & 22.10 & 0.522 & \textbf{0.377} & \textbf{63.57} & 0.602 & 0.417  \\
ResShift \citep{yue2023resshift} & 22.25 & 0.513 & 0.481 & 54.01 & 0.514 & 0.300 \\
\hline
\end{tabular}
}
\label{table:results-blind}
\end{table*}

\subsection{More visual results}
\begin{figure*}[h]
	\begin{minipage}{0.16\linewidth}
		\vspace{3pt}
    
		\centerline{\includegraphics[width=\textwidth]{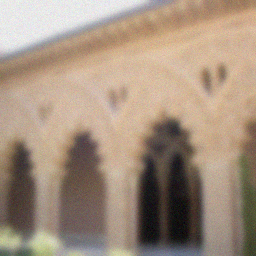}}
	\end{minipage}
	\begin{minipage}{0.16\linewidth}
		\vspace{3pt}
		\centerline{\includegraphics[width=\textwidth]{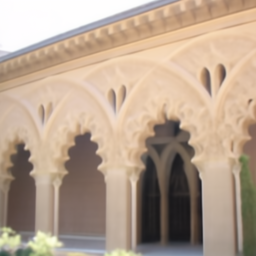}}
	 
	\end{minipage}
	\begin{minipage}{0.16\linewidth}
		\vspace{3pt}
		\centerline{\includegraphics[width=\textwidth]{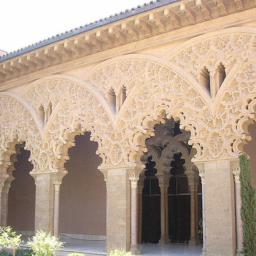}}
	 
	\end{minipage}
 \begin{minipage}{0.16\linewidth}
		\vspace{3pt}
		\centerline{\includegraphics[width=\textwidth]{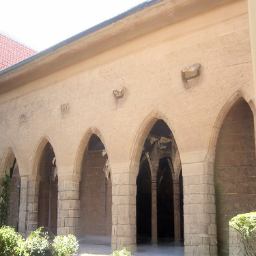}}
	\end{minipage}
 \begin{minipage}{0.16\linewidth}
		\vspace{3pt}
		\centerline{\includegraphics[width=\textwidth]{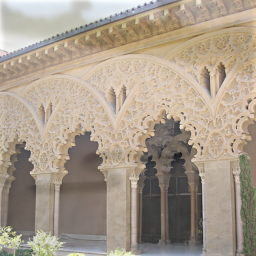}}
	\end{minipage}
 \begin{minipage}{0.16\linewidth}
		\vspace{3pt}
 
		\centerline{\includegraphics[width=\textwidth]{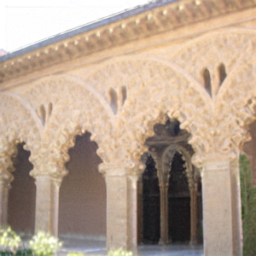}}

	\end{minipage}

\begin{minipage}{0.16\linewidth}
		\vspace{3pt}
		\centerline{\includegraphics[width=\textwidth]{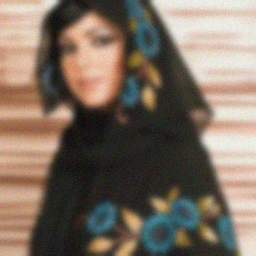}}
		\centerline{Reference}
	\end{minipage}
	\begin{minipage}{0.16\linewidth}
		\vspace{3pt}
		\centerline{\includegraphics[width=\textwidth]{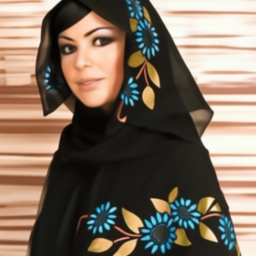}}
	 
		\centerline{DDRM}
	\end{minipage}
	\begin{minipage}{0.16\linewidth}
		\vspace{3pt}
		\centerline{\includegraphics[width=\textwidth]{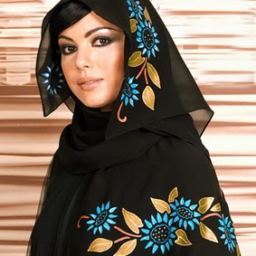}}
	 
		\centerline{DDNM}
	\end{minipage}
 \begin{minipage}{0.16\linewidth}
		\vspace{3pt}
		\centerline{\includegraphics[width=\textwidth]{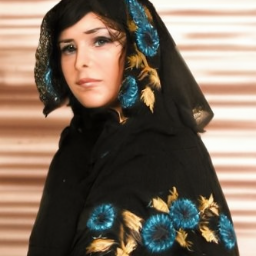}}
		\centerline{DPS}
	\end{minipage}
 \begin{minipage}{0.16\linewidth}
		\vspace{3pt}
		\centerline{\includegraphics[width=\textwidth]{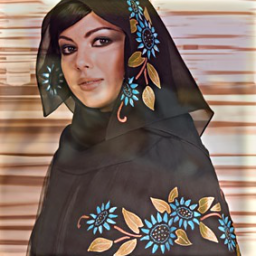}}
		\centerline{GDP}
	\end{minipage}
 \begin{minipage}{0.16\linewidth}
		\vspace{3pt}
		\centerline{\includegraphics[width=\textwidth]{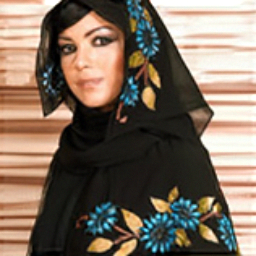}}
		\centerline{DiffPIR}
	\end{minipage}

	\caption{Qualitative results of zero-shot diffusion models on Image Deblurring  }
	\label{fig:qualitative-deblur}
\end{figure*}

\begin{figure*}[htp]
	\begin{minipage}{0.16\linewidth}
		\vspace{3pt}

		\centerline{\includegraphics[width=\textwidth]{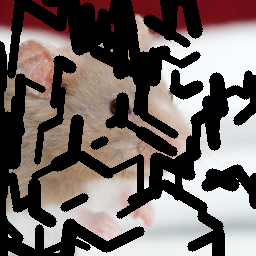}}

	\end{minipage}
	\begin{minipage}{0.16\linewidth}
		\vspace{3pt}
		\centerline{\includegraphics[width=\textwidth]{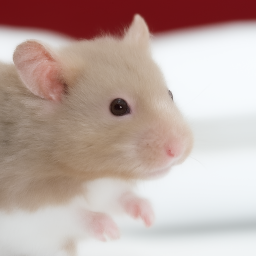}}

	\end{minipage}
	\begin{minipage}{0.16\linewidth}
		\vspace{3pt}
		\centerline{\includegraphics[width=\textwidth]{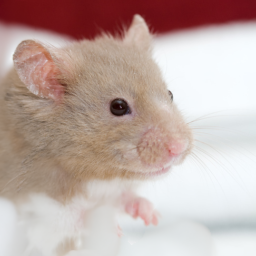}}

	\end{minipage}
 \begin{minipage}{0.16\linewidth}
		\vspace{3pt}

		\centerline{\includegraphics[width=\textwidth]{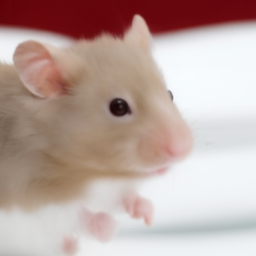}}

	\end{minipage}
 \begin{minipage}{0.16\linewidth}
		\vspace{3pt}

		\centerline{\includegraphics[width=\textwidth]{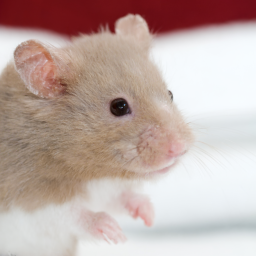}}

	\end{minipage}
 \begin{minipage}{0.16\linewidth}
		\vspace{3pt}

		\centerline{\includegraphics[width=\textwidth]{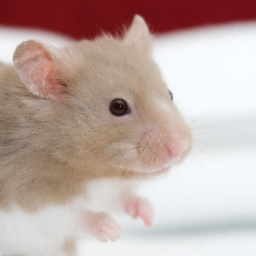}}

	\end{minipage}

\begin{minipage}{0.16\linewidth}
		\vspace{3pt}

		\centerline{\includegraphics[width=\textwidth]{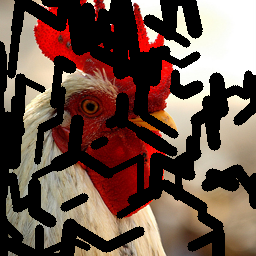}}
		\centerline{Reference}
	\end{minipage}
	\begin{minipage}{0.16\linewidth}
		\vspace{3pt}
		\centerline{\includegraphics[width=\textwidth]{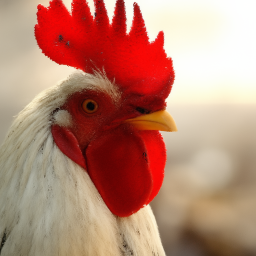}}
	 
		\centerline{DDRM}
	\end{minipage}
	\begin{minipage}{0.16\linewidth}
		\vspace{3pt}
		\centerline{\includegraphics[width=\textwidth]{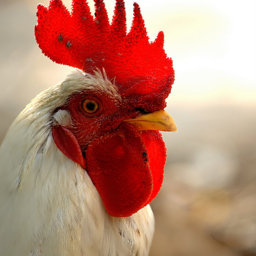}}
	 
		\centerline{DDNM}
	\end{minipage}
 \begin{minipage}{0.16\linewidth}
		\vspace{3pt}
  \centerline{\includegraphics[width=\textwidth]{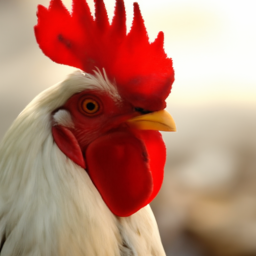}}
		\centerline{DPS}
	\end{minipage}
 \begin{minipage}{0.16\linewidth}
		\vspace{3pt}
		\centerline{\includegraphics[width=\textwidth]{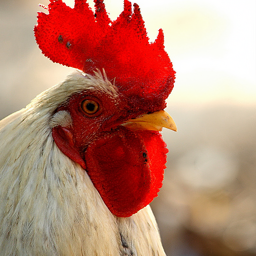}}
		\centerline{Repaint}
	\end{minipage}
 \begin{minipage}{0.16\linewidth}
		\vspace{3pt}
		\centerline{\includegraphics[width=\textwidth]{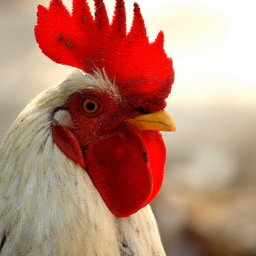}}
		\centerline{DiffPIR}
	\end{minipage}
	\caption{Qualitative results of diffusion models on Image Inpaiting.  }
	\label{fig:qualitative-inpainting}
\end{figure*}

\begin{figure*}[htp]
	\centering 
	\includegraphics[width=1\linewidth]{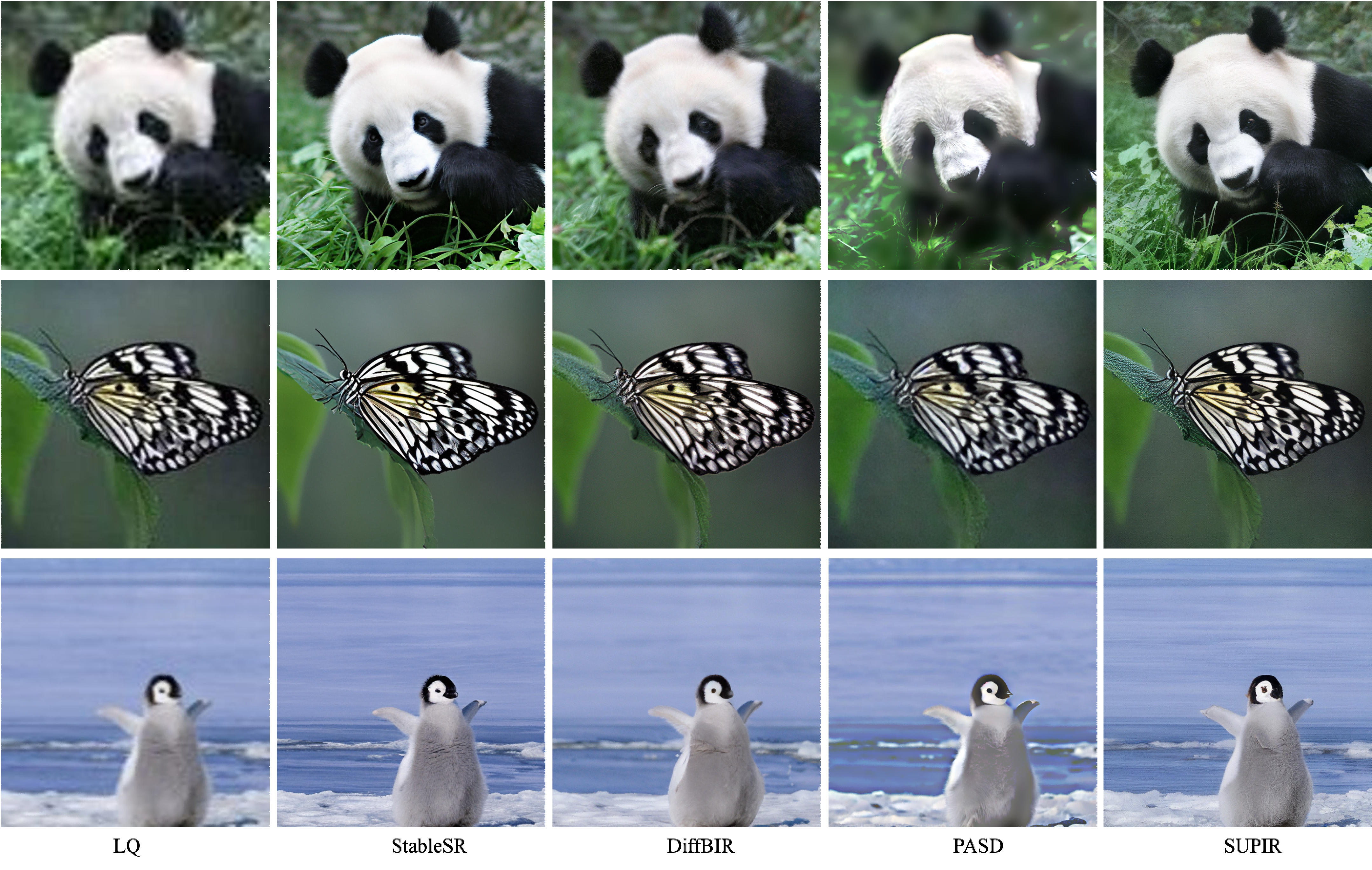}
	\caption{The visual comparisons among different DM-based IR methods on real-world dataset RealPhoto60 \citep{yu2024scaling-SUPIR}.}
	\label{fig:Visual2}
\end{figure*}

\begin{figure*}[htp]
	\centering 
	\includegraphics[width=1\linewidth]{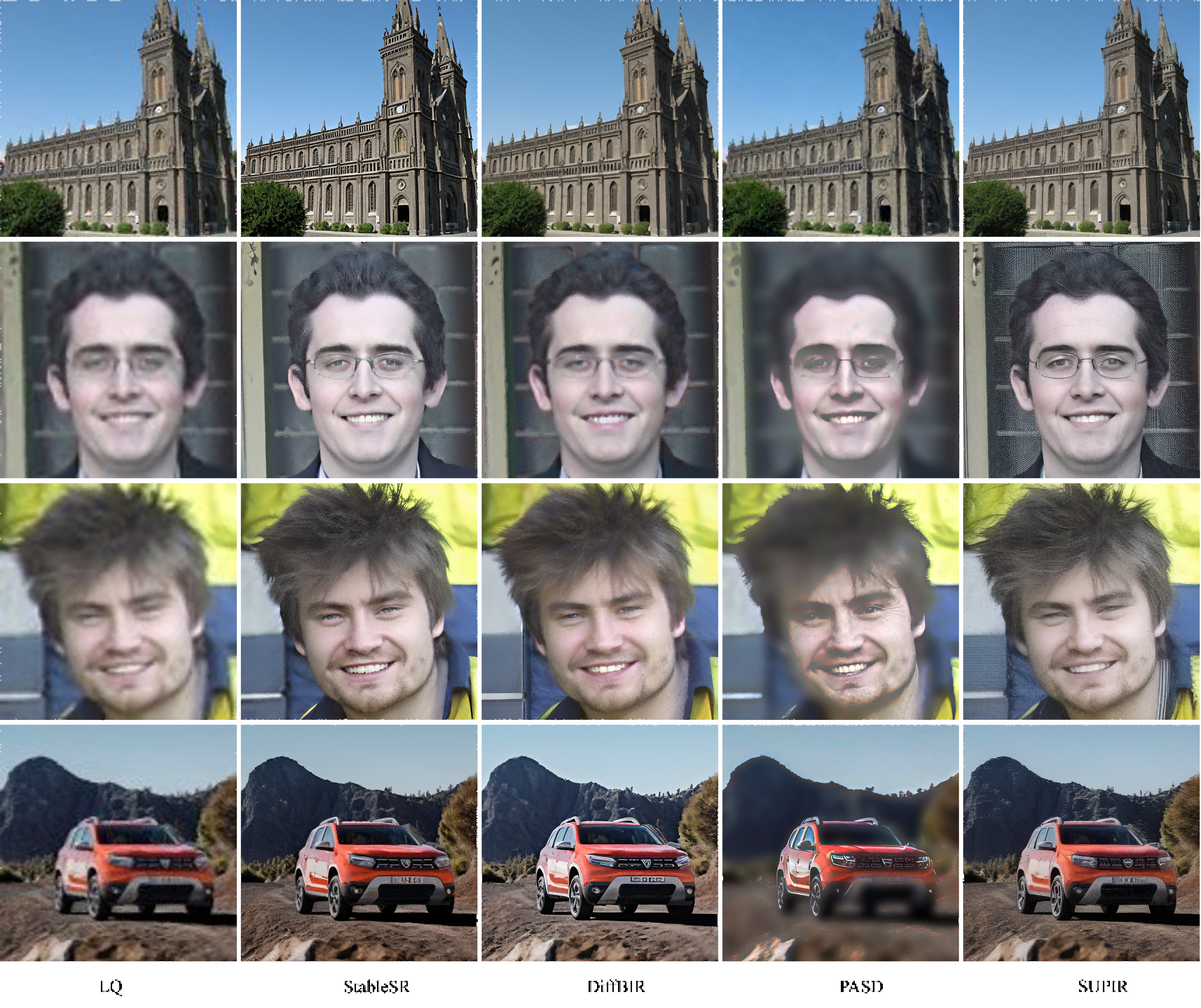}
	\caption{The visual comparisons among different DM-based IR methods on real-world dataset RealPhoto60 \citep{yu2024scaling-SUPIR}.}
	\label{fig:Visual3}
\end{figure*}

\begin{figure*}[htp]
	\centering 
	\includegraphics[width=1\linewidth]{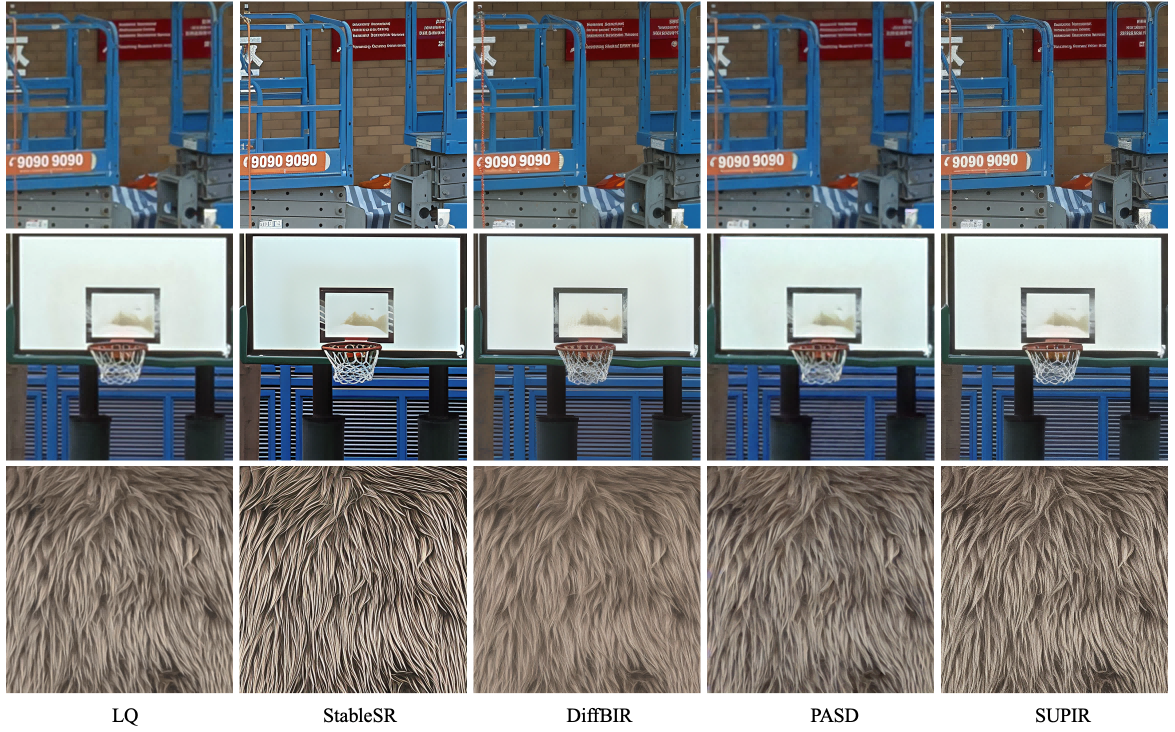}
	\caption{The visual comparisons among different DM-based IR methods on the real-world dataset RealSR \citep{cai2019toward-realsr}.}
	\label{fig:visual1}
\end{figure*}

In this section, we provides more visual results of diffusion-based IR methods on different tasks. Results on shown from Fig.~\ref{fig:qualitative-deblur} to Fig.~\ref{fig:visual1}.

Among them, Fig.~\ref{fig:qualitative-deblur} displays the qualitative images of some zero-shot models \citep{DDRMDRM,ddnm,dps,GDP,zhu2023denoising-diffpir} in image deblurring, while Fig.~\ref{fig:qualitative-inpainting} shows the results of these models on image inpainting tasks. It can be observed that DiffPIR \citep{zhu2023denoising-diffpir} excels among the compared models, especially in the recovery of structural information. On the other hand, although DPS \citep{dps} achieves restorations with perceptual quality, its fidelity is not high.
Fig.~\ref{fig:Visual3}, Fig.~\ref{fig:Visual2}, and Fig.~\ref{fig:visual1} present models of DM-based Efficient transfer learning: StableSR \citep{wang2023exploiting-stable-sr}, DiffBIR \citep{lin2023diffbir}, SUPIR \citep{yu2024scaling-SUPIR}, and PASD \citep{yang2023pixel-PASD}. Among these, SUPIR exhibits superior texture restoration details, while StableSR and DiffBIR have advantages in fidelity.

\begin{table*}[htp]
\centering
\caption{Quantitative Results of Diffusion Models on Image Deblurring Task}
\resizebox{\textwidth}{!}{\begin{tabular}{c|c|cccc|cccc|c|c}
\hline
                                      &                          & \multicolumn{4}{c|}{ImageNet 1K}                                                                                         & \multicolumn{4}{c|}{CelebaHQ Test}                                                                                              &                                 &                                     \\
\multirow{-2}{*}{Target}              & \multirow{-2}{*}{Models} & PSNR                         & SSIM                         & LPIPS                       & FID                          & PSNR                         & SSIM                         & LPIPS                       & FID                          & \multirow{-2}{*}{Time(s/image)} & \multirow{-2}{*}{Flops (G)}         \\ \hline
                                      & DPS \citep{dps}                      & 21.51                        & 0.516                        & 0.41                        & \textbf{ 52.61 }& 25.56                        & 0.688                        & 0.25                        & 65.67                        &  130.1    &  1113.75 \\
                                      & DDRM \citep{DDRMDRM}                     & 24.54                        & 0.668                        & 0.39                        & 71.65                        & 27.21                        & 0.767                        & 0.29                        & 87.32                        & \textbf{8.9 }    &  1113.75  \\
                                      & DDNM \citep{ddnm}                     & \textbf{25.37}                        & \textbf{0.731 }                       & 0.33                        & 56.54                        & 27.25                        & 0.782                        & 0.25                        & 62.65                        &  16.2     &  1113.75 \\
                                      & Dirac-DO \citep{fabian2023diracdiffusion}                 & 25.18                        & 0.719                        & 0.43                        & 75.29                        & \textbf{28.46}& \textbf{0.848 }& 0.31                        & 91.32                        & -   & -      \\
                                      & Dirac-PO \citep{fabian2023diracdiffusion}                 & 24.23                        & 0.673                        & 0.34                        & 53.91                        & 26.76                        & 0.742                        & \textbf{0.24}& 59.45                        & -   & -      \\
\multirow{-6}{*}{Gaussian Deblurring} & DiffPIR \citep{zhu2023denoising-diffpir}                  & 25.26&0.721&\textbf{ 0.32} & 53.21                        & 28.35                        & 0.832                        & 0.25                        & \textbf{58.56}&  23.5    & 1113.78          \\ \hline
\end{tabular}}
\label{tab:deblur-quantitative}
\end{table*}

\def\myurl#1{\texttt{\detokenize{#1}}}
\section{Comparisons with other surveys on diffusion models}
\textbf{(i) Earliest diffusion‐based survey dedicated to image restoration.}
To the best of our knowledge, this paper is the \emph{first} systematic and task-specific review devoted to diffusion models for image restoration and enhancement, released in August 2023 (arXiv: \url{https://arxiv.org/abs/2308.09388}). In addition, we provide an \emph{actively maintained} GitHub repository: 
~\url{https://github.com/lixinustc/Awesome-diffusion-model-for-image-processing/}, which currently exceeds 790 stars.

\textbf{(ii) Comprehensive but restoration-centric survey.} Existing surveys on diffusion models generally fall into two categories, \ieno, general-purpose survey and task-specific survey.  General-purpose surveys~\cite{chang2025designsurveychangziyi,chen2025comprehensivesurveychenhang,croitoru2023diffusioncroitoru,yang2023diffusionsurveyyangling} emphasize the development of the advanced diffusion model itself, followed by brief enumerations of downstream tasks, such as text-to-image, image translation, image editing, and a brief description of image restoration. Task-specific surveys~\cite{moser2024diffusionsurveysr,huang2025diffusionsurveyedit,liu2024diffusionsurveyremotesensing} (e.g., super-resolution~\cite{moser2024diffusionsurveysr}, editing~\cite{huang2025diffusionsurveyedit}, remote sensing~\cite{liu2024diffusionsurveyremotesensing}) provide in-depth analyses of a single application, but lack a unified view of the broader restoration domain. In contrast, our survey can balance them well. It offers the first comprehensive and focused review of diffusion models for image restoration. It systematically covers a wide spectrum of restoration tasks, including Super-resolution (SR), motion deblurring, JPEG artifacts removal, deraining/dehazing, low-light enhancement, blind real-world restoration, and image inpainting, and summarizes the common techniques and theoretical principles.

\textbf{(iii) Cross-task quantization analysis.} Existing surveys on image restoration/SR~\cite{moser2024diffusionsurveysr,luo2025tamingsurveytaming,daras2024surveyDarasinverse} often lack experimental evaluations or focus solely on a single task, such as super-resolution. In contrast, our survey provides a comprehensive cross-task analysis, covering both supervised and unsupervised settings across multiple restoration tasks, including super-resolution, inpainting, deblurring, and low-light enhancement.

\textbf{(iv) Universal methodology-centric analysis for diffusion-based image restoration}. Unlike prior surveys~\cite{daras2024surveyDarasinverse,luo2025tamingsurveytaming} that organize diffusion-based IR methods by individual tasks, our work adopts a method-centric perspective. We systematically categorize existing approaches along general methodological dimensions, including training strategies, conditioning mechanisms, architectural design, and techniques for handling real-world degradations. This unified analysis not only provides a comprehensive understanding across tasks but also establishes a general framework that has inspired subsequent task-specific surveys and covers the majority of existing diffusion-based IR methods.

\textbf{(iv) More comprehensive datasets summary.} We have comprehensively summarized 40 publicly available datasets spanning seven degradation types: super-resolution, motion deblurring, dehazing, deraining, shadow removal, and desnowing. In contrast, existing surveys~\cite{moser2024diffusionsurveysr,luo2025tamingsurveytaming,daras2024surveyDarasinverse} typically cover partial datasets or limit their scope to a single task, such as SR.

\end{document}